%% file: main.tex
\documentclass{article}

\usepackage{microtype}
\usepackage{graphicx}
\usepackage{subfigure}
\usepackage{booktabs}
\usepackage{xfrac}
\usepackage{physics}
\usepackage{amsmath,amssymb,amsthm,mathrsfs, bm}
\usepackage{stmaryrd}
\usepackage{enumitem}
\usepackage{comment}
\usepackage{dsfont}
\DeclareRobustCommand{\varlambda}{\text{\usefont{OML}{txmi}{m}{it}\symbol{"15}}}
\usepackage{cases}

\usepackage{hyperref}

\input{settings_custom.tex}

\usepackage{fancyhdr}
\usepackage{lipsum}

\usepackage{wrapfig}
\usepackage[utf8]{inputenc} 
\usepackage[T1]{fontenc}    
\usepackage{hyperref}       
\usepackage{url}            

\usepackage{amsmath, amssymb, amsthm, amsfonts}       
\usepackage{nicefrac, xfrac}       
\usepackage{microtype}      
\usepackage{xcolor}         
\usepackage{bm}
\usepackage{enumitem}
\usepackage{graphicx}
\usepackage{algorithm}
\usepackage{algorithmic}
\usepackage{geometry}
\usepackage{authblk,textcomp}
\usepackage{mathtools}
\usepackage{commath}
\usepackage{physics}
\usepackage[cal=euler]{mathalfa}
\usepackage{libertine}
\usepackage[capitalize,noabbrev]{cleveref}
\usepackage{mathrsfs}
\usepackage{enumitem}
\newlist{condenum}{enumerate}{1} 
\setlist[condenum]{label=(\roman*), ref=(\roman*)}
\crefname{condenumi}{Assumption}{Assumptions}

\usepackage{amsmath}
\usepackage{amssymb}
\usepackage{mathtools}
\usepackage{amsthm}
\usepackage{bbm}

\usepackage[capitalize,noabbrev]{cleveref}

\theoremstyle{plain}

\theoremstyle{definition}

\theoremstyle{plain}

\theoremstyle{remark}

\theoremstyle{plain}

\usepackage[textsize=tiny]{todonotes}

\newcommand{\seq}{sequence multi-index model~}

\hypersetup{pdfauthor={},pdftitle={},%
            colorlinks, linktocpage=true, pdfstartpage=1, pdfstartview=FitV,%
    breaklinks=true, pdfpagemode=UseNone, pageanchor=true, pdfpagemode=UseOutlines,%
    plainpages=false, bookmarksnumbered, bookmarksopen=true, bookmarksopenlevel=1,%
    hypertexnames=true, pdfhighlight=/O,%
    urlcolor=orange, linkcolor=blue, citecolor=blue
        }

\title{High-dimensional learning of narrow neural networks}

\author[1,2]{Hugo Cui}

\affil[1]{\small Statistical Physics Of Computation laboratory, \'Ecole Polytechnique F\'ed\'erale de Lausanne (EPFL), 1015 Lausanne, Switzerland}
\affil[2]{\small Center of Mathematical Sciences and Applications, Harvard University, Cambridge MA, USA}
\makeatletter
\newtheorem*{rep@theorem}{\rep@title}
\newcommand{\newreptheorem}[2]{%
\newenvironment{rep#1}[1]{%
 \def\rep@title{#2 \ref{##1}}%
 \begin{rep@theorem}}%
 {\end{rep@theorem}}}
\makeatother

\theoremstyle{plain}
\numberwithin{theorem}{section}
\newreptheorem{theorem}{Theorem}

\theoremstyle{remark}
\date{}

\begin{document}
\maketitle
\begin{abstract}
    Recent years have been marked with the fast-pace diversification and increasing ubiquity of machine learning applications. Yet, a firm theoretical understanding of the surprising efficiency of neural networks to learn from high-dimensional data still proves largely elusive. In this endeavour, analyses inspired by statistical physics have proven instrumental, enabling the tight asymptotic characterization of the learning of neural networks in high dimensions, for a broad class of solvable models. This manuscript reviews the tools and ideas underlying recent progress in this line of work. We introduce a generic model -- the sequence multi-index model--, which encompasses numerous previously studied models as special instances. This unified framework covers a broad class of machine learning architectures with a finite number of hidden units -- including multi-layer perceptrons, autoencoders, attention mechanisms--, and tasks --(un)supervised learning, denoising, contrastive learning--, in the limit of large data dimension, and comparably large number of samples. We explicate in full detail the analysis of the learning of sequence multi-index models, using statistical physics techniques such as the replica method and approximate message-passing algorithms. This manuscript thus provides a unified presentation of analyses reported in several previous works, and a detailed overview of central techniques in the field of statistical physics of machine learning. This review should be a useful primer for machine learning theoreticians curious of statistical physics approaches; it should also be of value to statistical physicists interested in the transfer of such ideas to the study of neural networks.
\end{abstract}

\input{Sections_masked/introduction}

\tableofcontents

\pagestyle{fancy}
\fancyhf{}
\fancyhead[R]{\selectfont\itshape \nouppercase\leftmark}
\fancyfoot[C]{\thepage}

\renewcommand{\headrulewidth}{0pt}
\input{Sections_masked/ML}

\input{Sections_masked/SP}

\input{Sections_masked/seqGLM}

\input{Sections_masked/Derivation}

\input{Sections_masked/GD_mapping}

\input{Sections_masked/Phenomenology}
\input{Sections_masked/Perspectives}
\pagestyle{fancy}
\fancyhf{}
\renewcommand{\headrulewidth}{0pt}
\bibliography{biblio}
\bibliographystyle{plain}

\pagestyle{fancy}
\fancyhf{}
\fancyhead[R]{\selectfont\itshape \nouppercase\leftmark}
\fancyfoot[C]{\thepage}
\newpage
\appendix

\input{Sections_masked/BP}

\end{document}

%% file: settings_custom.tex
\def \({\left(}
\def \){\right)}
\def \[{\left[}
\def \]{\right]}

\newcommand{\D}{\mathcal{D}}

\newcommand{\be}{\begin{equation}}
\newcommand{\ee}{\end{equation}}
\newcommand{\beqa}{\begin{eqnarray}}
\newcommand{\eeqa}{\end{eqnarray}}

\newcommand{\bea}{\begin{align}}
\newcommand{\eea}{\end{align}}

\DeclareMathAlphabet{\varmathbb}{U}{bbold}{m}{n}

\renewcommand{\vec}[1]{\boldsymbol{#1}}

\newcommand{\sign}{{\textrm{sign}}}

\newcommand{\R}[0]{{\mathbb{R}}}

\newcommand{\feat}[0]{\varphi}

\newcommand{\z}[0]{{\boldsymbol{z}}}
\newcommand{\xii}[0]{{\boldsymbol{\xi}}}
\newcommand{\x}{\boldsymbol{x}}
\newcommand{\w}{\boldsymbol{w}}

\newcommand{\teach}[0]{\boldsymbol{w}_\star}
\newcommand{\mean}[0]{\boldsymbol{\mu}}
\DeclareRobustCommand{\varlambda}{\text{\usefont{OML}{txmi}{m}{it}\symbol{"15}}}

\newcommand{\argmin}{\mathrm{argmin}}

\newcommand{\prox}{\mathrm{prox}}

\newcommand*{\colorboxed}{}
\def\colorboxed#1#{%
  \colorboxedAux{#1}%
}
\newcommand*{\colorboxedAux}[3]{%
  \begingroup
    \colorlet{cb@saved}{.}%
    \color#1{#2}%
       \boxed{%
      \color{cb@saved}%
      #3%
    }%
  \endgroup
}

\renewcommand{\boxed}[1]{
\tikz[baseline={([yshift=-1ex]current bounding box.center)}] \node [rectangle,line width=1.25, minimum width=1ex,draw, color=red] {\normalcolor $\displaystyle#1$};}
 \makeatother

\def\sign{\text{sign}}



%% file: Sections_masked/introduction.tex
\section*{Introduction}

Now ubiquitous \textit{Machine Learning} (ML) techniques have profoundly reshaped many technical fields, allowing to harness ever more abundant sources of data to automate complex tasks. Arguably driving this unremitting progress is the use of \textit{Neural Networks} (NNs), namely powerful statistical models shaped as structured stacks of neurons, to process (\textit{learn} from) large amounts of typically high-dimensional data -- images, language, video--. Standing in sharp contrast to the accruing successes of NN applications, however, is our relatively scarce theoretical understanding of their inner workings. On a technical level, prominent hurdles in building solid theoretical foundations lie in the need to analyze mathematically difficult \emph{high-dimensional}, \emph{non-linear}, \emph{non-convex} optimization problems. On a conceptual level, learning in NNs arises from the intricate interaction between its many constitutive parameters, a complexity that is challenging to analytically capture.  \\

On the other hand, these questions quite naturally fall into the orbit of statistical physics ideas and techniques, precisely developed, originally, to study the collective behaviour of interacting variables. Statistical physics ideas were first successfully applied to sharply characterize the learning curve of perceptron models by Gardner \cite{gardner1988optimal, gardner1989three}, giving the initial impulse to a long and rich line of works \cite{seung1992statistical, gyorgyi1990neural, schwarze1993learning, sompolinsky1990learning, barbier2019optimal, gabrie2018entropy, baldassi2016learning, baldassi2016unreasonable, gyorgyi1990first}. \cite{zdeborova2016statistical, Mzard2009InformationPA, Gabrie2019MeanfieldIM} are good reviews of the first successes of such analyses. While an overwhelming majority of these works focuses on the exemplar of Multi-Layer Perceptron (MLP) architectures, arguably mirroring the relative ubiquity of these models in ML empirics, recent years have witnessed a rapid diversification of learning paradigms, architectures, and tasks in ML practice. Perhaps to match this evolution, recent progress in statistical physics of machine learning has allowed the extension of the portfolio of analyzable models to encompass other architectures, such as autoencoders and attention modules. This manuscript provides a review of these developments.\\

The present manuscript reviews recent progress in the tight asymptotic analysis of neural network architectures trained with empirical risk minimization, in the limit of comparably large data dimension $d$ and number of training samples $n$, and finite number of hidden units $r=\Theta_d(1)$, using statistical physics techniques. Section \ref{sec:seq-GLM} introduces a unifying model, the \seq, which encompasses a broad class of models and architectures studied in previous works as special instances. This perspective allows a unified presentation of the analysis of these models, which we take as an opportunity to present and review central techniques in the statistical physics analysis of ML, such as the replica method and generalized approximate message-passing algorithms. This derivation, detailed in section \ref{sec: Derivation}, thus subsumes and unifies the analyses reported in many previous works \cite{mignacco2020role, loureiro2021learning2, cornacchia2023learning, aubin2020generalization, cui2023high, cui2024phase, gerace_generalisation_2020, loureiro2021learning, schroder2023deterministic, pesce2023gaussian,dAscoli2021OnTI}. Finally, section \ref{sec:perspectives} provides a brief overview of recent progress and current state of research in connex questions and topics. This review is addressed both to computer scientists, seeking a pedagogical review of statistical physics approaches to machine learning theory, and statistical physicists, looking for a physical perspective on neural networks. To the latter, we further provide in section \ref{sec:primer} a brief primer on notions and concepts in machine learning, so that this review is self-contained.

%% file: Sections_masked/ML.tex
\section{Basic concepts in machine learning}
\label{sec:primer}
\subsection{Why machine learning theory?}

ML is, above all, a collection of techniques and tools in the statistical processing of large quantities of data. It thus constitutes a branch of engineering that has, as such, historically progressed empirically, by trial and error. So why theory for ML? 
As Leo Breiman \cite{breiman2018reflections} summarizes : theory is desirable for "comfort (We knew it worked, but it’s nice to have a proof); insight: (Aha! So that’s why it works!); innovation: (At last, a mathematically proven idea that applies to data); suggestion: (Something like this might work with data)." As such, \textit{machine learning theory} is a valuable companion field to machine learning practice. It is also a very much essential guard-rail for the safe development of the field.
With the unremitting pace of practical advances and the increasing ubiquity of ML tools in all fields -- including sensitive ones such as health -- a trial-and-error approach is no longer sustainable, and a more principled basis for the development of ML is largely called for. A thorough mathematical theory of ML is, to that end, a requisite --  both to ensure a safe use of current tools, and to guide and inspire the development of novel methods. \\

Section \ref{sec:ML_review} provides a concise introduction to ML as a \textit{mathematical object} of study, highlighting in particular how it can be formulated as a random optimization problem in high dimensions.
It is primarily addressed to a physics audience seeking a simple and concise introduction to basic ideas in ML, and may be skipped by readers with a higher degree of familiarity with the field.
Section \ref{sec:statphys} and \ref{sec:seq-GLM} then illustrates, on a simple case study, how such problems can be analyzed using ideas borrowed from statistical physics. This analysis is finally pedagogically presented in section \ref{sec: Derivation}.

\subsection{The machine learning pipeline}
\label{sec:ML_review}

ML is a discipline concerned with \emph{automating complex tasks} -- by which it is understood tasks that admit no direct simple mathematical formulation --, through the statistical processing of \emph{data}. In its barest form, the ML pipeline consists of approximating a mapping $f_\star:\mathcal{X}\to\mathcal{Y}$ from the \emph{input data} $\x\in\mathcal{X}$ to its \emph{target value} (label) $f_\star(\x)\in\mathcal{Y}$. In practice, for example, the input/label pair $\x, f_\star(\x)$ can be a sentence and its translation, an image and its resolution-enhanced version, an image and its caption, or a pair of physical attributes. 
Because $f_\star$ admits in most settings no known closed-form mathematical or algorithmic formulation which can be readily coded and implemented by a computer, the ML pipeline rather aims at \emph{approximately} implementing the target function $f_\star$. To that end, ML techniques seek to leverage an informative representation (\emph{feature map}) $\feat:\mathcal{X}\to\mathcal{Z}$, which transforms and processes the original data point $\x$ into a more informative set of \emph{features} $\feat(\x)$, and subsequently approximating the target $f_\star$ typically as a linear combination$
    f_w=\w ^\top \feat(\x),$
with the \emph{weights vector} $\w$ collecting the corresponding coefficients. On an intuitive level, the features $\feat(\x)$ should thus regroup the important characteristics and attributes of the data point $\x$ in the context of the task.\\

The choice of the feature transformation $\feat$ is, for some applications, natural. Consider for instance the case of images, for which wavelet transforms offer a concise and informative representation, capturing the information encoded at multiple scales \cite{mallat2016understanding}. In practice, however, there could exist more efficient feature extraction maps $\feat$; further, in some other cases, as for natural language data, there exists no natural off-the-shelf transform candidate. It thus proves convenient to also seek an efficient representation $\feat$ inside a \emph{parametric family} $\{\varphi_\theta\}_\theta$. Allowing the feature extractor $\feat_\theta$ to be \emph{trainable} (\emph{learnable}) is the driving paradigm behind modern successes of ML techniques. The learning problem thus consists in finding the best function $f_{\hat{w},\hat{\theta}}$ among the parametric family $\{f_{w,\theta}=\w^\top \feat_\theta(\cdot)\}_{w,\theta}$. In the context of ML, the vector $\w$ is referred to as the \emph{readout weights}, and we shall sometimes refer to the feature map parameters $\theta$ as the \emph{internal weights} of the model. The selected values of these parameters $\hat{\w},\hat{\theta}$ are usually called the \emph{trained} --or \emph{learnt}-- weights.\\

To find satisfactory weights, ML techniques first seek to find a good approximation of the target $f_\star$ on a set of points $\D=\{\x^\mu, f_\star(\x^\mu)\}_{\mu=1}^n$ for which the target value is known. The set $\D$ is called the \emph{training set}, and is used as an empirical proxy for the true data distribution $P_x$. Mathematically, this procedure boils down to minimizing a notion of distance between the parametric function $f_{w,\theta}$ and the target $f_\star$ at all points of $\D$, which is referred to as the \textit{empirical} \emph{risk} (or  \emph{loss})
\begin{align}
\label{eq:intro:risk}
    \mathcal{R}(\w,\theta)=\sum\limits_{\mu=1}^n \ell\left(f_\star(\x^\mu), f_{\w,\theta}(\x^\mu)\right)
    +g\left(\w, \theta\right).
\end{align}
$\ell(\cdot)$ is a function that generically increases when its two arguments are dissimilar; $g(\cdot)$ is a --typically convex-- \emph{regularizer}, which penalizes too large weight parameters. Satisfactory values for the model weights can then be selected as minimizers of the risk \eqref{eq:intro:risk}
\begin{align}
\label{eq:intro:argmin_risk}
    \hat{\w},\hat{\theta}=\underset{\w,\theta}{\argmin}~ \mathcal{R}(\w,\theta).
\end{align}
In practice, this Empirical Risk Minimization (ERM) is numerically carried out using first-order Gradient-Descent (GD) based methods. The quality of the minimization \eqref{eq:intro:argmin_risk} is measured by the \emph{training error}
\begin{align}
    \label{eq:intro:train_error}
    \epsilon_t=\frac{1}{n}\sum\limits_{\mu=1}^n \ell_{\mathrm{tr.}}\left(
    f_\star(\x^\mu),f_{\hat{w},\hat{\theta}}(\x^\mu)
    \right),
\end{align}
for some metric $\ell_{\mathrm{tr.}}(\cdot)$, which quantifies the distance between the target $f_\star$ and the fitted parametric function $f_{\hat{\w},\hat{\theta}}$ on the points of the train set $\D$. Typically, in \emph{regression} settings where the target $f_\star$ takes continuous values -- e.g. $\mathcal{Y}=\R$ --, a popular choice is the squared error $\ell_{\mathrm{tr.}}(y,z)=\sfrac{1}{2}(y-z)^2$. In classification settings -- e.g. $\mathcal{Y}=\{-1,+1\}$--, a natural choice is the misclassification error $\ell_{\mathrm{tr.}}(y,z)=1-\delta_{y,\sign(z)}$, which measures the fraction of misclassified training samples. At the end of the training, the statistician thus has access to an approximate implementation $f_{\hat{w},\hat{\theta}}$ of the target map $f_\star$, which can be in turn readily applied to fresh data. The discrepancy between the true target and its learnt approximation is quantified by the \emph{test error}
\begin{align}
    \label{eq:intro:test_error}
    \epsilon_g=\mathbb{E}_{\x\sim P_x}\left[\ell_{\mathrm{ts.}}\left(
    f_\star(\x),f_{\hat{w},\hat{\theta}}(\x)
    \right)\right],
\end{align}
for some choice of metric $\ell_{\mathrm{ts.}}$. The test error \eqref{eq:intro:test_error} measures the discrepancy between the target $f_\star$ and the fitted parametric function $f_{\hat{\w},\hat{\theta}}$ on fresh, unseen \textit{test data}, which do not belong to the training data $\mathcal{D}$. It thus constitutes a central metric in ML to evaluate the \textit{generalization ability} of the learnt models.

\subsection{Some machine learning models}
Choosing -- and parametrizing-- a suitable feature transformation $\feat$ is thus the centerpiece of the ML pipeline. 
The development of the field has thus unsurprisingly gone hand-in-hand with a swift expansion of the zoology of possible feature map options.
In this subsection, we offer a selected digest of some standard choices, starting first from off-the-shelf fixed transforms, and secondly tunable NN feature maps.

\subsubsection{Off-the-shelf feature maps}

\paragraph{No feature map--} The simplest case is when the input $\x$ is used as is, with no further transformation, namely $\feat(\x)=\x$. The corresponding models $f_w(\x)=\w^\top \x$ are commonly regrouped under the umbrella of \emph{(generalized) linear models} (GLMs) in ML, and include canonical algorithms such as

\begin{itemize}
    \item Ridge regression $\ell(y,z)=\sfrac{1}{2}(y-z)^2, g(\w)=\sfrac{\lambda}{2}\lVert \w\lVert^2$,
    \item Logistic regression $\ell(y,z)=\ln(1+e^{-yz})$,
    \item Hinge regression $\ell(y,z)=(1-yz)_+$.
\end{itemize}
Because of their simplicity and their convexity, linear methods are very easy to train. A key limitation however lies in that they can only implement linear functions of the data, and thus suffer from poor \emph{expressivity}. In other words, they do not provide a versatile enough framework to approximate complex targets $f_\star$.

\paragraph{Kernel feature maps--} Kernel methods constitute another centerpiece of traditional ML. Given a kernel $K : \mathcal{X}\times \mathcal{X}\to \mathbb{R}$, the Moore–Aronszajn theorem ensures that the bilinear operation it defines can be rewritten in scalar product form 
\begin{align}
    \label{eq:intro:kernel_feats}
    K(\x_1,\x_2)=\langle \feat(\x_1), \feat(\x_2)\rangle_{\mathcal{H}}
\end{align}
where the \emph{kernel feature map} $\feat:\mathcal{X}\to\mathcal{H}$ maps the data in non-linear fashion to a typically large (or infinite) dimensional Hilbert space $\mathcal{H}$. Kernel learning then simply corresponds to learning using the thus defined feature map $\varphi$. Despite their relative simplicity, kernels thus provide a versatile framework to learn using non-linear features, while remaining in the realm of convex optimization. Furthermore, representer theorems such as \cite{kimeldorf1971some} imply that kernel methods can be efficiently trained, even as they tap into the expressivity of an infinite feature space.

\paragraph{Random Features (RF) -- } The closely related class of RF models were first introduced in \cite{rahimi2007random} as an efficient way to approximate kernel methods. They can alternatively be seen as NNs at initialization. Mathematically, a depth $L$ RF feature map is defined as the composition of maps
\begin{align}
    \label{eq:intro:RF}
    \feat=\psi_L\circ \dots \circ \psi_1
\end{align}
where the \emph{$\ell-$th layer} $\psi_\ell$ is the map
\begin{align}
    \label{eq:intro:layer}
    \psi_\ell(\x_{\ell-1})=\sigma_\ell(W_\ell\x_{\ell-1}).
\end{align}
In \eqref{eq:intro:layer}, $\sigma_\ell(\cdot)$ is a non-linear function, and $W_\ell$, referred to as the $\ell-th$ layer weights, is a \emph{fixed} --not trained-- random matrix. The first dimension of $W_\ell$ defines the \emph{width} of the $\ell-$th hidden layer. Aside from their connection to kernel methods, RF models provide stylized proxies for NNs, and thus afford an ideal theoretical sandbox to analytically probe some properties of the latter. 

\subsubsection{Tunable feature maps}
While ready-to-use feature maps provide computationally efficient pathways to enhance the model expressivity, the features they extract are not tailored to the data and task, and may prove sub-optimal. On the other hand, trainable maps such as Neural Network (NN) feature maps are themselves parametric and thus allow for additional tunability and versatility.

\paragraph{Multi-Layer Perceptron (MLP)-- } 
Historically, the first instance of NN is provided by the work of \cite{rosenblatt1958perceptron} on the \emph{perceptron} -- namely, a single-layer neural network in modern nomenclature. NNs \cite{lecun2015deep} build expressive feature maps by essentially stacking perceptron neurons into layers, and subsequently stacking several layers to construct deep \emph{architectures}. A depth $L$ MLP is thus defined as the composition of maps
\begin{align}
    \label{eq:intro:NN}
    \feat_{W_1,...,W_L}=\psi^{(L)}_{W_L}\circ \dots \circ \psi^{(1)}_{W_1},
\end{align}
where the $\ell-$th layer $\psi^{(\ell)}_{W_\ell}$ is defined as 
\begin{align}
    \label{eq:intro:NN_layer}
    \psi^{(\ell)}_{W_\ell}(\x_{\ell-1})=\sigma_\ell(W_\ell\x_{\ell-1}).
\end{align}
Here, the weights $W_1,..., W_L$ are importantly \emph{learnable} parameters. While MLPs are typically well suited to process vector data $\mathcal{X}=\R^d$, convolutional architectures \cite{fukushima1980neocognitron} are specifically designed to extract features from image data, leveraging convolutional and downsampling layers to take into account invariances inherent to this type of data.

\paragraph{Autoencoders (AE) --} 
AEs represent specific instances of NNs designed for self-supervised and denoising tasks, typically when the target and input spaces coincide $\mathcal{Y}=\mathcal{X}$. In its simplest two-layer instance, an AE $f_{w,\theta}(\x)=\w \feat_\theta(\x)$ is the succession of an \emph{encoder} feature map $\feat_\theta:\R^d\to\R^b$ and a \emph{decoder} layer with weights $\w\in \R^{d\times b}$. Standard AEs display a \emph{bottleneck} structure, with the hidden layer width $b$ being small. This enforces that the encoder learns a concise low-dimensional representation of the data, which can subsequently be mapped back into the original space by the decoder. Because AEs learn compact, thus a priori robust, latent representations, they are popular choices in denoising applications \cite{Vincent2010StackedDA}. AEs, and related modern denoiser architectures such as U-nets \cite{ronneberger2015u}, have further enjoyed a recent regain in interest as they find themselves at the heart of diffusion-based generative models \cite{sohl2015deep, ho2020denoising}.

\paragraph{Transformers --} Transformers \cite{vaswani2017attention} offer an efficient way of extracting features from sequential data -- such as language. Given an input $\x\in\R^{L\times d}$ of length $L$, a transformer feature map usually corresponds to a composition of a MLP and an \emph{attention layer}, defined as the map 
\begin{align}
    \label{eq:intro:attention}
    \feat_{W_Q,W_K,W_V}=\mathrm{softmax}\left(
    \x W_QW_K^\top \x^\top
    \right)\x W_V,
\end{align}
parametrized by the three trainable matrices $W_Q, W_K, W_V$. As for MLPs, the value matrix $W_V$ acts at the level of the token representations to build more informative features therefrom. Simultaneously, the $L\times L$ attention score matrix contextualizes each token, by mixing the sequence in an input-aware fashion. Crucially, transformer architectures are scalable, requiring less training time than recurrent network architectures such as long-short term memory networks \cite{bahdanau2014neural}.

\subsection{Challenges and open questions}

Despite the evident successes of day-to-day ML empirics, the field arguably still lacks a solid theoretical foundation. In fact, fundamental questions regarding the unreasonable effectiveness of statistical models such as NNs in approximating complex, typically high-dimensional functions have remained unanswered for decades. The interrogations raised by \cite{breiman2018reflections} in the $90$s --\emph{Why don't heavily parametrized neural networks overfit the data? [...] Why doesn't backpropagation head for poor local minima?} -- still remain of striking relevance in modern ML theory. On a technical level, prominent hurdles in answering those interrorgations lie in the need to analyze \emph{high-dimensional}, \emph{non-linear}, \emph{non-convex} optimization problems. On a conceptual level, learning in NNs arises from the collective effect of very large numbers of variables -- the constitutive neurons-- in interaction, a complexity that makes the characterization thereof difficult on several levels. These challenges, on the other hand, directly fall under the orbit of statistical physics techniques.

%% file: Sections_masked/SP.tex
\section{Statistical physics of neural networks}
\label{sec:statphys}

\subsection{Statistical physics in the ML researchscape}

The birth of statistical physics can be traced back to the seminal works of Maxwell, Gibbs and Boltzmann in the $19$th century, which were concerned with the study of emergent collective behaviours arising from the microscopic interaction of large assemblies of particles. More formally, the statistical physicist aims at a concise description of high-dimensional probability measures arising from large systems in interaction, in terms of a compact set of \emph{macroscopic} observables. Connecting to the previous section, learning in NNs also results from the complex interactions between a large number of variables --the eponymous neurons-- as they are jointly optimized to minimize the ERM loss. Perhaps then unsurprisingly, ML, as a field concerned with random optimization problems in high dimensions, naturally falls in the orbit of statistical physics techniques.\\

Statistical physics ideas were first successfully applied to sharply characterize the learning curve of perceptron models by Gardner \cite{gardner1988optimal, gardner1989three}, giving the initial impulse to a long and rich line of works \cite{ gyorgyi1990neural, schwarze1993learning, sompolinsky1990learning, monasson1995learning, opper1991calculation, opper1996mean} (see \cite{Mzard2009InformationPA,  opper1995statistical, watkin1993statistical, saad1999line, seung1992statistical} for reviews).
These seminal advances were accompanied by the development of powerful analysis techniques, based on the replica method \cite{parisi1979toward, parisi1983order} and approximate message-passing \cite{donoho2009message, rangan2011generalized, kabashima1998belief, kabashima2004bp, opper2001naive}, which we introduce and expound in section \ref{sec: Derivation}. These approaches share a deep connection to another central tool, the cavity method \cite{mezard1987sk, mezard1987spin}, which is not covered in the present review. The curious reader can find pedagogical expositions of this method in \cite{Mzard2009InformationPA, zdeborova2016statistical, engel2001statistical, advani2013statistical}.\\

These early works made it patent that the worst case, Probably Almost Correct (PAC) viewpoint on ML \cite{valiant1984theory}, then predominant in computer science, was too coarse to capture some finer aspects of the learning in high-dimensional models. While PAC bounds described a continuous and smooth improvement of the generalization with the number of training samples, abrupt \emph{phase transitions} to perfect generalization were found and highlighted. Such observations made the compelling case that distribution-free, worst-case analyses could sometimes prove insufficient, and that a theory of ML also required tight, typical-case analyses -- in the spirit of statistical physics. \\

As a part of theoretical physics, statistical physics analyses of ML are almost always model-driven, taking as a triple starting point:
\begin{itemize}
    \item[(a)] A specific data distribution;
    \item[(b)] A specific target function generating the labels, if the task is supervised;
    \item[(c)] A specific learning model and training procedure;
\end{itemize}
and aiming at \textit{tightly} computing -- down to the constant -- the typical (average) generalization learning curves of the model. 
Such \textit{exactly solvable models} should be simple enough to be amenable to mathematical scrutiny, while capturing key aspects of practical ML tasks. This perspective and approach now well supersede the boundaries of statistical physics, sparking works in e.g. mathematical optimization \cite{thrampoulidis2014gaussian, thrampoulidis2018precise, oymak2013squared} or random matrix theory \cite{Louart2017ARM, liao2020random, xiao2022precise}. Together with statistical physics-inspired analyses \cite{seung1992statistical, sompolinsky1990learning, gardner1988optimal, gardner1989three, engel2001statistical, opper1995statistical}, these lines of research are sometimes gathered under the umbrella of \emph{exact asymptotics} in ML theory.\\  

\subsection{Neural network modelscape}
\label{subsec:modelscape}
\begin{figure}
    \centering
    \includegraphics[scale=0.5]{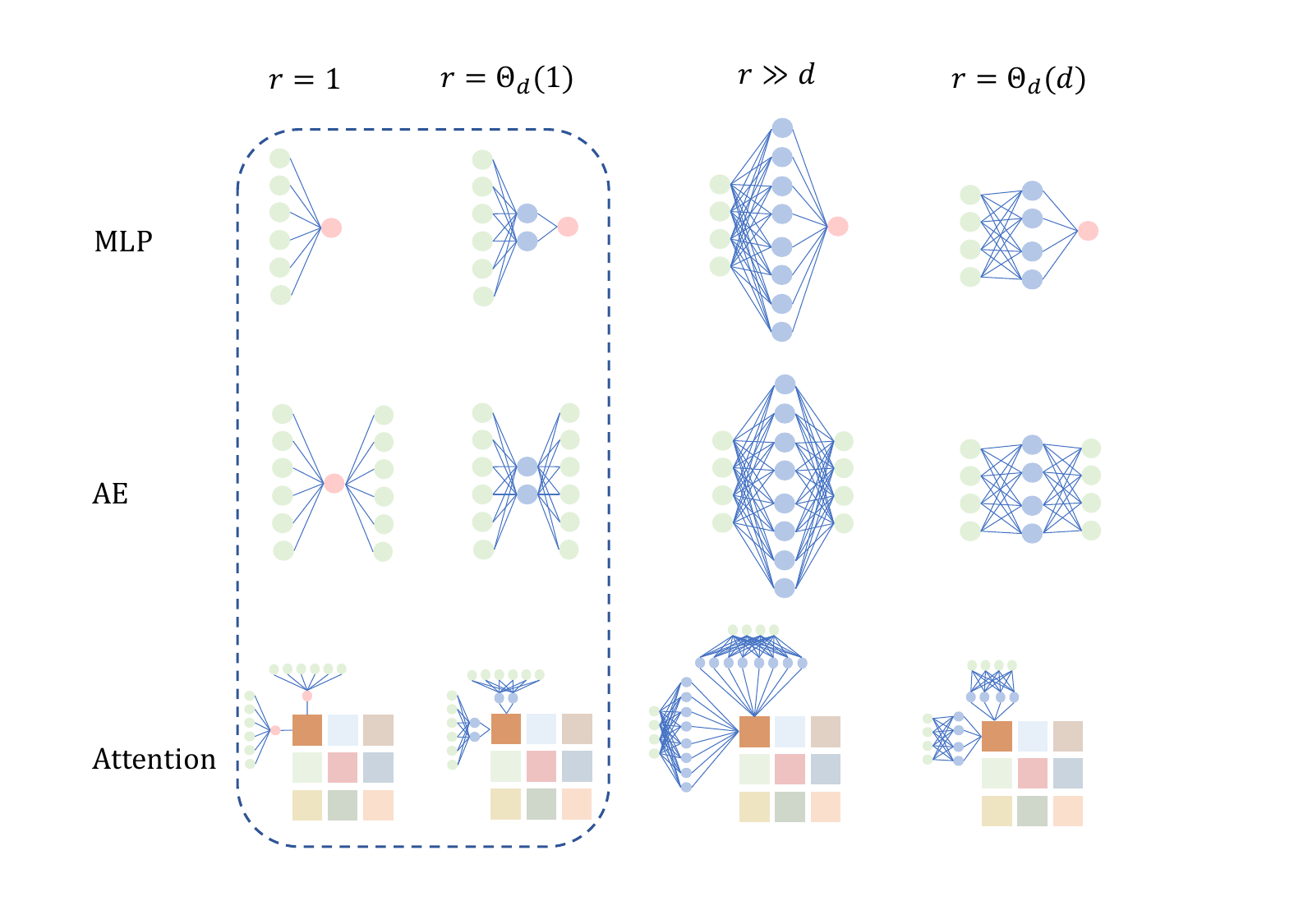}
    \caption{Graphical representation of some existing or possible models in asymptotic ML theories for MLPs (top), AEs (middle), or attention mechanisms (bottom). Each column corresponds to a different asymptotic limit for the NN architecture. From left to right: single hidden unit models, models with a finite number of hidden units, infinite-width models, and extensive-width models. Framed are the narrow architectures with $r=\Theta_d(1)$ hidden neurons which are reviewed and analyzed in unified fashion in the present review. }
    \label{fig:zoo}
\end{figure}

In an asymptotic theory, a NN model is specified by the relative scaling of its width $r$ with respect to the input dimension $d$ and the number of training samples $n$. For a given family of architectures, a whole spectrum of different asymptotic limits can thus be studied, yielding distinct theoretical models with diverse phenomenology. Some examples are represented in Fig \ref{fig:zoo}, where each row corresponds to a family of NN --MLPs, AEs, attention mechanisms-- and each column to a different scaling limit -- $r=0$, $r=\Theta_d(1)$, $r\gg d$, $r=\Theta_d(d)$--. The plurality of asymptotic limits of interest for a given type of NN is perhaps most apparent for MLPs, which have been the object of sustained theoretical scrutiny over the past decades. 

The learning of simple models with no hidden units (GLMs) has been well characterized in a very large body of works. Aspects of learning in perceptron models were studied in a series of foundational works \cite{gardner1988optimal, gardner1989three, gyorgyi1990neural, seung1992statistical, sompolinsky1990learning, opper1991calculation, monasson1994domains,opper1996mean, mezard1989space,krauth1989storage}, reviewed in \cite{opper1995statistical, engel2001statistical,saad1999line, watkin1993statistical, seung1992statistical, advani2013statistical}, with parts of these results put on rigorous footing in \cite{barbier2019optimal}. Many more results have since been reached for related models, for instance in the context of compressed sensing \cite{ganguli2010statistical, kabashima2010statistical, krzakala2012probabilistic}, or in the characterization of the geometry of the set of solutions \cite{huang2014origin, baldassi2015subdominant}, and are reviewed in \cite{malatesta2023high, advani2013statistical, Gabrie2019MeanfieldIM, gabrie2023neural}.
The analysis of single-layer NNs in high dimensional regimes is also the object of an extensive body of works in statistics and computer science. Let us mention for example \cite{candes2020phase, montanari2019generalization, Sur2020, mai2019large} for logistic regression and related classification methods, and \cite{donoho2016high, el2018impact, thrampoulidis2015regularized, bayati2011lasso, el2013robust, dobriban2018high} for regression with linear estimators.

These studies have been extended to narrow architectures with $r=\Theta_d(1)$ hidden neurons in \cite{schwarze1993learning, monasson1995learning, monasson1995weight, schwarze1992generalization, mato1992generalization, barkai1990statistical, sollich1996learning, engel1992storage,opper1994learning}, with a part of these results later proven rigorously in \cite{aubin2018committee}. The training dynamics of such NNs under online stochastic GD have been first characterized in the pioneering works \cite{saad.solla_1995_line, saad1995exact, saad1996learning, copelli1995line}, and reviewed in \cite{saad1999line, engel2001statistical, biehl2009statistical}. The study of these architectures has experienced a substantial revival of interest, for example around the topic of multiclass classification \cite{thrampoulidis2020theoretical, kini2021phase, loureiro2021learning2, cornacchia2023learning}.

Recent years have further witnessed a realization by the community that, on the opposite end of the spectrum, infinite-width networks ($r\gg d$) are also amenable to relatively easy analytical characterization due to their connection to kernel methods \cite{jacot2018neural, chizat_2019_lazy, geiger2019disentangling,mei2019mean, bordelon2020, Canatar2021SpectralBA, cui2023error, cui2021generalization,neal1996bayesian, williams96}, whose study has itself a long history (see \cite{hofmann2008kernel} for a review) -- including from a statistical physics viewpoint \cite{ dietrich1999statistical, opper1998mean, opper2001universal}. 
In the middle of the spectrum, relatively recent theoretical effort has been devoted to the study of the \emph{extensive-width limit} $r=\Theta_d(d)$ (rightmost in Fig.\,\ref{fig:zoo}) -- a regime which should allow to probe the learning of overparametrized models, while not reducing to some kernel limit \cite{li2021statistical, ariosto2022statistical, cui2023bayes, ba2022high, dandi2023twolayer, cui2024asymptotics, ZavatoneVeth2022ContrastingRA, maillard2024bayes}. Let us mention a rich recent line of works that in particular investigate the learning of extensive-width networks at initialization, namely RF models \cite{Gerace2022, mei2022generalization, hu2022universality, schroder2023deterministic, schroder2024asymptotics, aguirre2024random, defilippis2024dimension, hu2024asymptotics}. Naturally, other scalings of $r,d$ are of equal theoretical interest, as considered e.g. in \cite{camilli2023fundamental, seroussi2023separation}.\\

It is a fair observation that the study of supervised tasks with MLPs has historically constituted the primal focal point of asymptotic machine learning theory, including the statistical physics subfield thereof. This focus is perhaps reflective of the relative ubiquity and importance of such tasks and architectures in ML practice in the past two decades. While much remains to be understood, recent years have been marked by a growing realization of the need to analyze tasks and NN architectures beyond the supervised MLP exemplar, dictated by the need to mirror the rapid diversification of learning paradigms and models in ML practice -- with new architectures (e.g. transformers), learning paradigms (e.g. diffusion-based models) or data types (e.g. language). In other words, the portfolio of analyzable exactly solvable models represented in Fig.\,\ref{fig:zoo} needs to be extended also along its vertical direction, namely by considering, for a given scaling limit, different architectures.
As far as exact asymptotic studies go, let us mention recent progress in this direction for AEs \cite{Refinetti2022TheDO, cui2023high, ichikawa2023dataset}, flow-based generative models \cite{cui2023analysis}, attention models \cite{rende2023optimal, cui2024phase, lu2024asymptotic, tiberi2024dissecting}, and generative adversarial networks \cite{ichikawa2024statistical}.\looseness=-1 \\

This review presents a mathematical framework that unifies a large number of previous investigation of networks with a finite number $r=\Theta_d(1)$ of hidden units \cite{aubin2018committee, aubin2020generalization, loureiro2021learning2, mignacco2020role, cornacchia2023learning, cui2023high, cui2024phase, adomaityte2023high,pesce2023gaussian,gerace_generalisation_2020, loureiro2021learning}, by formulating and analyzing a very general model, encompassing the models considered in these works as special cases. Importantly, this framework enables the unified study of MLP, AE and attention architectures; we also highlight how it allows the study of siamese architecture models. The framework also spans a wide range of tasks, including (un)supervised learning, denoising, reconstruction, and contrastive learning.

%% file: Sections_masked/seqGLM.tex
\section{The \seq  }
\label{sec:seq-GLM}

This section introduces the \textit{\seq}, a very generic model for architectures with $\Theta_d(1)$ hidden units learning from sequential data, which encompasses as special instances a broad class of architectures -- fully-connected, auto-encoder, attention-- and tasks -- (un)supervised learning, contrastive learning, denoising--  of interest. This first section defines the model, and then proceeds to explicitly establish how the models considered in e.g. \cite{aubin2018committee, cornacchia2023learning, aubin2020generalization, cui2023high, cui2024phase, loureiro2021learning2, mignacco2020role, gerace_generalisation_2020, loureiro2021learning, schroder2023deterministic, dAscoli2021OnTI, pesce2023gaussian} represent special instances of the \seq. The next section then proceeds to present a tight asymptotic analysis of the model, unifying those presented in the aforementioned works, using the replica method from statistical physics \cite{parisi1979toward,parisi1983order, mezard1987spin, Mzard2009InformationPA}, and highlighting its connection with Generalized Approximate Message Passing (GAMP) algorithms \cite{bayati2011dynamics, rangan2016fixed, donoho2009message, donoho2010message, rangan2011generalized, javanmard2013state}.

\subsection{Model definition}

The \seq provides a simplified, yet rich, model for a neural network $f_{\w}$ with $r$ hidden units, parametrized by $\w\in\R^{d\times r}$, learning from a train set $\mathcal{D}=\{\x^\mu,y^\mu\}_{\mu=1}^n$. The latter is assumed to comprise $n$ i.i.d. samples $\x^\mu \in\R^{L\times d}$, and the corresponding labels $y^\mu$, if a supervised setting is considered. We sequentially describe the distribution of the data $\x$, and the training procedure of the network weights $\w$.

\paragraph{Data model -- } The \seq is a model of NNs learning from sequential data. Each sample $\x\in\R^{L\times d}$ is a $L\times d$ matrix, viewed as a length-$L$ sequence of $d-$ dimensional \textit{tokens} (rows) $\{\x_\ell\}_{\ell=1}^L$. For example, in the context of language data, $L$ would correspond to the length of the sentence, while $d$ represents the dimension of the space in which its constitutive words are embedded. As we will expound below, we will focus in this review on the limit of large embedding dimension $d\to\infty$, while keeping the sequence length finite, namely $L=\Theta_d(1)$. In order to design a simple model, we assume that each token $\x_\ell$ is drawn from a Gaussian mixture distribution of $K_\ell$ clusters
\begin{align}
\label{eq:data_distrib}
    \x_\ell, c_\ell\sim \sum\limits_{k=1}^{K_\ell} \rho_{\ell,k}\delta_{c_\ell,k} \mathcal{N}(\mean_{\ell,k},\Sigma_{\ell,k}),
\end{align}
where $\rho_{\ell,k}\in[0,1]$ indicates the relative probability of each cluster, and the companion random variable $c_\ell\in\llbracket 1, K_\ell \rrbracket$ indicates which cluster from the mixture a particular sample is drawn from. In particular, the law of $c_\ell$ is specified by $\rho_{\ell,1},..,\rho_{\ell,K_\ell}$. This distribution affords a simple model for a large class of practical setups. For example, the different clusters may correspond to different grammatical classes -- verb, pronoun, adjective -- if $\x_\ell$ is thought of as a word; it can model different types of objects if $\x_\ell$ represents a part of a composite picture. Therefore, on a conceptual level, each token is drawn from a vocabulary of $K_\ell$ classes, while the randomness and variability within a given class further model the idiosyncrasies of the token (e.g. which exact adjective among the class of all adjectives). Finally, we assume that the class assignments $c=\{c_\ell\}_{\ell=1}^L$ of the $L$ different tokens $\{\x_\ell\}_{\ell=1}^L$ are correlated, and follow a joint probability distribution $\rho_c$ over $\mathcal{K}\equiv\llbracket 1, K_1 \rrbracket\times \dots \times \llbracket 1, K_L \rrbracket$. We denoted $c \in\mathcal{K}$ the collection of the $L$ cluster assignments $c=\{c_\ell\}_{\ell=1}^L$. For instance, in sentences, two consecutive words are seldom from the same grammatical class. Arbitrarily complex class interactions and statistics -- e.g. hierarchical, random -- can thus be encoded in $\rho_c$.  \\

Finally, while the analysis is detailed for a finite number of classes $K_\ell$ for ease of presentation, the model -- and its analysis-- seamlessly generalizes to the infinite cluster limit $K_\ell\to\infty$. In this limit, the variable $c_\ell$ gains a continuous support $(0,\infty)$. This limit is of particular interest, since particular classes of elliptic heavy-tailed distribution \cite{beck2004superstatistics} can be constructed as infinite superposition of Gaussian clusters, as for example leveraged in \cite{adomaityte2023classification}. The data distribution \eqref{eq:data_distrib} thus also includes heavy-tailed (rather than Gaussian) token distributions as a special limiting case. As we further explicate in subsection \ref{subsec:particular_cases}, the data distribution \eqref{eq:data_distrib} encompasses as special cases a large class of data models considered in previous works, including : isotropic unimodal Gaussians \cite{cornacchia2023learning, aubin2018committee, aubin2020generalization}, structured unimodal Gaussians \cite{loureiro2021learning,  dAscoli2021OnTI, schroder2023deterministic}, (colored) Gaussian mixtures \cite{mignacco2020role, loureiro2021learning2, cui2023high}, sequences of Gaussian tokens \cite{cui2024phase, loureiro2022fluctuations}, and heavy-tailed distributions \cite{adomaityte2023classification, adomaityte2023high}.

\paragraph{Empirical Risk Minimization --} Given the training set $\mathcal{D}$, we are interested in training a neural network parametrized by a set of weights $\w\in\R^{d\times r}$ -- with $d$ the dimension of the tokens $\x_\ell$ and the width $r$ representing the number of hidden units of the network -- to complete a given task. As is standard in ML, this can be enforced by an ERM of the form
\begin{align}
    \hat{\w}=\underset{\w\in\R^{d\times r}}{\mathrm{argmin}}~\mathcal{R}(\w).
\end{align}
The \seq corresponds to empirical risks of the form
\begin{align}
\label{eq:ERM}
    \mathcal{R}(\w)=
\sum\limits_{\mu=1}^n\ell\left(
\frac{\x^\mu \teach}{\sqrt{d}},\frac{\x^\mu \w}{\sqrt{d}}, \frac{\w^\top \w}{d},c^\mu 
\right)    
+\frac{\lambda}{2}\lVert \w\lVert^2    .
\end{align}
We introduced a loss function $\ell: \R^{L\times t}\times \R^{L\times r}\times \R^{r\times r}\times \mathcal{K} \to\R_+ $. This loss function takes as an argument, in particular, the cluster assignment $c^\mu\in\mathcal{K}$ of the $\mu-$th sample, and is allowed to depend on the self-product $\sfrac{\w^\top \w}{d}$ of the weights $\w$. This latter feature shall prove convenient, as we shall cover in the following, for the analysis of AE architectures. Aside from the product $\sfrac{\x\w}{\sqrt{d}}$ between the data $\x$ and the weights $\w$, the loss function $\ell$ also takes as an argument the product $\sfrac{\x\w_\star}{\sqrt{d}}$. In \eqref{eq:ERM}, the \textit{target weights} $\teach\in\R^{d\times t}$ represent the parameters of the target function $f_\star$ (see also \eqref{eq:intro:risk}), when the latter is parametric, and intervenes when considering supervised tasks. In particular, in settings where the target function is itself of NN form -- sometimes called \textit{teacher-student} settings in statistical physics \cite{zdeborova2016statistical, Gabrie2019MeanfieldIM} --, the matrix $\teach$ represent the weights of the target network. Finally, the last term of \eqref{eq:ERM} corresponds to a $\ell_2$ regularization term. We shall discuss in subsection \ref{subsec:particular_cases} how many problems explored in previous works can be rewritten in the generic form \eqref{eq:ERM}, and thus fall under the umbrella of the \seq.\\

As in previous sections, the performance of the learnt model is characterized by two particular metrics of interest, namely the training loss
\begin{align}
    \label{eq:et}
    \epsilon_t=\mathcal{R}(\hat{\w}),
\end{align}
and the generalization error
\begin{align}
\label{eq:eg}\epsilon_g=\mathbb{E}_{\x}\ell_{\mathrm{tst.}}\left(
\frac{\x \teach}{\sqrt{d}},\frac{\x \hat{\w}}{\sqrt{d}}, \frac{\hat{\w}^\top \hat{\w}}{d},c
\right).
\end{align}

We note that the expressions \eqref{eq:ERM} and \eqref{eq:eg} are admittedly fairly abstract in form. In particular, all the details on the architecture, activations, loss function, form of the target function are subsumed and hidden inside the functions $\ell,\ell_{\mathrm{tst.}}$. In exchange, this rather terse formulation allows for a compact and unique expression that captures a very broad class of settings as special cases, as further discussed in subsection \ref{subsec:particular_cases}. In the following subsection, we detail many setups of special interest which fall under the umbrella of the formulation \eqref{eq:ERM}. Before doing so, however, we complete the formal statement of the model by specifying the high-dimensional scaling in which the learning problem \eqref{eq:ERM} shall be tightly characterized.

\paragraph{Asymptotic limit --} We will study the ERM \eqref{eq:ERM} in a particular high-dimensional asymptotic limit, sometimes referred to as the \textit{proportional regime}, which corresponds to the data dimension $d$ and number of samples $n$ diverging at comparable rates, i.e. $d,n\to\infty$ with $\alpha=\sfrac{n}{d}=\Theta_d(1)$. The finite ratio $\alpha$ is called the \textit{sample complexity}. The proportional limit constitutes the common asymptotic regimes in which almost all of the aforementioned analyses \cite{gardner1989three, seung1992statistical, sompolinsky1990learning, barbier2019optimal, aubin2018committee, mignacco2020role, aubin2020generalization, maillard2020phase, cui2020large, loureiro2021learning,loureiro2021learning2, cui2023high, cornacchia2023learning, cui2024phase, ZavatoneVeth2022ContrastingRA, li2021statistical, ariosto2022statistical} are set. Furthermore, we assume that for any $\ell, $ and $k\in \llbracket 1, K_\ell\rrbracket$, the mean of the $k-$th cluster has finite norm : $\lVert \mean_{\ell,k}\lVert=\Theta_d(1)$. Finally, all the other dimensions of the problem, including the sequence length $L$ and number of hidden units $t,r$, are also assumed to remain $\Theta_d(1)$. We now need to specify the asymptotic limit of the fixed high-dimensional parameters of the model, namely the data distribution statistics $\mean_{\ell,k},\Sigma_{\ell,k}$ \eqref{eq:data_distrib} and the target weights $\teach$ \eqref{eq:ERM} when applicable. Like in \cite{loureiro2021learning2, cui2023high, cui2024phase}, let us assume that the set of matrices $\{\{\Sigma_{\ell, k}\}_{k=1}^{K_\ell}\}_{\ell=1}^L$ admits a common set of eingevectors $\{\boldsymbol{e}_i\}_{i=1}^d$. We denote  $\{\varlambda^{\ell,k}_i\}_{i=1}^d$ the eigenvalues of $\Sigma_{\ell,k}$. The eigenvalues $\{\varlambda^{\ell,k}_i\}_{\ell,k,i}$ and the projection of the cluster means $\{\mean_{\ell,k}\}_{\ell,k}$ and the teacher columns $\{(\teach)_i\}_{i=1}^t$ on these eigenvectors are assumed to admit a well-defined joint distribution $\nu$ as $d\to \infty$ -- namely, for $\gamma=(\gamma_{\ell,k})_{\ell,k}$, $\pi=(\pi_1,...,\pi_t)\in\mathbb{R}^t$ and $\tau=(\tau_{\ell,k})_{\ell,k}$:
    \begin{align}
     \label{eq:spectral_density}   &\frac{1}{d}\sum\limits_{i=1}^d\prod\limits_{\ell=1}^L
        \prod\limits_{k=1}^{K_\ell}
        \delta\left(\varlambda_i^{\ell,k}-\gamma_{\ell,k}\right)
        \delta\left(\sqrt{d}\boldsymbol{e}_i^\top \mean_{\ell,k}-\tau_{\ell,k}\right)
        \prod\limits_{j=1}^t\delta\left(\boldsymbol{e}_i^\top (\teach)_j-\pi_j\right)\xrightarrow[]{d\to\infty} \nu\left(\gamma,\tau,\pi\right).
    \end{align}
While this limiting distribution may seem rather daunting at first, in many cases of interest considered in previous works \cite{loureiro2021learning2, cornacchia2023learning, cui2023high, cui2024phase, mignacco2020role, gerace_generalisation_2020, loureiro2021learning, schroder2023deterministic,pesce2023gaussian}, $\nu$ exhibits a rather simple form. For instance, in the often considered case of isotropic clusters, and independently Gaussian-sampled $\mean_{\ell,k}$ and $\teach$, $\nu$ decouples into the product of a Dirac and two Gaussian distribution. The spectral density $\nu$, combined with the class distribution $\rho_c$, afford a compact asymptotic description of the data distribution \eqref{eq:data_distrib}.

\subsection{Particular cases}

The \seq, specified by \eqref{eq:data_distrib} and \eqref{eq:ERM}, provides a versatile model for NN architectures with a finite number of hidden units, which encompasses a broad class of previously studied models \cite{aubin2018committee, cornacchia2023learning, aubin2020generalization, cui2023high, cui2024phase, loureiro2021learning2, mignacco2020role, gerace_generalisation_2020, loureiro2021learning, schroder2023deterministic, dAscoli2021OnTI, pesce2023gaussian} as special instances. In this subsection, we explicate in particular how previously studied models of AEs \cite{cui2023high}, GLMs \cite{mignacco2020role, cornacchia2023learning, loureiro2021learning2}, and attention \cite{cui2024phase}, but also a hitherto unstudied model
for siamese networks, can be rewritten in the form \eqref{eq:ERM}, and consequently mapped to a \seq. For each case, we first provide a brief formal description of the model, followed by a detailed derivation of the mapping.

\label{subsec:particular_cases}
\begin{figure}
    \centering
    \includegraphics[scale=0.4]{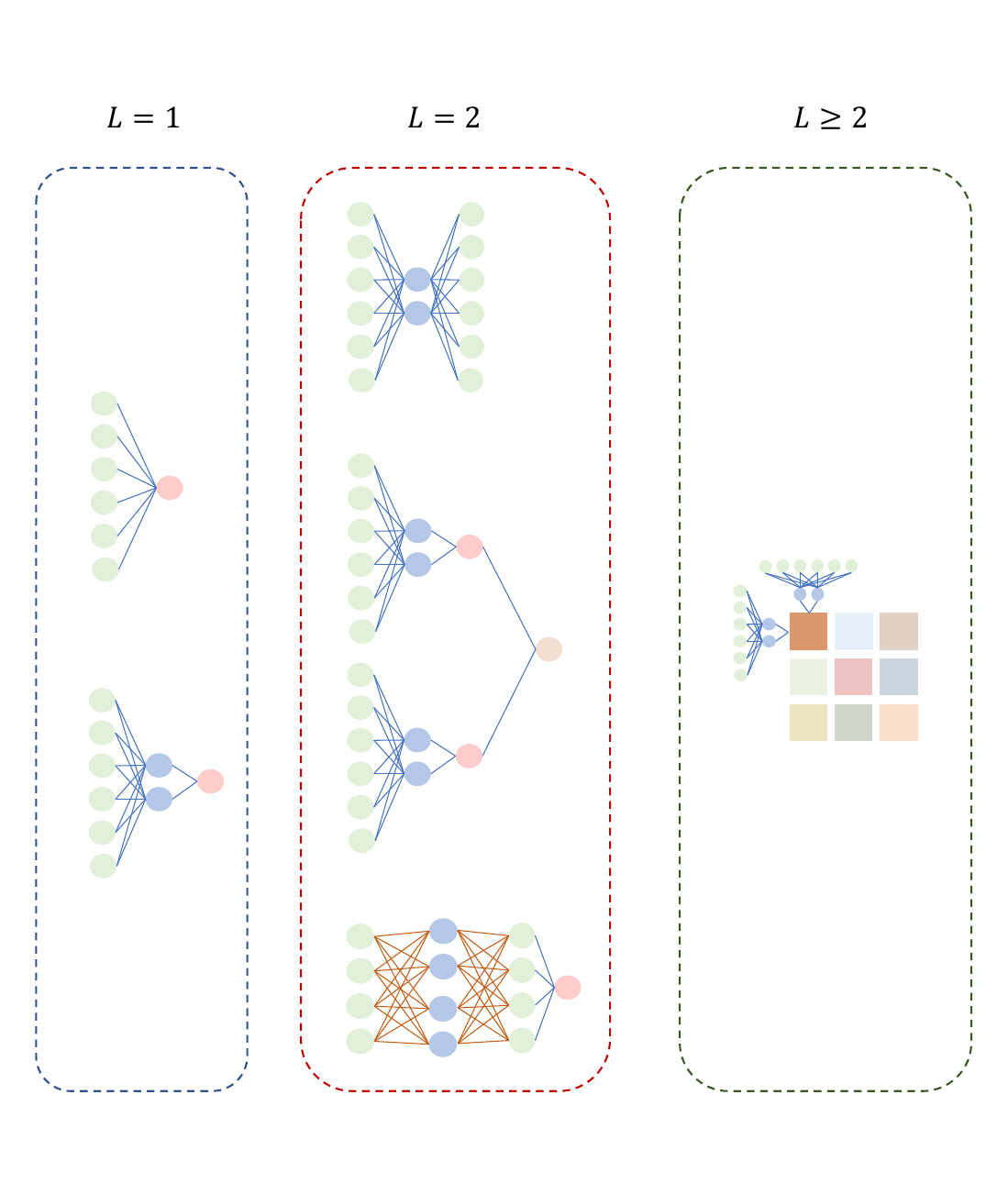}
    \caption{Special cases of interest of the \seq \eqref{eq:data_distrib}\eqref{eq:ERM}, classified by different associated sequence lengths $L$. For $L=1$, GLMs and two-layer neural networks \cite{aubin2020generalization, cornacchia2023learning, mignacco2020role, loureiro2021learning2}, discussed in \ref{subsec:perceptron}. For $L=2$, AEs \cite{cui2023high}, discussed in \ref{subsec:DAE}, siamese networks, discussed in \ref{subsec:siamese}, and RFs \cite{Gerace2022, loureiro2021learning, schroder2023deterministic}, discussed in \ref{subsec:RF}. For $L\ge 2$, attention models \cite{cui2024phase}, discussed in \ref{subsec:attention}.}
    \label{fig:seq_GLM}
\end{figure}

\subsubsection{Denoising AEs \cite{cui2023high}}
\label{subsec:DAE}
\paragraph{Model --} (\textit{reproduced from \cite{cui2023high}}) ~Let us consider data $\z\in\R^d$ drawn, for definiteness, from a Gaussian mixture distribution
\begin{align}
\label{eq:AE:data}
 \z\sim\sum\limits_{k=1}^K \rho_k\mathcal{N}(\mean_k,\Sigma_k),   
\end{align}
and corrupted by an additive Gaussian noise $\xii\sim\mathcal{N}(0, \Delta \mathbb{I}_d)$. We consider the denoising task of training a neural network $f_{\w}$ to take the noisy sample $\z+\xii$ as an input, and to output a --hopefully good-- approximation of the clean sample $\z$. As discussed in section \ref{sec:primer}, an AE architecture \cite{Vincent2010StackedDA} is a particularly simple and natural choice for such tasks. Let us in particular consider here a two-layer AE
\begin{align}
\label{eq:AE:model}
    f_{b,\w_e,\w_d}(\z)=b\z+\frac{\w_d}{\sqrt{d}}\sigma\left(\frac{\w_e^\top \z}{\sqrt{d}}\right),
\end{align}
with $\sigma$ an element-wise non-linearity, $\w_e,\w_d\in\R^{d\times r}$ the encoder (resp. decoder) weights and $b\in \R$ the strength of a trainable skip connection. Provided a train set $\mathcal{D}=\{\z^\mu+\xii^\mu,\z^\mu\}_{\mu=1}^n$ is available, training can be carried out by minimizing the \textit{denoising objective}
\begin{align}
\label{eq:AE:risk}
    \check{\mathcal{R}}(b,\w_e,\w_d)=\sum\limits_{\mu=1}^n \left\lVert 
    f_{b,\w_e,\w_d}(\z^\mu+\xii^\mu)-\z^\mu
    \right\lVert^2+\frac{\lambda}{2}\lVert \w_e\lVert^2+\frac{\lambda}{2}\lVert \w_d\lVert^2. 
\end{align}
\eqref{eq:AE:risk} and \eqref{eq:AE:model} correspond to the model analyzed in \cite{cui2023high}. As we detail below, it is possible to reduce the ERM problem \eqref{eq:AE:risk} into the \seq form \eqref{eq:ERM}. For clarity, we distinguish the full denoising objective \eqref{eq:AE:risk} from the corresponding \seq risk with a $\check{\circ}$.

\paragraph{Mapping to a \seq --} Let us now show how the data and network model specified in \eqref{eq:AE:data}, \eqref{eq:AE:model} and \eqref{eq:AE:risk} fall in the framework of the \seq, starting from the data model. Let us fix $L=2$ and define $\x\in\R^{2\times d}$ as the stacked input with rows $\x_1=\z$ and $\x_2=\xii$. It is immediate to see that the distribution of $\x$ is a particular case of that defined around \eqref{eq:data_distrib}, with in particular $K_1=K, K_2=1$, and the cluster assignment variables $c_1$ and $c_2=1$ being uncorrelated. Bringing \eqref{eq:AE:risk} into the form \eqref{eq:ERM} requires slightly more work. Expanding the square loss, one reaches
\begin{align}
    &\check{\mathcal{R}}(b,\w_e,\w_d)\notag\\
    &=\sum\limits_{\mu=1}^n (1-b)^2\lVert \z^\mu\lVert^2 +b^2 \lVert \xii^\mu\lVert^2 -b(1-b)\z^\top\xii
    \notag\\
    &+\sum\limits_{\mu=1}^n\left[ \sigma\left(\frac{\w_e^\top (\z^\mu+\xii^\mu)}{\sqrt{d}}\right)^\top \frac{\w_d^\top\w_d}{d}\sigma\left(\frac{\w_e^\top (\z^\mu+\xii^\mu)}{\sqrt{d}}\right)-2\frac{((1-b)\z^\mu-b\xii^\mu)^\top\w_d}{\sqrt{d}}\sigma\left(\frac{\w_e^\top (\z^\mu+\xii^\mu)}{\sqrt{d}}\right)
    \right]\notag\\
    &+\frac{\lambda}{2}\lVert \w_e\lVert^2+\frac{\lambda}{2}\lVert \w_d\lVert^2. 
\end{align}
Note that the first line regroups terms scaling like $\Theta_d(d^2)$ and $\Theta_d(d^{\sfrac{3}{2}})$, and furthermore only involves $b$ as the sole trainable parameter, while the terms regrouped in the second and third lines only are $\Theta_d(1)$. Therefore, the minimization problem for $b$ is dominated by the first term, which by the law of large numbers is to leading order asymptotically equal to 
\begin{align}
\check{\mathcal{R}}(b,\w)\asymp n d\left((1-b)^2\sum\limits_{k=1}^K \rho_k \int d\nu(\gamma, \tau,\pi) \gamma_k+\Delta b^2\right),
\end{align}
thus imposing that the minimizer $\hat{b}$ of the denoising risk \eqref{eq:AE:risk} should be equal to
\begin{align}
    \hat{b}=1-\frac{\Delta}{\sum\limits_{k=1}^K \rho_k \int d\nu(\gamma, \tau,\pi) \gamma_k+\Delta}.
\end{align}
The minimization of $\w$ then results from the $\Theta_d(d)$ terms in \eqref{eq:AE:risk}, which read (omitting terms independent on $\w_e, \w_d$)
\begin{align}
\label{eq:AE:risk2}
    \mathcal{R}(\w_e,\w_d)=&\sum\limits_{\mu=1}^n\left[ \sigma\!\!\left(\frac{\w_e^\top (\z^\mu+\xii^\mu)}{\sqrt{d}}\right)^\top \!\!\!\!\!\frac{\w_d^\top\w_d}{d}\sigma\!\!\left(\frac{\w_e^\top (\z^\mu+\xii^\mu)}{\sqrt{d}}\right)\!\!-2\frac{((1-\hat{b})\z^\mu-\hat{b}\xii^\mu))^\top\w_d}{\sqrt{d}}\sigma\!\!\left(\frac{\w_e^\top (\z^\mu+\xii^\mu)}{\sqrt{d}}\right)
    \right]\notag\\
    &+\frac{\lambda}{2}\lVert \w_e\lVert^2+\frac{\lambda}{2}\lVert \w_d\lVert^2.
\end{align}
To finish the mapping, let us introduce the stacked weight variable $\w\in\R^{d\times (r+r)}$, horizontally stacking $\w_e,\w_d$. Now one can rewrite \eqref{eq:AE:risk2} in the generic ERM form \eqref{eq:ERM}, with the \seq master loss $\ell$ being given by
\begin{align}
    \ell(h_\star,h, q, c)&=\ell(h,q)\notag\\
    &= \sigma\left(h_{11}+h_{21}\right)^\top q_{22}\sigma\left(h_{11}+h_{21}\right)-2((1-\hat{b})h_{12}-\hat{b}h_{22})^\top\sigma\left(h_{11}+h_{21}\right).
\end{align}
We viewed the argument $h\in\R^{2\times(r+r)}$ (resp $q\in \R^{(r+r)\times (r+r)}$) of $\ell$ as a $2\times 2$ block matrix, with blocks $h_{11}^\top,h_{12}^\top,h_{21}^\top, h_{22}^\top \in\R^{1\times r}$ (resp. $q_{11}, q_{12}, q_{21}, q_{22}\in\R^{r\times r}$). Note that for the present model, the \seq loss $\ell$ is only a function of its second and fourth arguments. Notably, the fact that it does not depend on its first argument -- which represents the label term -- directly stems from the fact that we are considering an unsupervised setting, and that there hence exists no target function in the problem.

\subsubsection{Multiclass perceptron \cite{cornacchia2023learning, loureiro2021learning2, mignacco2020role}}
\label{subsec:perceptron}
 We now turn to the multiclass perceptron model, whose tight asymptotic analysis was reported in \cite{cornacchia2023learning, loureiro2021learning2, mignacco2020role}, and show how, again, it constitutes a special case of the \seq. Since \cite{cornacchia2023learning} and \cite{loureiro2021learning2, mignacco2020role} consider the same architecture, but slightly different settings, we discuss each group of works sequentially, starting from \cite{cornacchia2023learning}. In modern nomenclature, multi-output perceptrons sometimes go by the name of \textit{committee machines}, or \textit{multi-index models} -- namely, narrow two-layer networks --  in the literature.

 \paragraph{Model--} (\textit{reproduced from \cite{cornacchia2023learning}})~ Consider Gaussian isotropic data $\x\sim\mathcal{N}(0,\mathbb{I}_d)$, and a target function parametrized by $\teach\in\R^{t\times d}$ specified by $f_\star:\R^d\to \R^t$, which for any $1\le i\le t$ is given by
 \begin{align}
     f_\star(\x)_i=\mathrm{Argmax}\left(\frac{\teach \x}{\sqrt{d}}\right)_i\equiv\delta_{i, \mathrm{argmax}_{1\le j\le t}((\teach)_j^\top \x)}.
 \end{align}
 In other words, the label corresponds to a vector with all components equal to $0$, except the one corresponding to the argmax of the data/target overlap $\teach \x \in\R^t$. Finally, consider training a perceptron of the form $f_{\w}(\x)=\phi(\sfrac{\w\x}{\sqrt{d}})$ to approximate the target, by minimizing a loss function
 \begin{align}
     \mathcal{R}(\w)=\sum\limits_{\mu=1}^n \mathcal{C}\left(f_\star(\x^\mu),\frac{\w\x}{\sqrt{d}}\right)+\frac{\lambda}{2}\lVert \w\lVert^2.
 \end{align}
 Typically, $\mathcal{C}$ would be the cross-entropy loss, commonly used for classification tasks.

 \paragraph{Mapping to a \seq--} The multiclass perceptron constitutes a particularly natural instance of the \seq. The data distribution simply corresponds to the case $L=1, K_1=1, \mean_1=0,\Sigma_1=\mathbb{I}_d$ of the \seq data density \ref{eq:data_distrib}. The \seq loss function $\ell$ is given by
 \begin{align}
     \ell(h_\star,h, q, c)&=\ell(h_\star,h)=\mathcal{C}\left(\mathrm{Argmax}(h_\star), h\right).
 \end{align}

\paragraph{The case of \cite{loureiro2021learning2,mignacco2020role} --} While \cite{loureiro2021learning2} --and \cite{mignacco2020role} in a simpler setting-- also consider classification using a single-layer network, they consider data sampled from a Gaussian mixture
\begin{align}
\x\sim\sum\limits_{k=1}^K\rho_k\mathcal{N}(\mean_k,\Sigma_k),
\end{align}
and consider the ERM
 \begin{align}
     \mathcal{R}(\w)=\sum\limits_{\mu=1}^n \mathcal{C}\left(\frac{\w\x}{\sqrt{d}}, c^\mu\right)+\frac{\lambda}{2}\lVert \w\lVert^2.
 \end{align}
 This also is a \seq, with $L=1, K_1=K$, and 
 \begin{align}
     \ell(h_\star,h, q, c)&=\mathcal{C}\left( h,c\right).
 \end{align}
 Let us end this section by mentioning that all analyses of learning with single-layer neural networks -- also called \textit{linear methods}--, as considered e.g. in \cite{aubin2020generalization}, are also special cases of the \seq, following identical mappings. This encompasses cases of particular interest such as logistic, hinge, or linear regression.

\subsubsection{Random Features \cite{gerace_generalisation_2020, loureiro2021learning, schroder2023deterministic}}
\label{subsec:RF}
 (Deep) RF networks are simple models for neural networks at initialization, and have been studied in the proportional regime in a stream of recent works, through the lens of statistical physics \cite{gerace_generalisation_2020, loureiro2021learning, schroder2023deterministic}, and rigorous methods \cite{schroder2024asymptotics, hu2022universality, mei2022generalization}. RF models constitute in fact a special class of a more generic \textit{Gaussian Covariate Model} (GCM) \cite{loureiro2021learning}, which we directly consider for generality and conciseness. The mapping from a RF model to a GCM is explicited in \cite{loureiro2021learning}, and follows from recent progress in the understanding of Gaussian universality \cite{goldt2020modeling, goldt2022gaussian, mei2022generalization, hu2022universality}, which falls out of the scope of the present review. In this section, we focus on showing how the GCM, in turn, is a special instance of the \seq.

\paragraph{Model--} (\textit{reproduced from \cite{loureiro2021learning}})~ Consider coupled Gaussian samples $\boldsymbol{u}\in\R^p,\boldsymbol{v}\in\R^d$ as 
\begin{align}
\label{eq:GCM_data}(\boldsymbol{u},\boldsymbol{v})\sim\mathcal{N}\left(0_{2d},\left(\begin{array}{cc}
        \Psi&\Phi\\
        \Phi^\top &\Omega
    \end{array}\right)\right).
\end{align}
The target function acts on the covariates $\boldsymbol{u}$ as 
\begin{align}
\label{eq:GCM_targt}f_\star(\boldsymbol{u})=\phi\left(\frac{\tilde{\teach}^\top \boldsymbol{u}}{\sqrt{p}}\right),
\end{align}
for a given function $\phi:\R\to\R$, and is parametrized by a set of weights $\tilde{\teach}\in\R^p$. The neural network, on the other hand, only has access to the covariates $\boldsymbol{v}$ -- rather than the covariates $\boldsymbol{u}$ utilized in the target function \eqref{eq:GCM_targt}-- in order to learn the target. Given a train set $\mathcal{D}=\{f_\star(\boldsymbol{u}^\mu),\boldsymbol{v}^\mu\}_{\mu=1}^n$, its training can be carried out by minimizing the empirical risk
\begin{align}
\label{eq:GCM_risk}
    \mathcal{R}(\w)=\sum\limits_{\mu=1}^n\mathcal{C}\left(f_\star(\boldsymbol{u}^\mu),\frac{\w^\top \boldsymbol{v}^\mu}{\sqrt{d}}\right)+\frac{\lambda}{2}\lVert \w\lVert^2,
\end{align}
for a set of trainable weights $\w\in\R^d$. In the special instance of RFs, $\boldsymbol{v}$ correspond to the eponymous random features. Finally, the dimension $p$ is supposed to tend to infinity jointy with $d$, with a fixed ratio $\gamma=\sfrac{p}{d}=\Theta_d(1)$.

\paragraph{Mapping to a \seq-- } Following \cite{clarte2023theoretical}, let us first observe that the random variable $\boldsymbol{u}$, following the law \eqref{eq:GCM_data}, can be equivalently rewritten as
\begin{align}
    \boldsymbol{u}=\Phi \Omega^{-1}\boldsymbol{v}+(\Psi-\Phi \Omega^{-1} \Phi^\top)^{\sfrac{1}{2}}\boldsymbol{\zeta},
\end{align}
where $\boldsymbol{\zeta}\sim\mathcal{N}(0,\mathbb{I}_d)$, assuming $\Omega$ is invertible. Therefore, the target function can be rewritten as
\begin{align}
    f_\star(\boldsymbol{v})=\phi\left(\frac{\teach^\top \boldsymbol{v}}{\sqrt{d}}+z\right),
\end{align}
where we introduced the effective target weights $\teach\equiv \sfrac{1}{ \sqrt{\gamma}}\Omega^{-1}\Phi^\top \tilde{\teach} \in \R^d$, and $z\sim\mathcal{N}(0, \sfrac{1}{p} \Tilde{\teach}^\top (\Psi-\Phi \Omega^{-1} \Phi^\top) \Tilde{\teach})$. Note that one can always rewrite $z=\sfrac{\teach^\top \boldsymbol{\xi} }{\sqrt{d}}$, provided the variance of the newly introduced isotropic Gaussian noise $\boldsymbol{\xi}$ is matched as $\boldsymbol{\xi}\sim\mathcal{N}(0, \sfrac{ 
\Tilde{\teach}^\top (\Psi-\Phi \Omega^{-1} \Phi^\top) \Tilde{\teach}}{\gamma \lVert \teach\lVert ^2}\mathbb{I}_d)$. With all these steps, the ERM problem can finally be equivalently rewritten as
\begin{align}
\label{eq:G3m_equivalent_risk}
    \mathcal{R}(\w)=\sum\limits_{\mu=1}^n\mathcal{C}\left(\phi\left(\frac{\teach^\top (\boldsymbol{v}^\mu+\boldsymbol{\xi}^\mu)}{\sqrt{d}}\right),\frac{\w^\top \boldsymbol{v}^\mu}{\sqrt{d}}\right)+\frac{\lambda}{2}\lVert \w\lVert^2.
\end{align}
The last step of the mapping consists in introducing, the stacked data $\x\in\R^{2\times d}$, with rows $\x_1=\boldsymbol{v},\x_2=\boldsymbol{\xi}$. This corresponds to the data distribution \eqref{eq:data_distrib} with sequence length $L=2$, and $K_1=K_2=1$ clusters. Finally, the ERM problem \eqref{eq:G3m_equivalent_risk} reduces to a sequence multi-index problem \eqref{eq:ERM} with effective loss
\begin{align}
    \ell(h_\star, h,q,c)&=\ell(h_\star, h)\notag\\
    &=\mathcal{C}\left(
    \phi((h_\star)_1+(h_\star)_2), h
    \right).
\end{align}

 \subsubsection{Dot-product attention \cite{cui2024phase}}
 \label{subsec:attention}

 \cite{cui2024phase} report the tight asymptotic analysis of supervised learning with a single attention layer, which is the core architectural component of transformer architectures, as discussed in section \ref{sec:primer}. We show in the following how it also falls into the \seq class.

 \paragraph{Model-- } (\textit{reproduced from \cite{cui2024phase}})~ Consider sequential data $\x\in\R^{L\times d}$, where each token $\x_\ell\in\R^d$ is sampled, independently from the others, from a Gaussian density $\mathcal{N}(\mean_\ell,\Sigma_\ell)$. \cite{cui2020large} focus on the learning of a target $f_\star:\R^{L\times d}\to \R^{L\times d}$, parametrized by $\teach\in\R^{d\times t}$, and given by
 \begin{align}
    f_\star(\x)=\mathcal{T}\left(\frac{\x\teach}{\sqrt{d}}\right)\x,
 \end{align}
 where $\mathcal{T}:\R^{L\times t}\to\R^{L\times L}$ is a matrix-to-matrix mapping. On a conceptual level, $\mathcal{T}\left(\sfrac{\x\teach}{\sqrt{d}}\right)$ represents the\textit{ target attention matrix}, which a student attention layer is tasked to learn. We consider the learning of a dot-product attention layer
 \begin{align}
 \label{eq:Attention:student}
     f_{\w_Q,\w_K}(\x)=\mathrm{softmax}\left[
     \frac{\x\w_Q\w_K^\top\x^\top}{d}
     \right]\x.
 \end{align}
 $\w_Q,\w_K\in\R^{d\times r}$ are trainable query and key weights. The model is trained by minimizing the quadratic loss
 \begin{align}
     \mathcal{R}(\w_Q,\w_K)=\frac{1}{d}\sum\limits_{\mu=1}^n\left\lVert 
     f_\star(\x^\mu)- f_{\w_Q,\w_K}(\x^\mu)
     \right\lVert^2+\frac{\lambda}{2}\lVert \w_Q\lVert^2+\frac{\lambda}{2}\lVert \w_K\lVert^2.
 \end{align}
 The mean $\mean_\ell$ of the $\ell-$the token corresponds in practice to \textit{positional encodings}, which are used to feed to the model positional information about the ordering of the tokens inside a sentence. Note that in the original work \cite{cui2024phase}, these encodings are not present in the target function. We choose here to include them for ease of presentation, at the price of a very minimal deviation from the original analysis \cite{cui2024phase}.

 \paragraph{Mapping to a \seq -- } Quite straightforwardly, the data distribution corresponds to that of a \seq \eqref{eq:data_distrib} with sequence length $L$, and $K_\ell=1$ for $1\le \ell \le L$. To ascertain the equivalent sequence multi-index loss $\ell$, let us expand the quadratic cost as
 \begin{align}
 \label{eq:Att:expansion}
     &\frac{1}{d}\left\lVert 
     f_\star(\x^\mu)- f_{\w_Q,\w_K}(\x^\mu)
     \right\lVert^2\notag\\
     &=\Tr[\mathcal{T}\left(\frac{\x\teach}{\sqrt{d}}\right)\frac{\x^\mu(\x^\mu)^\top}{d}\mathcal{T}\left(\frac{\x\teach}{\sqrt{d}}\right)^\top]+\Tr[\mathrm{softmax}\left(
     \frac{\x\w_Q\w_K^\top\x^\top}{d}
     \right)\frac{\x^\mu(\x^\mu)^\top}{d}\mathrm{softmax}\left(
     \frac{\x\w_Q\w_K^\top\x^\top}{d}
     \right)^\top]\notag\\
     &~~~~-2\Tr[\mathrm{softmax}\left(
     \frac{\x\w_Q\w_K^\top\x^\top}{d}
     \right)\frac{\x^\mu(\x^\mu)^\top}{d}\mathcal{T}\left(\frac{\x\teach}{\sqrt{d}}\right)^\top]\notag\\
     &\asymp \Tr[\mathcal{T}\left(\frac{\x\teach}{\sqrt{d}}\right)\Lambda \mathcal{T}\left(\frac{\x\teach}{\sqrt{d}}\right)^\top]+\Tr[\mathrm{softmax}\left(
     \frac{\x\w_Q\w_K^\top\x^\top}{d}
     \right)\Lambda\mathrm{softmax}\left(
     \frac{\x\w_Q\w_K^\top\x^\top}{d}
     \right)^\top]\notag\\
     &~~~~-2\Tr[\mathrm{softmax}\left(
     \frac{\x\w_Q\w_K^\top\x^\top}{d}
     \right)\Lambda\mathcal{T}\left(\frac{\x\teach}{\sqrt{d}}\right)^\top].
 \end{align}
 We denoted $\Lambda\in\R^{L\times L}$ the diagonal matrix with diagonal elements
 \begin{align}
     \Lambda_{\ell\ell}=\int d\nu(\gamma,\tau,\pi)\gamma_\ell.
 \end{align}
Finally, let us introduce the effective total weight matrix $\w\in\R^{d\times (r+r)}$, defined as the horizontal stack of $\w_Q$ and $\w_K$. The sequence multi-index loss function for the dot-product attention model can then be read from \eqref{eq:Att:expansion} to be
\begin{align}
    \ell(h_\star,h, q,c)&=\ell(h_\star, h)\notag\\
    &=\Tr[\mathrm{softmax}\left(
     h_1 h_2^\top
     \right)\Lambda\mathrm{softmax}\left(
     h_1 h_2^\top
     \right)^\top]-2\Tr[\mathrm{softmax}\left(
     h_1 h_2^\top\right)\Lambda\mathcal{T}\left(h_\star\right)^\top].
\end{align}
We viewed the argument $h\in\R^{L\times (r+r)}$ as a block matrix with two horizontally concatenated blocks $h_1,h_2\in\R^{L\times r}$, and omitted without altering the optimization problem \eqref{eq:ERM} a term constant in the weights $\w$.\\

We have so far detailed how several models and architectures considered in previous analyses \cite{aubin2020generalization, cornacchia2023learning, loureiro2021learning2, mignacco2020role, cui2023high, cui2024phase,dAscoli2021OnTI, loureiro2021learning, gerace_generalisation_2020} are special instances of the \seq \eqref{eq:data_distrib}, \eqref{eq:ERM}. The \seq, however, not only allows to unify existing models -- it also provides a flexible plug-and-play builder model for high-dimensional NN architectures.  
To illustrate this last point, we now outline one more possible model of interest encompassed in the \seq framework, which is yet -- to our awareness-- to receive a full asymptotic characterization in the literature.

\subsubsection{Siamese networks}
\label{subsec:siamese}
\textit{Siamese networks }\cite{bromley1993signature} are employed to learn to discriminate positive (similar) pairs of inputs from negative (dissimilar) pairs of inputs, by training two neural networks sharing the same weights -- the eponymous siamese networks. For example, positive pairs of inputs can be two images of signatures from the same person, while a negative pair can be an image of a signature, and an imitation thereof by a third party.

\paragraph{Model --} As a simple model for positive and negative input pairs, let us consider samples $(\x_1,\x_2)$, stacked in $\x\in\R^{2\times d}$. We will assume that both inputs $\x_1,\x_2$ are independently sampled from a binary Gaussian mixture $\sfrac{1}{2}\mathcal{N}(\mean, \mathbb{I}_d)+\sfrac{1}{2}\mathcal{N}(-\mean, \mathbb{I}_d)$, and that positive (negative) pairs correspond to pairs where $\x_1,\x_2$ were drawn from the same (different) clusters:
\begin{align}
    f_\star(\x)=\delta_{c_1,c_2},
\end{align}
where we remind that $c_1,c_2\in\{1,2\}$ are the cluster assignment. One can train two siamese perceptrons (or committee machines) $f_{\w}(\x_1)=\phi(\sfrac{\w^\top x}{\sqrt{d}}), \phi(\sfrac{\w^\top x}{\sqrt{d}})$, sharing the same weights $\w\in\R^{d}$ but acting on different tokens, to discriminate between positive and negative pairs $\x$. A standard loss to enforce the training is the\textit{ contrastive loss}
\begin{align}
\label{eq:siamese:risk}
    \mathcal{R}(\w)=\sum\limits_{\mu=1}^n \left[
    \delta_{f_\star(\x^\mu),1} (f_{\w}(\x_1^\mu)-f_{\w}(\x_2^\mu))^2+ \delta_{f_\star(\x^\mu),0}\max(0, \kappa-(f_{\w}(\x_1^\mu)-f_{\w}(\x_2^\mu))^2) 
    \right]+\frac{\lambda}{2}\lVert \w\lVert^2.
\end{align}
The contrastive loss \eqref{eq:siamese:risk} enforces that the output of the siamese networks should be close for positive pairs, but larger than a margin $\kappa>0$ for negative pairs.

\paragraph{Mapping to a \seq} The data model is, by construction, of the form \eqref{eq:data_distrib}. The sequence multi-index loss function reads
\begin{align}
    \ell(h_\star,h,q,c)&=\ell(h,c)\notag\\
    &=\delta_{c_1,c_2} (\phi(h_1)-\phi(h_2))^2+ (1-\delta_{c_1,c_2})\max(0, \kappa-(\phi(h_1)-\phi(h_2))^2).
\end{align}
Again, we denoted by $h_1,h_2$ the components of the argument $h\in\R^{2}$.\\

These several special cases thus cover many previous works, and showcase the breadth, versatility and modularity of the \seq in the design of neural network architectures in high dimensions. These many special cases are summarized in Fig.\,\ref{fig:seq_GLM}. Note however that this list is by far not exhaustive, and while we have highlighted a possible yet unstudied model, several more could be built from the basis of the \seq, in plug-and-play fashion. For example, interpreting different tokens as different patches of a flattened image can allow the study of simple, one-dimensional convolutional architectures. We leave these exciting explorations to future works, and move on to detail how the test error, train loss and several summary statistics describing the learning of a \seq \eqref{eq:ERM} can be tightly characterized in closed-form, using ideas borrowed from statistical physics. More precisely, we employ the replica method \cite{parisi1979toward, parisi1983order, Mzard2009InformationPA, mezard1987spin} to reach sharp formulae for these metrics, and then highlight how these are connected to the GAMP algorithm \cite{rangan2016fixed, bayati2011dynamics, donoho2009message, donoho2010message, rangan2011generalized, javanmard2013state}, and ultimately to the critical point of the non-convex empirical loss landscape \eqref{eq:ERM}. Importantly, this analysis consequently provides a unified derivation of those reported in \cite{mignacco2020role, cornacchia2023learning, aubin2020generalization, loureiro2021learning2, cui2023high, cui2024phase, gerace_generalisation_2020, loureiro2021learning,pesce2023gaussian, dAscoli2021OnTI} for the special instances described above.

%% file: Sections_masked/Derivation.tex
\section{Asymptotic analysis}
\label{sec: Derivation}

In this section, we discuss how properties of the trained \seq weights $\hat{\w}$  \eqref{eq:ERM} -- notably the test error $\epsilon_g$ \eqref{eq:eg} and train loss $\epsilon_t$ \eqref{eq:et} --  can be tightly asymptotically characterized using the replica method from statistical physics \cite{parisi1979toward, parisi1983order, mezard1987spin}, and how the latter is connected to GAMP algorithms \cite{ rangan2011generalized, javanmard2013state, donoho2009message} and GD. This gives us the opportunity to present, in self-contained and unified fashion, the analyses of \cite{mignacco2020role, cornacchia2023learning, aubin2020generalization, loureiro2021learning2, cui2023high, cui2024phase, gerace_generalisation_2020, loureiro2021learning,pesce2023gaussian,dAscoli2021OnTI}, carried out using these tools. Importantly, these techniques --or variants thereof-- also constitute centerpieces of the field of statistical physics of learning broadly construed, and have been the main technical tools behind many more works that are not immediate special cases of the \seq  -- e.g. \cite{baldassi2016learning, Saglietti2021SolvableMF, cui2020large, ZavatoneVeth2022ContrastingRA, gabrie2018entropy, manoel2017multi, ichikawa2023dataset, krzakala2012probabilistic}--.
So that this section might also be suitable as a primer on these techniques for a broader machine learning theory audience curious about statistical physics approaches, we present all technical steps in exhaustive detail. We also point the interested reader to the very good reviews \cite{zdeborova2016statistical, Gabrie2019MeanfieldIM} for further overviews of the same tools, in the related setting of statistical inference with single-layer GLMs (whereas the following addresses ERM \eqref{eq:ERM} with a broader class of neural networks), and unstructured data.

\subsection{The replica method}
The replica method \cite{parisi1979toward, parisi1983order, mezard1987spin} (see also \cite{Mzard2009InformationPA} for a review) starts from the simple observation that for any test function (observable) $\phi(\hat{\w})$ of the trained weights $\hat{\w}$ -- such as the test error $\epsilon_g$ \eqref{eq:eg} or the training loss $\epsilon_t$ \eqref{eq:et}--, one can write its average over the draw of the training set $\mathcal{D}$ as the integral
\begin{align}
\label{eq:intro:test_func}\mathbb{E}_{\mathcal{D}}\phi(\hat{w})=\underset{\beta\to\infty}{\lim}\mathbb{E}_{\mathcal{D}} \frac{1}{Z}\int d\w e^{-\beta \mathcal{R}(\w)}\phi(\w),
\end{align}
where we introduced the \emph{partition function} (normalization factor)
\begin{align}
    \label{eq:Z }
    Z=\int d\w e^{-\beta \mathcal{R}(\w)}.
\end{align}
Characterizing the average $\mathbb{E}_{\mathcal{D}}\phi(\hat{w})$ is thus tantamount to studying the family of $\beta-$ parametrized measures $\mathbb{P}_\beta(\w)=\sfrac{e^{-\beta\mathcal{R}(\w)}}{Z}$, for different values of the \emph{inverse temperature} $\beta>0$. To that end, it is natural to focus on studying the cumulant-generating function
\begin{align}
\label{eq:free_energy}
    f=-\underset{\beta\to\infty}{\lim}\frac{1}{\beta d}\mathbb{E}_{\mathcal{D}} \ln Z.
\end{align}
In statistical physics, $f$ is called the \emph{free energy}. Analytically evaluating the logarithm of a random variable $Z$ is, however, usually a complex enterprise. The \emph{replica method} builds on the simplifying identity
\begin{align}
\label{eq:replica_trick}
   \mathbb{E}_{\mathcal{D}}  \ln Z=\underset{s\to 0}{\lim}\frac{\mathbb{E}_{\mathcal{D}} 
 Z^s-1}{s}
\end{align}
to map it back to the simpler computation of the moment $Z^s$, for a parameter $s\to 0$. Note that $Z^s$ corresponds to the partition function (normalization) of the product measure of $s$ copies (the eponymous \textit{replicas}) of the original problem. The centerpiece of the replica method then lies in the computation of $\mathbb{E}_{\mathcal{D}} 
 Z^s$, which we detail below. \\

 \noindent The replicated partition function $Z^s$ reads
\begin{align}
    \label{eq:Zs_p1}
    \mathbb{E}_{\mathcal{D}}Z^s&=\int \prod\limits_{a=1}^s d\w_a e^{-\beta \sum\limits_{a=1}^s \frac{\lambda}{2}\lVert\w_a\lVert^2} \prod\limits_{\mu=1}^n \mathbb{E}_{\x} e^{-\beta\sum\limits_{a=1}^s  \ell\left(
    \frac{ \x\teach}{\sqrt{d}},\frac{ \x\w_a}{\sqrt{d}},\frac{\lVert \w_a\lVert^2}{d},c
    \right)}\notag\\
&= \int \prod\limits_{a=1}^s d\w_a e^{-\beta \sum\limits_{a=1}^s \frac{\lambda}{2}\lVert\w_a\lVert^2} \prod\limits_{\mu=1}^n \mathbb{E}_c\left[\mathbb{E}_{\x|c} e^{-\beta\sum\limits_{a=1}^s  \ell\left(
    \frac{ (\x-\mean_c)\teach}{\sqrt{d}}+\frac{ \mean_c\teach}{\sqrt{d}},\frac{ (\x-\mean_c)\w_a}{\sqrt{d}}+\frac{ \mean_c\w_a}{\sqrt{d}},\frac{\w_a^\top \w_a}{d},c
    \right)}\right]
\end{align}
In the last line, we used the law of total expectation to decompose $\mathbb{E}_{\x}=\mathbb{E}_c\mathbb{E}_{\x|c}$.
We denoted $\mean_c\in\R^{L\times d}$ the matrix with rows $\mean_{\ell,c_\ell}$.
Observe how all the technically challenging terms -- the non-linear loss $\ell$, and the high-dimensional average over the data $\x$ -- appear in the bracketed expression in \eqref{eq:Zs_p1}, which we shall consequently discuss in priority. To simplify the bracketed expression, let us explicitly name the random variables appearing therein as
\begin{align}
    \label{eq:intro:local_fields}
    h^a\equiv \frac{(\x-\mean_c) \w_a}{\sqrt{d}}\in\mathbb{R}^{L\times r}, &&
    h^\star\equiv\frac{(\x-\mean_c) \teach}{\sqrt{d}}\in\mathbb{R}^{L\times t}.
\end{align}
We denoted $\mu_c\in\R^{L\times d}$ the vector of means, with rows $\mean_{\ell,c_\ell}$. Note that since, conditional on $c$, $\x$ follows Gaussian statistics \eqref{eq:data_distrib}, so do $h^a, h^\star$. The rows $\{h^a_\ell,h^\star_\ell\}_{\ell=1}^L$ of the variables \eqref{eq:intro:local_fields} are thus fully characterized by their joint Gaussian statistics
\begin{align}
\label{eq:intro_overlaps}
&\mathbb{E}_{x|c}[h^a_\ell]=\mathbb{E}_{x|c}[h^\star_\ell]=0,\notag\\
    &\mathbb{E}_{x|c}[h_\ell^a(h_\kappa^b)^\top]=\delta_{\ell\kappa}\frac{\w_a^\top \Sigma_{\ell,c_\ell} \w_b}{d}\equiv q^{\ell,c_\ell}_{ab}\in\R^{r\times r},\\
    &\mathbb{E}_{x|c}[h_\ell^\star(h_\kappa^\star)^\top]=\delta_{\ell\kappa}\frac{\teach^\top \Sigma_{\ell,c_\ell} \teach}{d}\equiv \rho_{\ell,c_\ell}\in\R^{t\times t},\\
    &\mathbb{E}_{x|c}[h_\ell^a(h_\kappa^\star)^\top]=\delta_{\ell\kappa}\frac{\w_a^\top \Sigma_{\ell,c_\ell} \teach}{d}\equiv \theta^{\ell,c_\ell}_a\in\R^{r\times t}.
\end{align}
There are three more parameters that appear in the bracketed expression of \eqref{eq:Zs_p1} yet to examine, namely $\sfrac{\w_a^\top \w_a}{d}$,$\sfrac{\mean_c \w_a}{\sqrt{d}}$, $\sfrac{\mean_c \teach}{\sqrt{d}} $. These parameters do not depend on the data $\x$, and we shall treat them as order parameters. Let us rename them as
\begin{align}
\label{eq:intro:m_a}
v_a=\frac{\w_a^\top \w_a}{d} \in\R^{r\times r},&&
m_a^c\equiv\frac{\mean_c \w_a}{\sqrt{d}}\in\mathbb{R}^{L\times r},&&m_\star^c\equiv\frac{\mean_c \teach}{\sqrt{d}}\in\mathbb{R}^{L\times t}.
\end{align}
We shall further denote the rows of $m^c_a$ (resp. $m^c_\star$) as $m^{\ell, c_\ell}_a=\sfrac{\mean_{\ell,c_\ell}\w_a}{\sqrt{d}}$ (resp. $m^{\ell, c_\ell}_\star=\sfrac{\mean_{\ell,c_\ell}\teach}{\sqrt{d}}$). Using all the notations \eqref{eq:intro:local_fields}\eqref{eq:intro_overlaps}\eqref{eq:intro:m_a} thus defined, the bracketed expression in \eqref{eq:Zs_p1} can be rewritten more compactly as
\begin{align}
\label{eq:bracketed}
    \mathbb{E}_c\left[\mathbb{E}_{x|c} e^{-\beta\sum\limits_{a=1}^s  \ell\left(
    \frac{ (\x-\mean_c)\teach}{\sqrt{d}}+\frac{ \mean_c\teach}{\sqrt{d}},\frac{ (\x-\mean_c)\w_a}{\sqrt{d}}+\frac{ \mean_c\w_a}{\sqrt{d}},\frac{\w_a^\top \w_a}{d},c
    \right)}\right]=\mathbb{E}_c \mathbb{E}_{h^\star,\{h_a\}_{a=1}^s|c} e^{-\beta\sum\limits_{a=1}^s  \ell\left(h_\star+m_\star^c, h^a+m^c_a
    , v_a,c
    \right)}.
\end{align}
We thus introduced a \textit{finite} set of \textit{low-dimensional} order parameters $\{q^{\ell,k}_{ab},\rho_{\ell,k},\theta^{\ell,k}_a,v_a,m_a^{\ell, k}, m_\star^{\ell,k}\}_{1\le a,b\le s, \ell\in[L], k\in[K_\ell]}$, that fully characterize the non-trivial bracketed term in the expression of the replicated partition function $Z^s$ \eqref{eq:Zs_p1}. These parameters describe some simple statistics of the weights $\w_a$, and, as we shall later discuss, completely characterize $Z^s$. To see this, the next natural step is to shift the integration from the high-dimensional weights $\w_a$ \eqref{eq:Zs_p1} to the low-dimensional parameters $\{q^{\ell,k}_{ab},\rho_{\ell,k},\theta^{\ell,k}_a,v_a,m_a^{\ell, k}, m_\star^{\ell,k}\}$. This agenda can be carried out by inserting in \eqref{eq:Zs_p1} (Fourier-transformed) Dirac functions enforcing the definitions \eqref{eq:intro_overlaps} and \eqref{eq:intro:m_a}:
\begin{align}
\label{eq:Dirac}
    &\prod\limits_{a=1}^s \delta\left(d v_a-\w_a^\top \w_a\right) \prod\limits_{\ell=1}^L\prod\limits_{k=1}^{K_\ell}\prod\limits_{a=1}^s\delta\left(\sqrt{d} m^{\ell, k}_a-\mean_{\ell,k}\w_a\right) \delta\left(d\theta_a^{\ell,k} -\w_a^\top \Sigma_{\ell,c_\ell} \teach\right)\prod\limits_{a\le b}^s \delta\left(dq^{\ell,k}_{ab}- \w_a^\top \Sigma_{\ell,c_\ell} \w_b\right)\notag\\
    &=\int \prod\limits_{a=1}^s  d\hat{v}_a \prod\limits_{\ell=1}^L\prod\limits_{k=1}^{K_\ell}\prod\limits_{a=1}^s d\hat{m}^{\ell, k}_a d\hat{\theta}_a^{\ell,k}\prod\limits_{a\le b}^s dq^{\ell,k}_{ab}e^{\!\!\!\!-d\sum\limits_a \sum\limits_\ell\sum\limits_{k}\left[\hat{m}^{\ell,k\top}_a\!\!m_a^{\ell,k}+ \Tr(\theta^{\ell,k}_a\hat{\theta}^{\ell, k \top}_a)\right]-d\sum\limits_\ell\sum\limits_{k}\sum\limits_{1\le a\le b\le s} \!\!\!\!\!\!\Tr(q_{ab}^{\ell,k}\hat{q}_{ab}^{\ell, k\top}) -d\sum\limits_a \Tr[v_a\hat{v}_a^\top]} 
\end{align}
This rewriting necessitates the introduction of a collection of conjugate variables $\{\hat{q}^{\ell,k}_{ab},\hat{\theta}^{\ell,k}_a,\hat{v}_a,\hat{m}_a^{\ell, k}\}$.
Plugging \eqref{eq:bracketed} and \eqref{eq:Dirac} into \eqref{eq:Zs_p1}, the replicated partition function \eqref{eq:Zs_p1} can be rewritten as
\begin{align}
    \label{eq:Zs_p2}
    &\mathbb{E}_{\mathcal{D}}Z^s=\int \prod\limits_{a=1}^s dv_a d\hat{v}_a \prod\limits_{\ell=1}^L\prod\limits_{k=1}^{K_\ell}\prod\limits_{a=1}^sdm^{\ell, k}_a d\hat{m}^{\ell, k}_a d\theta_a^{\ell,k} d\hat{\theta}_a^{\ell,k}\prod\limits_{a\le b}^s dq^{\ell,k}_{ab}d\hat{q}^{\ell,k}_{ab}\notag\\
    & \underbrace{e^{-d\sum\limits_a \sum\limits_\ell\sum\limits_{k}\left[\hat{m}^{\ell,k\top}_am_a^{\ell,k}+ \Tr(\theta^{\ell,k}_a\hat{\theta}^{\ell, k \top}_a)\right]-d\sum\limits_\ell\sum\limits_{k}\sum\limits_{1\le a\le b\le s} \Tr(q_{ab}^{\ell,k}\hat{q}_{ab}^{\ell, k\top}) -d\sum\limits_a \Tr[v_a\hat{v}_a^\top]} 
    }_{e^{sd\beta\Psi_t}} \notag\\
    &\underbrace{\int \prod\limits_{a=1}^s d\w_a e^{ \sum\limits_{a=1}^s -\beta \frac{\lambda}{2}\lVert\w_a\lVert^2+\Tr[\hat{v}_a\w_a^\top \w_a]+\sum\limits_a \sum\limits_\ell\sum\limits_{k}(\sqrt{d}\hat{m}^{\ell,k\top}_a \w_a^\top \mean_{\ell,k}+ \Tr[\hat{\theta}^{\ell,k}_a \teach^\top \Sigma_{\ell,k} \w_a])+\sum\limits_{1\le a\le b\le s}\sum\limits_\ell \sum\limits_{k}\Tr[\hat{q}^{\ell,k}_{ab}\w_b^\top\Sigma_{\ell,k} \w_a]}}_{e^{sd\beta\Psi_w}} \notag\\
    &\underbrace{\left[\mathbb{E}_c \mathbb{E}_{h^\star,\{h_a\}_{a=1}^s|c} e^{-\beta\sum\limits_{a=1}^s  \ell\left(h_\star+m_\star^c, h^a+m^c_a
    , v_a,c
    \right)}\right]^{\alpha d}}_{e^{s\alpha d \beta\Psi_y}}.
\end{align}
 For easier bookkeeping, we have decomposed the replicated free entropy into the \textit{trace, entropic and energetic potentials} $\Psi_t, \Psi_w, \Psi_y$, which we shall study in turn in the following. 
 Loosely speaking, the trace and entropic potentials $\Psi_t, \Psi_w$ measure the volume (or entropy) of weight configurations corresponding to a given value of the descriptors $\{q_{ab}^{\ell,k}, \theta^{\ell,k}_a, m_a^{\ell,k},v_a\}$. The energetic potential $ \Psi_y$, on the other hand, measures the average loss corresponding to weights associated with those descriptors.
 Importantly, note that all exponents in the integrand \eqref{eq:Zs_p2} scale as $\Theta_d(d)$. Therefore the integral in \eqref{eq:Zs_p2} concentrate as $d \to \infty$, and can consequently be computed using a \textit{Laplace saddle-point approximation}, and be rephrased as an \textit{extremization} -- rather than a more intricate integration -- problem.

\subsubsection{Replica-Symmetric ansatz}
\label{subsec:replica}
We have thus rephrased the analysis of the high-dimensional average \eqref{eq:intro:test_func} as an optimization problem over the low-dimensional order parameters $\{q_{ab}^{\ell,k}, \theta^{\ell,k}_a, m_a^{\ell,k},v_a\}$, and the associated conjugate variables. While conceptually simpler, this optimization still bears over $2\times (r^2s+ (K_1+\dots+K_L)\times (r^2\sfrac{s(s+1)}{2}+rts+rs))$ scalar variables. Besides, note that the number of order parameters depends on $s$, and we need to take the analytical continuation $s\to 0$ -- in other words, the extremization formally bears on an \textit{non-integer, vanishing} number of order parameters. To circumvent these conceptual pitfalls and in order to make progress, one can look for the extremizer of the exponent of \eqref{eq:Zs_p2} in a specific and simple form. A particular prescription is the \textit{Replica Symmetric} (RS) ansatz \cite{parisi1983order,parisi1979toward}
\begin{align}
    \label{eq:intro:RS}
    &q^{\ell,k}_{ab}=(r_{\ell,k}-q_{\ell,k})\delta_{ab}+q_{\ell,k},\\
    &m^{\ell,k}_a=m_{\ell,k},\\
    & \theta^{\ell,k}_a=\theta_{\ell,k},\\
    &v_a=v,\\
    &\hat{q}^{\ell,k}_{ab}=-\left(\sfrac{\hat{r}_{\ell,k}}{2}+\hat{q}_{\ell,k}\right)\delta_{ab}+\hat{q}_{\ell,k},\\
    &\hat{m}^{\ell,k}_a=\hat{m}_{\ell,k},\\
    & \hat{\theta}^{\ell,k}_a=\hat{\theta}_{\ell,k},\\
    &\hat{v}_a=-\frac{1}{2}\hat{v}.
\end{align}
The RS ansatz implies the extremization problem associated with the evaluation of \eqref{eq:Zs_p2} now bears over $8(K_1+...+K_L)+2$ order parameters, allowing to greatly simplify the remainder of the computation.
Crucially, motivated by the definition of these quantities, we assume that $r_{\ell,k},q_{\ell,k}, v, \hat{r}_{\ell,k},\hat{q}_{\ell,k},\hat{v}$ are symmetric matrices.
In words, 
the RS ansatz assumes that the overlaps between any two distinct replicas are identical, and that all replicas further share the same overlap with the target weights. The RS ansatz \eqref{eq:intro:RS} is in particular always correct for convex problems \cite{zdeborova2016statistical}, but remains otherwise an assumption. In addition, the RS ansatz has been bolstered by compelling numerical support in a large majority of settings explored in the literature \cite{mignacco2020role, cornacchia2023learning, aubin2020generalization, loureiro2021learning2, cui2023high, cui2024phase, gerace_generalisation_2020, loureiro2021learning,pesce2023gaussian,dAscoli2021OnTI, ZavatoneVeth2022ContrastingRA, ichikawa2023dataset, cui2020large}, displaying a very good agreement with numerical experiments. The RS ansatz also bears profound connections with algorithmic schemes like AMP \cite{donoho2009message, donoho2010message, bayati2011dynamics, rangan2011generalized, rangan2016fixed, javanmard2013state} or GD, as we subsequently discuss in section \ref{susbsec:GAMP}. Let us mention that other ansatz, incorporating \textit{replica symmetry breaking}, can be considered when the RS ansatz does not hold. We refer the interested reader to \cite{mezard1987spin} for further discussion.\\

With the aid of the RS ansatz \eqref{eq:intro:RS}, one is now in a position to sequentially simplify the expressions of the potentials $\Psi_t, \Psi_w,\Psi_y$. 

\subsubsection{Trace potential}
To leading order in $s$, under the RS ansatz \eqref{eq:intro:RS}, the trace potential $\Psi_t$ can be compactly written as
\begin{align}
    \beta \Psi_t= -\sum\limits_\ell \sum\limits_{k}\left(\hat{m}_{\ell,k}^\top m_{\ell,k}+\Tr[\theta_{\ell,k}\hat{\theta}_{\ell,k}^\top-\frac{(V_{\ell,k}+q_{\ell,k})(\hat{V}_{\ell,k}-\hat{q}_{\ell,k})^\top}{2}-\frac{q_{\ell,k}\hat{q}_{\ell,k}^\top}{2}]\right)+\frac{1}{2}\Tr[v\hat{v}^\top],
\end{align}
where we introduced the additional order parameters
\begin{align}
    V_{\ell,k}\equiv r_{\ell,k}-q_{\ell,k},&&\hat{V}_{\ell,k}\equiv \hat{r}_{\ell,k}+\hat{q}_{\ell,k}.
\end{align}

\subsubsection{Entropic potential}

We now turn to the entropic potential $\Psi_w$, which can be expressed as
\begin{align}
    e^{\beta sd\Psi_w}
    &=\int \prod\limits_{a=1}^s d\w_a e^{ \sum\limits_{a=1}^s -\beta \frac{\lambda}{2}\lVert\w_a\lVert^2+\Tr[\hat{v}\w_a^\top\w_a]+\sum\limits_a \sum\limits_\ell\sum\limits_{k}(\sqrt{d}\hat{m}_{\ell,k }^\top\w_a^\top \mean_{\ell,k}+ \Tr[\hat{\theta}_{\ell,k} \teach^\top \Sigma_{\ell,k} \w_a])}\notag\\
    &\qquad\qquad\qquad e^{-\frac{1}{2}\sum\limits_{a}\sum\limits_\ell \sum\limits_{k}\Tr[\hat{V}_{\ell,k}\w_a^\top\Sigma_{\ell,k} \w_a]+\sum\limits_{1\le a, b\le s}\sum\limits_\ell \sum\limits_{k}\Tr[\hat{q}_{\ell,k}\w_b^\top\Sigma_{\ell,k} \w_a]}\notag\\
    &=\mathbb{E}_{\vec{\Xi}} \left[
    \int d\w e^{-\beta \frac{\lambda}{2}\lVert \w\lVert^2 -\frac{1}{2}\w^\top\odot\left[\hat{v}\otimes \mathbb{I}_d+\sum\limits_\ell\sum\limits_{k}\hat{V}_{\ell,k}\otimes \boldsymbol{\Sigma}_{\ell,k}\right]\odot \w^\top
    +\left(
    \sum\limits_\ell\sum\limits_{k}\sqrt{d}\hat{m}_{\ell,k} \mean_{\ell,k}^\top 
    +\hat{\theta}_{\ell,k} \teach^\top\Sigma_{\ell,k}+\boldsymbol{\Xi}_{\ell,k}\odot (\hat{q}_{\ell,k}\otimes \boldsymbol{\Sigma}_{\ell,k})^{\frac{1}{2}}
    \right)\odot \w^\top
    }
    \right]^s.
\end{align}
In going from the first to the second line, we employed a Hubbard-Stratonovitch transformation, introducing the Gaussian expectation $\mathbb{E}_{\vec{\Xi}}$ which bears over independent matrices $\vec{\Xi}_{\ell,k}\in \R^{r\times d}$ with i.i.d standard Gaussian entries. $\otimes$ denotes the direct (Kronecker) product of two matrices. For two matrices $\w,\w^\prime \in\mathbb{R}^{d \times r}$ and a block matrix $\boldsymbol{A}\in \mathbb{R}^{rd \times rd}$ (viewed as a block matrix with $d\times d$ blocks), we denoted $( \boldsymbol{A} \odot \w^\top)_{kl}=\sum_{ij}\w^{ji}\boldsymbol{A}_{ij,kl}$ and $\w^\top\odot (\w^\prime)^\top=\sum_{ij}\w^{ij}\w^\prime_{ij}$.
Carrying out the Gaussian integrals allows to reach the compact expression
\begin{align}
\label{eq:entro_final}
 \beta \Psi_w=&-\frac{1}{2d}\ln\det\left[\beta \lambda \mathbb{I}_r\otimes \mathbb{I}_d+\check{V}
    \right]\notag\\&
    +\frac{1}{2d}\Tr[(\beta \lambda \mathbb{I}_r\otimes \mathbb{I}_d+\check{V})^{-1} \left(\left(\sum\limits_\ell\sum\limits_{k}\sqrt{d}\hat{m}_{\ell,k} \mean_{\ell,k}^\top 
    +\hat{\theta}_{\ell,k} \teach^\top\Sigma_{\ell,k}\right)^{\otimes 2}+\sum\limits_\ell\sum\limits_{k}\hat{q}_{\ell,k}\otimes \boldsymbol{\Sigma}_{\ell,k}
    \right)],
\end{align}
where we used the shorthand
\begin{align}
    \check{V}\equiv
    \hat{v}\otimes \mathbb{I}_d+
    \sum\limits_\ell\sum\limits_{k} \hat{V}_{\ell,k}\otimes \Sigma_{\ell,k} .
\end{align}

\subsubsection{Energetic potential}

The computation of the energetic potential $\Psi_y$ requires more lengthy, albeit straightforward,  steps. The main technical hurdle is to evaluate the expectation over $h^\star_\ell, h^a_\ell$, under the RS ansatz \eqref{eq:intro:RS}. 
The Gaussian measure over $h^\star_\ell, h^a_\ell$ \eqref{eq:intro_overlaps} has, conditional on the class assignment $c$, the covariance
\begin{align}
\Sigma_{\ell,c_\ell}=\left(
    \begin{array}{cccc}
     \rho_{\ell,c_\ell} &\theta_{\ell,c_\ell}^\top &\dots & \theta_{\ell,c_\ell}^\top \\
      \theta_{\ell,c_\ell}   & r_{\ell,c_\ell} &q_{\ell, c_\ell}&\dots\\
      \vdots & q_{\ell, c_\ell}&\ddots&q_{\ell, c_\ell}\\
      \theta_{\ell, c_\ell}& \hdots&q_{\ell, c_\ell}&r_{\ell,c_\ell}
    \end{array}\right)
    \in\mathbb{R}^{(t+rs)\times (t+rs)}.
\end{align}
The Gaussian measure over $h^\star_\ell, h^a_\ell$ formally involves the inverse $\Sigma^{-1}_{\ell,c_\ell}$ and the determinant $\det \Sigma_{\ell,c_\ell}$, which we need to compute.
It can be seen that the inverse $\Sigma^{-1}_{\ell,c_\ell}$ exhibits a similar block structure, with corresponding elements \cite{aubin2018committee, cornacchia2023learning}
\begin{align}
\label{eq:inverse_elements}
    &\tilde{\rho}_{\ell,c_\ell}
    =\left(\rho_{\ell,c_\ell}- s\theta_{\ell,c_\ell}^\top (V_{\ell,c_\ell}+sq_{\ell,c_\ell})^{-1}\theta_{\ell,c_\ell} \right)^{-1},\notag\\
    &\Tilde{r}_{\ell,c_\ell}=V_{\ell,c_\ell}^{-1}+(V_{\ell,c_\ell}+sq_{\ell,c_\ell})^{-1}\left(
    \theta _{\ell,c_\ell}\tilde{\rho}_{\ell,c_\ell} \theta_{\ell,c_\ell}^\top (V_{\ell,c_\ell}+sq_{\ell,c_\ell})^{-1}-q_{\ell,c_\ell}V_{\ell,c_\ell}^{-1}
    \right),\notag\\
    &\tilde{q}_{\ell,c_\ell}=\Tilde{r}_{\ell,c_\ell}-V_{\ell,c_\ell}^{-1},\notag\\
    &\Tilde{\theta}_{\ell,c_\ell}=-(V_{\ell,c_\ell}+sq_{\ell,c_\ell})^{-1}\theta_{\ell,c_\ell} \Tilde{\rho}_{\ell,c_\ell},
\end{align}
while the determinant of $\Sigma_{\ell,c_\ell}$ can be evaluated as \cite{aubin2018committee, cornacchia2023learning}
\begin{align}
\label{eq:determinants}
    \ln \det \Sigma _{\ell,c_\ell}
    =(s-1) \ln \det V_{\ell,c_\ell}+\ln\det(V_{\ell,c_\ell}+sq_{\ell,c_\ell})+\ln\det(\rho_{\ell,c_\ell}- s\theta_{\ell,c_\ell}^\top (V_{\ell,c_\ell}+sq_{\ell,c_\ell})^{-1}\theta_{\ell,c_\ell} ).
\end{align}
Let us note that the expressions for the inverse \eqref{eq:inverse_elements} and the determinant of \eqref{eq:determinants} of $\Sigma_{\ell,c_\ell}$ can be derived from standard matrix identities, for integer $s$. In the framework of the replica method, they are assumed to hold also when considering the non-integer sized matrix $\Sigma_{\ell,c_\ell}$ as $s\to 0$. Like the RS ansatz, the expression \eqref{eq:inverse_elements} \eqref{eq:determinants} thus also constitute formally ambiguous, yet physically intuitive, heuristic steps.
One is now in a position to evaluate the energetic potential, plugging in the RS ansatz \eqref{eq:intro:RS}. Defining $\tilde{V}\equiv \tilde{r}-\Tilde{q}$, one reaches
\begin{align}
    e^{s\beta\Psi_y}&=\mathbb{E}_c\int dh^\star\prod\limits_a^s dh_a\prod\limits_\ell e^{-\frac{1}{2}h^{\star\top}_\ell \tilde{\rho}_{\ell,c_\ell}h^\star_\ell -(h^\star_\ell) ^\top\Tilde{\theta}_{\ell,c_\ell}^\top\sum\limits_a h^a_\ell
    -\frac{1}{2}\sum\limits_a h^{a\top}_{\ell} \tilde{V}_{\ell,c_\ell} h^{a}_{\ell}-\frac{1}{2}\sum\limits_{a,b}h^{a\top}_{\ell} \Tilde{q}_{\ell, c_\ell}h^b_\ell
    -\frac{1}{2}\ln\det 2\pi \Sigma_{\ell,c_\ell}
    }\notag\\
    &\qquad\qquad\qquad\qquad\qquad e^{-\beta\sum\limits_{a=1}^s  \ell\left(h_\star+m_\star^c, h^a+m^c_a,v,c\right)} \notag\\
    & =\mathbb{E}_c\mathbb{E}_\Xi
    \int dh^\star \prod\limits_\ell e^{-\frac{1}{2}h^{\star\top }_\ell\tilde{\rho}_{\ell,c_\ell}h^\star_\ell
     -\frac{1}{2}\ln\det  2\pi\Sigma_{\ell,c_\ell}-\frac{s}{2}\ln\det 2\pi \tilde
V_{\ell,c_\ell}+\frac{s}{2}(\Tilde{\theta}_{\ell,c_\ell}h^\star_\ell -(-\Tilde{q}_{\ell, c_\ell})^{\sfrac{1}{2}}\xi_\ell) ^\top \Tilde{V}_{\ell,c_\ell} ^{-1} 
 (\Tilde{\theta}_{\ell,c_\ell}h^\star_\ell -(-\Tilde{q}_{\ell, c_\ell})^{\sfrac{1}{2}}\xi_\ell) }\notag\\
     &\qquad\qquad\qquad
     \Bigg[
     \int dh \prod \limits_\ell \frac{e^{-
     \frac{1}{2}((h_\ell-\Tilde{V}_{\ell,c_\ell} ^{-1}(\Tilde{\theta}_{\ell,c_\ell}h^\star _\ell-(-\Tilde{q}_{\ell, c_\ell})^{\sfrac{1}{2}}\xi_\ell))^\top \Tilde{V}_{\ell,c_\ell} ((h_\ell-\Tilde{V}_{\ell,c_\ell} ^{-1}(\Tilde{\theta}_{\ell,c_\ell}h^\star_\ell -(-\Tilde{q}_{\ell, c_\ell})^{\sfrac{1}{2}}\xi_\ell))}}{\det (2\pi\Tilde{V}_{\ell,c_\ell} ^{-1})^{\sfrac{1}{2}} }\notag\\
     &\qquad\qquad\qquad e^{-\beta  \ell\left(h_\star+m_\star^c, h+m^c,v,c\right)}
     \Bigg]^s\notag\\
     &=1+s\mathbb{E}_c\mathbb{E}_\Xi
    \int dh^\star \prod\limits_\ell\frac{e^{-\frac{1}{2}h^{\star\top}_\ell \rho_{\ell,c_\ell}^{-1}h^\star_\ell}}{\det(2\pi\rho_{\ell,c_\ell})^{\sfrac{1}{2}}}
    \notag\\
    &\qquad\qquad\qquad\ln\Bigg[    
     \int dh \prod \limits_\ell \frac{e^{-
     \frac{1}{2}((h-V_{\ell,c_\ell} (\Tilde{\theta}^0_{\ell,c_\ell}h^\star -(-\Tilde{q}^0_{\ell, c_\ell})^{\sfrac{1}{2}}\xi_\ell))^\top V_{\ell,c_\ell}^{-1} ((h-V_{\ell,c_\ell}(\Tilde{\theta}^0_{\ell,c_\ell}h^\star  -(-\Tilde{q}^0_{\ell, c_\ell})^{\sfrac{1}{2}}\xi_\ell))}}{\det (2\pi V_{\ell,c_\ell})^{\sfrac{1}{2}}}\notag\\
     &\qquad\qquad\qquad e^{-\beta  \ell\left(h_\star+m_\star^c, h+m^c,v,c\right)}
    \Bigg]+o_s(s)
\end{align}
We have again introduced a Hubbard-Stratonovitch tranformation in going from the first line to the second, introducing $\Xi\in\R^{L\times r}$ with rows $\xi_\ell\sim\mathcal{N}(0,\mathbb{I}_r)$. The third line results from a Taylor expansion to first order in $s$, leveraging in particular the expansions \eqref{eq:determinants} and \eqref{eq:inverse_elements}. The superscript $~^0$ indicates the zeroth order ($s=0$) expression of the variables \eqref{eq:inverse_elements}. 
Doing a change of variables on the Gaussian variables $h_\star, \Xi$, and renaming the integration variables for clarity and ease of bookkeeping finally leads to the compact expression
\begin{align}
    \label{eq:energetic_final}\beta\Psi_y&=\underbrace{\mathbb{E}_c\int_{\mathbb{R}^{L\times t}} \!\!\!\!\!\! dY  \mathbb{E}_\Xi \prod\limits_{\ell=1}^L\frac{e^{-\frac{1}{2}\left(y_\ell-\theta_{\ell,c_\ell}^\top q_{\ell,c_\ell}^{-\frac{1}{2}}\xi_\ell\right)^\top(\rho_{\ell,c_\ell}- \theta_{\ell,c_\ell}^\top q_{\ell,c_\ell}^{-1}\theta_{\ell,c_\ell} )^{-1}\left(y_\ell-\theta_{\ell,c_\ell}^\top q_{\ell,c_\ell}^{-\frac{1}{2}}\xi_\ell\right)}}{\det(2\pi (\rho_{\ell,c_\ell}- \theta_{\ell,c_\ell}^\top q_{\ell,c_\ell}^{-1}\theta_{\ell,c_\ell}))^{\sfrac{1}{2}}}}_{\equiv\mathbb{E}_{c,Y,\Xi}}\notag\\
    &\times \ln\left[
    \int_{\mathbb{R}^{L\times r}} \!\!\!\!\!\!dX
    \prod\limits_{\ell=1}^L \frac{e^{-\frac{1}{2}\left( x_\ell-q_{\ell,c_\ell}^{\frac{1}{2}}\xi_\ell\right)^\top V_{\ell,c_\ell}^{-1}\left( x_\ell-q_{\ell,c_\ell}^{\frac{1}{2}}\xi_\ell\right)}}{\det(2\pi V_{\ell,c_\ell})^{\sfrac{1}{2}}}e^{-\beta \ell\left(
    Y+m^c_\star, X+m^c,v,c
    \right)}
    \right].
\end{align}
The expectation again bears over a tensor $\vec{\Xi}\in \R^{L\times r}$ with i.i.d standard Gaussian entries.\\

Summarizing, we have computed all the three terms $\Psi_t,\Psi_w,\Psi_y$ making up the replicated partition function $Z^s$ \eqref{eq:Zs_p2}, as a function of the RS order parameters $\{q_{\ell,k}, V_{\ell,k}, m_{\ell,k},\theta_{\ell,k},t,\hat{q}_{\ell,k},\hat{V}_{\ell,k},\hat{m}_{\ell,k},\hat{\theta}_{\ell,k},\hat{v}\}_{\ell,k}$ \eqref{eq:intro:RS}. Before carrying out the extremization over the latter, one can first deal with the zero-temperature $\beta \to\infty$ limit.

\subsubsection{Zero-temperature limit}
To carry out the $\beta\to\infty$ 
 limit, it is convenient, following \cite{aubin2020generalization}, to first rescale the RS order parameters \eqref{eq:intro:RS} as 
\begin{align}
    \label{eq:temperature_rescaling}
    \beta \hat{V}_{\ell,k}\leftarrow \hat{V}_{\ell,k},&& \frac{1}{\beta}V_{\ell,k} \leftarrow V_{\ell,k}, && \beta\hat{m}_{\ell,k}\leftarrow \hat{m}_{\ell,k}
    , && \beta\hat{\theta}_{\ell,k}\leftarrow \hat{\theta}_{\ell,k},&& \beta^2\hat{q}_{\ell,k}\leftarrow \hat{q}_{\ell,k}, && \beta \hat{v}\leftarrow \hat{v}.
\end{align} 
In the zero-temperature limit $\beta\to\infty$, the entropic potential $\Psi_w$ \eqref{eq:entro_final} straightforwardly reduces to 
\begin{align}
    \label{eq:entro_zeroT}
    \Psi_w=&\frac{1}{2d}\Tr[( \lambda \mathbb{I}_r\otimes \mathbb{I}_d+\check{V})^{-1} \left(\left(\sum\limits_\ell\sum\limits_{k}\sqrt{d}\hat{m}_{\ell,k} \mean_{\ell,k}^\top 
    +\hat{\theta}_{\ell,k} \teach^\top\Sigma_{\ell,k}\right)^{\otimes 2}+\sum\limits_\ell\sum\limits_{k}\hat{q}_{\ell,k}\otimes \boldsymbol{\Sigma}_{\ell,k}
    \right)].
\end{align}
Using the spectral density $\nu$ \eqref{eq:spectral_density} introduced in section \ref{sec:seq-GLM}, $\Psi_w$ can further be written in fully asymptotic form as
\begin{align}
    \Psi_w=&\frac{1}{2}\int d\nu(\gamma, \tau,\pi) \Tr[\left(\lambda \mathbb{I}_r+\hat{v}+\sum\limits_{\ell}\sum\limits_{k} \gamma_{\ell,k}\hat{V}_{\ell, k}\right)^{-1} \left(\left(\sum\limits_\ell\sum\limits_{k}\hat{m}_{\ell,k} \tau_{\ell,k}
    +\gamma_{\ell,k}\hat{\theta}_{\ell,k} \pi\right)^{\otimes 2}\!\!\!\!\!\!\!\!+\sum\limits_\ell\sum\limits_{k}\gamma_{\ell,k}\hat{q}_{\ell,k}
    \right)].
\end{align}
The energetic potential $\Psi_y$ \eqref{eq:energetic_final} is similarly greatly simplified in the $\beta\to\infty$ limit, and can be recast into a more compact form
\begin{align}
    \label{eq:energetic_zeroT}
    \Psi_y=-\mathbb{E}_{c,Y,\Xi}\mathcal{M}(c,Y,\Xi),
\end{align}
where the Moreau envelope is defined as
\begin{align}
    \label{eq:Moreau_repeat}
    &\mathcal{M}(c,Y, \Xi)=\underset{X}{\inf}\Bigg\{
    \frac{1}{2}\sum\limits_{\ell=1}^L \Tr[V_{\ell,c_\ell}^{-1}\left(X_\ell-q_{\ell,c_\ell}^{\sfrac{1}{2}}\xi_\ell-m_{\ell,c_\ell}\right)^{\otimes 2}]+\ell\left(
    Y+m_\star^c, X,v,c
    \right)
    \Bigg\}.
\end{align}
Note that the optimization problem associated to the evaluation of the Moreau envelope \eqref{eq:Moreau_repeat} bears over a finite-dimensional argument $X\in\R^{L\times r}$, and in thus \textit{low-dimensional} -- in contrast to the original high-dimensional ERM problem \eqref{eq:ERM}. On an intuitive plane, the Moreau envelope \eqref{eq:Moreau_repeat} represents an effective, average empirical loss landscape, in a typical realization of the training set.

\subsubsection{Replica free energy}
Summarizing, we have reached a closed-form expression for each of the components $\Psi_t,\Psi_w, \Psi_y$ composing the average replicated partition function $\mathbb{E}_{\mathcal{D}}Z^s$. Putting these characterization together with the replica trick \eqref{eq:replica_trick} allows one to finally reach the characterization of the free energy \eqref{eq:free_energy} (or cumulant-generating function)
\begin{align}
    f=-\underset{\beta\to\infty}{\lim}\frac{1}{\beta d}\mathbb{E}_{\mathcal{D}} \ln Z=-\Psi_t-\Psi_w-\alpha \Psi_y.
\end{align}
As discussed above, $f$ is expressed as the solution of a low-dimensional optimization problem
\begin{align}
    f=-\mathrm{extr}~~ \Phi
\end{align}
where the extremization bears on the variables $q_{\ell,k}, V_{\ell,k}, m_{\ell,k},\theta_{\ell,k},t,\hat{q}_{\ell,k},\hat{V}_{\ell,k},\hat{m}_{\ell,k},\hat{\theta}_{\ell,k},\hat{v}$ and the \textit{free entropy} $\Phi$ reads
\begin{align}
    \label{eq:intro:Phi_l2_fullasymp}    \Phi=&\sum\limits_\ell\sum\limits_{k}\left(\frac{1}{2}\Tr[q_{\ell,k}\hat{V}_{\ell,k}^\top - V_{\ell,k}\hat{q}_{\ell,k}^\top] -\Tr[ \theta_{\ell,k}\hat{\theta}_{\ell,k}^\top]- \hat{m}_{\ell,k}^\top m_{\ell,k}\right) +\frac{1}{2}\Tr[v\hat{v}^\top]-\alpha\mathbb{E}_{c,Y,\Xi}\mathcal{M}(c,Y,\Xi)\\
    &+\frac{1}{2}\int d\nu(\gamma, \tau,\pi) \Tr[\left(\lambda \mathbb{I}_r+\hat{v}+\sum\limits_{\ell}\sum\limits_{k} \gamma_{\ell,k}\hat{V}_{\ell, k}\right)^{-1} \left(\left(\sum\limits_\ell\sum\limits_{k}\hat{m}_{\ell,k} \tau_{\ell,k}
    +\gamma_{\ell,k}\hat{\theta}_{\ell,k} \pi\right)^{\otimes 2}\!\!\!\!\!\!\!\!+\sum\limits_\ell\sum\limits_{k}\gamma_{\ell,k}\hat{q}_{\ell,k}
    \right)].
\end{align}
Assuming all terms are differentiable, the extremization over $q_{\ell,k}, V_{\ell,k}, m_{\ell,k},\theta_{\ell,k},t,\hat{q}_{\ell,k},\hat{V}_{\ell,k},\hat{m}_{\ell,k},\hat{\theta}_{\ell,k},\hat{v}$ can be enforced by writing zero-gradient conditions on these variables. These conditions are commonly referred to as \textit{Saddle-Point} (SP) equations, and read 
\begin{align}
    \label{eq:intro:replica_SP_repeat}
    &\begin{cases}
        \hat{q}_{\ell, k}=\alpha\mathbb{E}_{c}\delta_{c_\ell,k}\mathbb{E}_{\Xi,Y} V_{\ell,k}^{-1}\left(
        \prox_\ell^c-q_{\ell,k}^{\frac{1}{2}}\xi_\ell-m_{\ell,k}
        \right)^{\otimes 2}V_{\ell,k}^{-1}\\
        \hat{V}_{\ell,k}=\hat{\theta}_{\ell,k}\theta_{\ell,k}^\top q_{\ell,k}^{-1}-\alpha \mathbb{E}_{c}\delta_{c_\ell,k}\mathbb{E}_{\Xi,Y} V_{\ell,k}^{-1}\left(
        \prox_\ell^c-q_{\ell,k}^{\frac{1}{2}}\xi_\ell-m_{\ell,k}
        \right)\xi_\ell^\top q_{\ell,k}^{-\frac{1}{2}}\\
        \hat{m}_{\ell, k}=\alpha \mathbb{E}_{c}\delta_{c_\ell,k}\mathbb{E}_{\Xi,Y}V_{\ell, k}^{-1}\left(
        \prox_\ell^c-q_{\ell,k}^{\frac{1}{2}}\xi_\ell-m_{\ell,k}
        \right)\\
        \hat{\theta}_{\ell,k}=\alpha \mathbb{E}_{c}\delta_{c_\ell,k}\mathbb{E}_{\Xi,Y}V_{\ell,k}^{-1}\left(
        \prox_\ell^c-q_{\ell,k}^{\frac{1}{2}}\xi_\ell-m_{\ell,k}
        \right)\left(y_\ell-\theta_{\ell,k}^\top q_{\ell,k}^{-\sfrac{1}{2}} \xi_\ell \right)^\top\left(\rho_{\ell,k}-\theta_{\ell,k}^\top q_{\ell,k}^{-1}\theta_{\ell,k}\right)^{-1}\\
        \hat{v}=2\alpha\mathbb{E}_c\mathbb{E}_{\Xi,Y}\partial_3 \ell(Y+m^c_\star, \prox^c,v,c)
    \end{cases}
\end{align}
\begin{align}
    &\begin{cases}
        q_{\ell,k}=\int d\nu(\gamma,\tau,\pi) 
        \gamma_{\ell,k}\left(
        \lambda\mathbb{I}_r+\hat{v}+\sum\limits_{\kappa}\sum\limits_{j} \gamma_{\kappa,j}\hat{V}_{\kappa, j}
        \right)^{-1}
        \\\qquad\left[\left(\sum\limits_\kappa\sum\limits_{j}\hat{m}_{\kappa,j} \tau_{\kappa,j}
    +\gamma_{\kappa,j}\hat{\theta}_{\kappa,j} \pi\right)^{\otimes 2}\!\!\!\!\!\!\!\!+\sum\limits_\kappa\sum\limits_{j}\gamma_{\kappa,j}\hat{q}_{\kappa,j}\right]\left(
        \lambda\mathbb{I}_r+\hat{v}+\sum\limits_{\kappa}\sum\limits_{j} \gamma_{\kappa,j}\hat{V}_{\kappa, k}
        \right)^{-1}
        \\
        V_{\ell,k}=\int d\nu(\gamma,\tau,\pi) 
        \gamma_{\ell,k}\left(
        \lambda\mathbb{I}_r+\hat{v}+\sum\limits_{\kappa}\sum\limits_{j} \gamma_{\kappa,j}\hat{V}_{\kappa, j}
        \right)^{-1}\\
        m_{\ell,k}=\int d\nu(\gamma,\tau,\pi) 
        \tau_{\ell,k}\left(\lambda\mathbb{I}_r+\hat{v}+\sum\limits_{\kappa}\sum\limits_{j} \gamma_{\kappa,j}\hat{V}_{\kappa, j}
        \right)^{-1}\left(\sum\limits_\kappa\sum\limits_{j}\hat{m}_{\kappa,j} \tau_{\kappa,j}
    +\gamma_{\kappa,j}\hat{\theta}_{\kappa,j} \pi
        \right) \\
        \theta_{\ell,k}=\int d\nu(\gamma,\tau,\pi) 
        \gamma_{\ell,k}\left(\lambda\mathbb{I}_r+\hat{v}+\sum\limits_{\kappa}\sum\limits_{j} \gamma_{\kappa,j}\hat{V}_{\kappa, j}
        \right)^{-1} \left(\sum\limits_\kappa\sum\limits_{j}\hat{m}_{\kappa,j} \tau_{\kappa,j}
    +\gamma_{\kappa,j}\hat{\theta}_{\kappa,j} \pi
        \right)\pi^\top\\
        v=\int d\nu(\gamma,\tau,\pi) 
       \left(
        \lambda\mathbb{I}_r+\hat{v}+\sum\limits_{\kappa}\sum\limits_{j} \gamma_{\kappa,j}\hat{V}_{\kappa, j}
        \right)^{-1}
        \\\qquad\left[\left(\sum\limits_\kappa\sum\limits_{j}\hat{m}_{\kappa,j} \tau_{\kappa,j}
    +\gamma_{\kappa,j}\hat{\theta}_{\kappa,j} \pi\right)^{\otimes 2}\!\!\!\!\!\!\!\!+\sum\limits_\kappa\sum\limits_{j}\gamma_{\kappa,j}\hat{q}_{\kappa,j}\right]\left(
        \lambda\mathbb{I}_r+\hat{v}+\sum\limits_{\kappa}\sum\limits_{j} \gamma_{\kappa,j}\hat{V}_{\kappa, k}
        \right)^{-1}
    \end{cases}
\end{align}
where $\prox_\ell^c$ corresponds to the arginf of the minimization defining the Moreau envelope $\mathcal{M}(c,Y,\Xi)$ \eqref{eq:Moreau_repeat}, namely 
\begin{align}
    \label{eq:prox}
    &\prox^c=\underset{X}{\mathrm{arginf}}\Bigg\{
    \frac{1}{2}\sum\limits_{\ell=1}^L \Tr[V_{\ell,c_\ell}^{-1}\left(X_\ell-q_{\ell,c_\ell}^{\sfrac{1}{2}}\xi_\ell-m_{\ell,c_\ell}\right)^{\otimes 2}]+\ell\left(
    Y+m_\star^c, X,v,c
    \right)
    \Bigg\}.
\end{align}
How to interpret the extremizers $q_{\ell,k},m_{\ell,k},\theta_{\ell,k}$, solutions of \eqref{eq:intro:replica_SP_repeat}? Consider the summary statistics
\begin{align}
\label{eq:summary_stats_intro}\check{q}_{\ell,k}(\hat{\w})=\frac{\hat{\w}^\top\Sigma_{\ell,k} \hat{\w}}{d},&& \check{m}_{\ell,k}(\hat{\w})=\frac{\mean_{\ell,k}^\top \hat{\w}}{\sqrt{d}}, && \check{\theta}_{\ell,k}(\hat{\w})=\frac{\hat{\w}^\top \Sigma_{\ell,k} \teach}{d},
\end{align}
of the trained weights $\hat{\w}$. The average value of any test function $\phi(\hat{\w})\equiv\phi(\{\check{q}_{\ell,k}(\hat{\w}),\check{m}_{\ell,k}(\hat{\w}), \check{\theta}_{\ell,k}(\hat{\w})\}_{\ell,k}) $ of these statistics can be rewritten similarly to \eqref{eq:intro:test_func} as
\begin{align}
    \mathbb{E}_{\mathcal{D}}\phi(\hat{\w})&=\underset{\beta\to\infty}{\lim}\mathbb{E}_{\mathcal{D}} \frac{1}{Z}\int d\w e^{-\beta \mathcal{R}(\w)}\phi(\w)\notag\\
    &=\underset{\beta\to\infty, s\to 0}{\lim}\mathbb{E}_{\mathcal{D}}
    \int d\w e^{-\beta \mathcal{R}(\w)}\phi(\w)Z^{s-1}\notag\\
    &=\underset{\beta\to\infty, s\to 0}{\lim} \mathbb{E}_{\mathcal{D}}\int \prod\limits_{a=1}^s d\w^a e^{-\beta \mathcal{R}(\w^a)}\phi(\w^1)\notag\\
    &=\underset{\beta\to\infty, s\to 0}{\lim}\int \prod\limits_{\ell=1}^L\prod\limits_{k=1}^{K_\ell}\prod\limits_{a=1}^sdm^{\ell, k}_a d\hat{m}^{\ell, k}_a d\theta_a^{\ell,k} d\hat{\theta}_a^{\ell,k}\prod\limits_{a\le b}^s dq^{\ell,k}_{ab}d\hat{q}^{\ell,k}_{ab} \phi(\{q^{\ell,k}_{11}, m^{\ell,k}_1, \theta^{\ell,k}_1\}_{\ell,k})e^{sd\beta (\Psi_t+\Psi_w+\Psi_y)}\notag\\
    &\asymp \underset{\beta\to\infty, s\to 0}{\lim}\phi(\{q_{\ell,k}, m_{\ell,k}, \theta_{\ell,k}\}_{\ell,k})e^{-s\beta d f}=\phi(\{q_{\ell,k}, m_{\ell,k}, \theta_{\ell,k}\}_{\ell,k}),
\end{align}
where the last equality results from first taking the $s\to 0$ limit. In words, the average of any function of the summary statistics $\check{q}_{\ell,k}(\hat{\w}),\check{m}_{\ell,k}(\hat{\w}), \check{\theta}_{\ell,k}(\hat{\w})$ \eqref{eq:summary_stats_intro}, and in particular the average of the summary statistics  themselves, is asymptotically given by (a function of) the quantities $q_{\ell,k}, m_{\ell,k},\theta_{\ell,k}$ characterized by the SP equations \eqref{eq:intro:replica_SP_repeat}. The replica method thus provides a powerful framework to \textit{tightly} describe the minimizer $\hat{\w}$ of the ERM problem \eqref{eq:ERM}, in terms of a set of summary statistics. In practice, the SP equations \eqref{eq:intro:replica_SP_repeat} characterizing the summary statistics can be solved numerically, for instance by iterating the equations until a fixed point is reached. A possible numerical bottleneck lies in the evaluation of the expectation over the Gaussian variables $\Xi, Y$ in \eqref{eq:intro:replica_SP_repeat}, which require the computation of finite but multi-dimensional integrals over $L(t+r)$ variables -- warranting, in the most intricate cases, efficient numerical integration schemes.

\subsubsection{Test error}
As discussed, any metric depending on the summary statistics characterized by the SP equations \eqref{eq:intro:replica_SP_repeat} can thus be sharply characterized in closed-form. One such metric of particular interest is the generalization error $\epsilon_g$ \eqref{eq:eg} associated to the trained \seq:
\begin{align}
    \label{eq:test_mse}
   \epsilon_g=\mathbb{E}_{\x}\ell_{\mathrm{tst.}}\left(
\frac{\x \teach}{\sqrt{d}},\frac{\x \hat{\w}}{\sqrt{d}}, \frac{\hat{\w}^\top \hat{\w}}{d},c
\right).
\end{align}
This expression is a function of the correlated Gaussian variables $\x\hat{\w}, \x\teach$, whose joint second-order statistics are simply given by the summary statistics $q_{\ell,k}, \theta_{\ell,k},\rho_{\ell,k}, m^\star_{\ell,k}, m_{\ell,k}$, which are in turn characterized by the SP equations \eqref{eq:intro:replica_SP_repeat}. This seamlessly implies the following compact asymptotic characterization:
\begin{align}
    \label{eq:eg_repl}
    \epsilon_g=\mathbb{E}_{c,X,Y}\ell_{\mathrm{ts.}}\left(Y,X,v,c\right),
\end{align}
where, conditioned on the class assignments $c$, the average bears on $X\in\mathbb{R}^{L\times r}, Y\in\mathbb{R}^{L\times t}$ with independent rows with statistics
\begin{align}
    (x_\ell, y_\ell)\sim \mathcal{N}\left[
    \begin{pmatrix}
        m_{\ell,c_\ell}\\\hline
        m^\star_{\ell,c_\ell} 
    \end{pmatrix},
    \left(
    \begin{array}{c|c}
         q_{\ell,c_\ell}& \theta_{\ell,c_\ell} \\\hline
         \theta_{\ell,c_\ell}^\top& \rho_{\ell,c_\ell}
    \end{array}\right)
    \right],
\end{align}
where the summary statistics $q_{\ell,c_\ell}, \theta_{\ell,c_\ell},\rho_{\ell,c_\ell}$ are characterized by \eqref{eq:intro:replica_SP_repeat}.

\subsubsection{Training loss}
We finally turn to the other metric of interest, namely the training loss $\epsilon_t$ \eqref{eq:et}. Unlike the test error $\epsilon_g$ \eqref{eq:eg}, the training loss cannot be straightforwardly expressed in terms of the summary statistics $q_{\ell,c_\ell}, \theta_{\ell,c_\ell},\rho_{\ell,c_\ell}$\eqref{eq:intro:replica_SP_repeat}, and some additional steps are required to reach such an expression. Before presenting the derivation, first observe that it is reasonable to expect, from statistical physics, that the training loss (energy) should coincide with the free energy $f$ at zero temperature $\beta \to \infty$. We provide below an alternative derivation. First note that the training loss $\epsilon_t$ can be expressed as
\begin{align}
    \epsilon_t=-\underset{\beta \to\infty}{\lim}\partial_\beta \underbrace{ \frac{1}{d}\ln Z(\beta)}_{\Phi(\beta)}=-\underset{\beta \to\infty}{\lim}\partial_\beta \left[
    \Psi_w(\beta)+\Psi_t(\beta)+\Psi_y(\beta)
    \right],
\end{align}
where $\Phi(\beta)$ is the free entropy at finite temperature, and we emphasized the $\beta-$ dependence of the potentials $\Psi_t,\Psi_w,\Psi_y$. The trace potential $\Psi_t$ bears no explicit dependence on $\beta$. On the other hand, we remind that the entropic potential reads 
\begin{align}
    \beta \Psi_w=&-\frac{1}{2d}\ln\det\left[\beta \lambda \mathbb{I}_r\otimes \mathbb{I}_d+\check{V}
    \right]\notag\\&
    +\frac{1}{2d}\Tr[(\beta \lambda \mathbb{I}_r\otimes \mathbb{I}_d+\check{V})^{-1}\odot \left(\left(\sum\limits_\ell\sum\limits_{k}\sqrt{d}\hat{m}_{\ell,k} \mean_{\ell,k}^\top 
    +\hat{\theta}_{\ell,k} \teach^\top\Sigma_{\ell,k}\right)^{\otimes 2}+\sum\limits_\ell\sum\limits_{k}\hat{q}_{\ell,k}\otimes \boldsymbol{\Sigma}_{\ell,k}
    \right)],
\end{align}
and thus explicitly depends on $\beta$. Its derivative is straightforwardly given by 
\begin{align}
\partial_\beta(\beta\Psi_w)=&- \frac{\lambda}{2d}\Tr\left[
    \beta \lambda\mathbb{I}_r\otimes \mathbb{I}_d +\check{V}
    \right]^{-1}\notag\\
    &-\frac{\lambda}{2d} \Tr[\left(\beta\lambda \mathbb{I}_r\odot \mathbb{I}_d +\check{V}\right)^{-2}\!\!\!\!\odot \left(\left(\sum\limits_\ell\sum\limits_{k}\sqrt{d}\hat{m}_{\ell,k} \mean_{\ell,k}^\top 
    +\hat{\theta}_{\ell,k} \teach^\top\Sigma_{\ell,k}\right)^{\otimes 2}+\sum\limits_\ell\sum\limits_{k}\hat{q}_{\ell,k}\otimes \boldsymbol{\Sigma}_{\ell,k}
    \right)].
\end{align}
Finally, going through the same rescaling steps to take the $\beta\to\infty $ limit,
\begin{align}
    \underset{\beta\to\infty}{\lim}\partial_\beta(\beta\Psi_w)=\!\!-\frac{\lambda}{2d} \!\!\Tr[\left(\lambda \mathbb{I}_r\odot \mathbb{I}_d +\check{V}\right)^{-2}\!\!\!\!\!\!\odot \left(\left(\sum\limits_\ell\sum\limits_{k}\sqrt{d}\hat{m}_{\ell,k} \mean_{\ell,k}^\top \!\!
    +\hat{\theta}_{\ell,k} \teach^\top\Sigma_{\ell,k}\right)^{\otimes 2}\!\!\!\!\!\!\!\!\!\!+\sum\limits_\ell\sum\limits_{k}\hat{q}_{\ell,k}\otimes \boldsymbol{\Sigma}_{\ell,k}
    \right)].
\end{align}
By the same token, it is straightforward to see that
\begin{align}
    \underset{\beta\to\infty}{\lim}\partial_\beta(\alpha \beta\Psi_y)&=-\alpha \mathbb{E}_{c,Y,\Xi}\left[
    \mathcal{M}(c,Y,\Xi)-\frac{1}{2}\sum\limits_{\ell=1}^L \Tr[V_{\ell,c_\ell}^{-1}\left(\prox^c_\ell-q_{\ell,c_\ell}^{\sfrac{1}{2}}\xi_\ell-m_{\ell,c_\ell}\right)^{\otimes 2}]
    \right]\notag\\
    &=-\alpha \mathbb{E}_{c,Y,\Xi}
    \mathcal{M}(c,Y,\Xi) +\frac{1}{2}\sum\limits_{\ell=1}^L\sum\limits_{k=1}^{K_\ell}\Tr[\hat{q}_{\ell,k}V_{\ell, k}].
\end{align}
We used the self-consistent equations \eqref{eq:intro:replica_SP_repeat} in the last line. Putting everything together, 
\begin{align}
    -\epsilon_t&=\underset{\beta \to\infty}{\lim}\partial_\beta \Psi(\beta)\notag\\
    &=-\frac{\lambda}{2} \!\!\int d\nu(\gamma,\tau,\pi) \Tr[
       \left(
        \lambda\mathbb{I}_r+\hat{v}+\sum\limits_{\kappa}\sum\limits_{k} \gamma_{\kappa,k}\hat{V}_{\kappa, k}
        \right)^{-2}\left[\left(\sum\limits_\kappa\sum\limits_{k}\hat{m}_{\kappa,k} \tau_{\kappa,k}
    +\gamma_{\kappa,k}\hat{\theta}_{\kappa,k} \pi\right)^{\otimes 2}\!\!\!\!\!\!\!\!+\sum\limits_\kappa\sum\limits_{k}\gamma_{\kappa,k}\hat{q}_{\kappa,k}\right]]\notag\\
    &\qquad -\alpha \mathbb{E}_{c,Y,\Xi}
    \mathcal{M}(c,Y,\Xi) +\frac{1}{2}\sum\limits_{\ell=1}^L\sum\limits_{k=1}^{K_\ell}\Tr[\hat{q}_{\ell,k}V_{\ell, k}].
\end{align}
This constitutes a sharp asymptotic characterization of the training loss $\epsilon_t$. For completeness, we finally explicit the connection between $\epsilon_t$ and the negative free entropy (i.e. the \textit{free energy} in statistical physics), thus validating the physical intuition that the two quantities should coincide. We go back to massage the expression for the free entropy \eqref{eq:intro:Phi_l2_fullasymp}
\begin{align}
    \Phi&=\sum\limits_\ell\sum\limits_{k}\left(\frac{1}{2}\Tr[q_{\ell,k}\hat{V}_{\ell,k}^\top - V_{\ell,k}\hat{q}_{\ell,k}^\top] -\Tr[ \theta_{\ell,k}\hat{\theta}_{\ell,k}^\top]- \hat{m}_{\ell,k}^\top m_{\ell,k}\right)+\frac{1}{2}\Tr[v\hat{v}]-\alpha\mathbb{E}_{c,Y,\Xi}\mathcal{M}(c,Y,\Xi)\\
    &+\frac{1}{2}\int d\nu(\gamma,\tau,\pi) \Tr[
       \left(
        \lambda\mathbb{I}_r+\hat{v}+\sum\limits_{\ell}\sum\limits_{k} \gamma_{\ell,k}\hat{V}_{\ell, k}
        \right)^{-1}\left[\left(\sum\limits_\ell\sum\limits_{k}\hat{m}_{\ell,k} \tau_{\ell,k}
    +\gamma_{\ell,k}\hat{\theta}_{\ell,k} \pi\right)^{\otimes 2}\!\!\!\!\!\!\!\!+\sum\limits_\ell\sum\limits_{k}\gamma_{\ell,k}\hat{q}_{\ell,k}\right]]\notag\\
    &=\sum\limits_\ell\sum\limits_{k}\frac{1}{2}\Tr[q_{\ell,k}\hat{V}_{\ell,k}^\top + V_{\ell,k}\hat{q}_{\ell,k}^\top]+\frac{1}{2}\Tr[v\hat{v}]-\alpha\mathbb{E}_{c,Y,\Xi}\mathcal{M}(c,Y,\Xi)\notag\\
    &-\frac{1}{2}\int d\nu(\gamma,\tau,\pi) \Tr[
       \left(
        \lambda\mathbb{I}_r+\hat{v}+\sum\limits_{\ell}\sum\limits_{k} \gamma_{\ell,k}\hat{V}_{\ell, k}
        \right)^{-1}\left[\left(\sum\limits_\ell\sum\limits_{k}\hat{m}_{\ell,k} \tau_{\ell,k}
    +\gamma_{\ell,k}\hat{\theta}_{\ell,k} \pi\right)^{\otimes 2}\!\!\!\!\!\!\!\!+\sum\limits_\ell\sum\limits_{k}\gamma_{\ell,k}\hat{q}_{\ell,k}\right]]\notag\\
   &= -\frac{\lambda}{2} \int d\nu(\gamma,\tau,\pi) \Tr[
       \left(
        \lambda\mathbb{I}_r+\hat{v}+\sum\limits_{\ell}\sum\limits_{k} \gamma_{\ell,k}\hat{V}_{\ell, k}
        \right)^{-2}\left[\left(\sum\limits_\ell\sum\limits_{k}\hat{m}_{\ell,k} \tau_{\ell,k}
    +\gamma_{\ell,k}\hat{\theta}_{\ell,k} \pi\right)^{\otimes 2}\!\!\!\!\!\!\!\!+\sum\limits_\ell\sum\limits_{k}\gamma_{\ell,k}\hat{q}_{\ell,k}\right]]\notag\\
    &\qquad -\alpha \mathbb{E}_{c,Y,\Xi}
    \mathcal{M}(c,Y,\Xi) +\frac{1}{2}\sum\limits_{\ell=1}^L\sum\limits_{k=1}^{K_\ell}\Tr[\hat{q}_{\ell,k}V_{\ell, k}]\notag\\
    &=-\epsilon_t.
\end{align}
We replaced $q,V,\theta,m,v$ by their self-consistent expressions \eqref{eq:intro:replica_SP_repeat} in going to the second and third lines.
In other words, the training loss is equal to the zero-temperature free energy.

\subsection{An algorithmic perspective}
\label{susbsec:GAMP}

\begin{figure}
\centering
    \hspace{10mm}
    \includegraphics[scale=0.7]{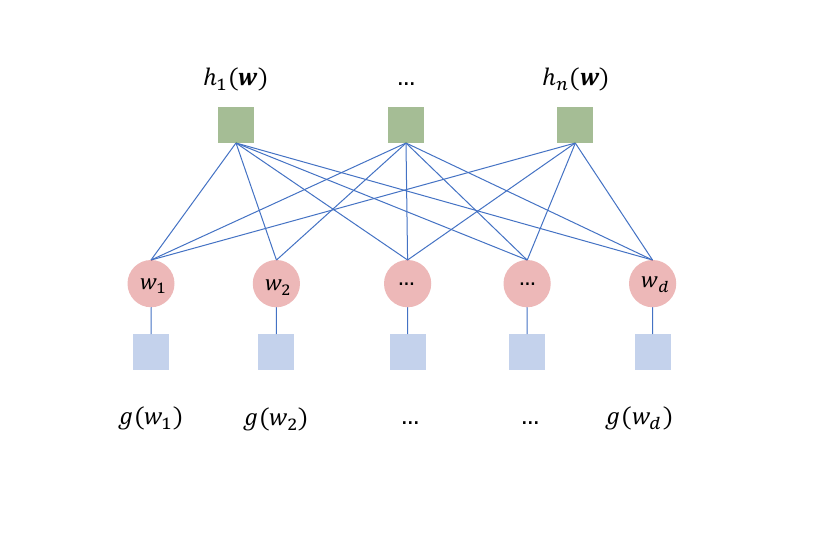}
    \vspace{-10mm}
    \caption{Graphical model associated to the measure $\mathbb{P}_\beta$ \eqref{eq:Z }. We used the shorthands $h_\mu(\w)\equiv \mathrm{exp}(\beta \ell(\sfrac{\x^\mu\teach}{\sqrt{d}},\sfrac{\x^\mu \w}{\sqrt{d}},\sfrac{\w^\top \w}{d}, c^\mu)), g(w_i)=\mathrm{exp}(\beta \sfrac{\lambda}{2}\lVert w_i\lVert^2)$. Iterative schemes such as GAMP (\ref{alg:GAMP}) \cite{rangan2011generalized, rangan2016fixed, javanmard2013state} can be used to estimate marginals from such distributions.}
    \label{fig:Graphical_model}
\end{figure}

The precedent section \ref{subsec:replica} showed how the learning of a \seq could be asymptotically characterized in terms of a set of low-dimensional equations \eqref{eq:intro:replica_SP_repeat}. In this section, we complement this discussion by providing an alternative \emph{algorithmic} viewpoint on the SP equations \eqref{eq:intro:replica_SP_repeat}. More precisely, we will first show that the SP equations \eqref{eq:intro:replica_SP_repeat} describe the fixed points of a GAMP algorithm \cite{rangan2011generalized, rangan2016fixed, javanmard2013state}. Furthermore, the set of fixed point of GAMP will be shown to correspond to critical (zero gradient) points of the empirical ERM landscape. As a result, it follows that aside from the global minimizer, the other solutions of \eqref{eq:intro:replica_SP_repeat} may describe non-global critical points of the empirical landscape. \\

Historically, AMP was first derived in computer science in \cite{donoho2009message} in the context of compressed sensing, and later generalized in \cite{rangan2011generalized}. In physics, related algorithms appeared already in \cite{kabashima1998belief, kabashima2003cdma, opper2001naive, kabashima2004bp}. Interested readers may read the reviews \cite{Gabrie2019MeanfieldIM, zdeborova2016statistical} for further in-depth discussions.

\subsubsection{GAMP algorithm}
How is the replica viewpoint of subsection \ref{subsec:replica} related to iterative algorithms ? A first step in understanding this connection is to observe that the measure $\mathbb{P}_\beta$ \eqref{eq:Z }, which served as a departure point for the discussion of subsection \ref{subsec:replica},  can be represented as a \textit{graphical model}, see Fig.\,\ref{fig:Graphical_model}. For such classes of probability distributions, \emph{message-passing} algorithms \cite{pearl:88, richardson2008modern} provide a versatile framework to evaluate the marginal $\hat{\w}$ (which here correspond to the trained weights), for any given realization of the train set $\mathcal{D}$. We refer the interested reader to \cite{Mzard2009InformationPA, zdeborova2016statistical, Gabrie2019MeanfieldIM} for introductions and reviews of message-passing algorithms. For the measure $\mathbb{P}_\beta$ \eqref{eq:Z }, the corresponding rBP algorithm reads, in the $\beta \to \infty$ limit:

\begin{algorithm}[H]
\caption{rBP}
\label{alg:BP}
\begin{algorithmic}
\STATE \textbf{Inputs} : $\mathcal{D
}=\{\x^\mu, y^\mu,c^\mu\}_{\mu=1}^n $
\STATE \textbf{Initialize} $ \forall 1\le \mu\le n,~~1\le i\le d,~~\hat{w}^0_{i\to \mu}=0_r, \hat{c}^0_{i\to \mu}=\mathbb{I}_{r}$
\FOR{$t\le t_{\max}$}
\STATE $\forall 1\le \ell,\kappa\le L,1\le \mu\le n, 1\le i \le d,~~(V_{\mu\to i }^t)_{\ell\kappa}=\frac{1}{d}\sum\limits_{j\ne i} (x^\mu_{\ell j})(x^\mu_{\kappa j}) \hat{c}^t_{j\to \mu} $
\STATE $1\le \mu\le n, 1\le i \le d,~~\Gamma_{\mu\to i}^t=\frac{1}{d}\sum\limits_{j\ne i}\hat{w}^t_{j\to\mu}(\hat{w}^t_{j\to\mu})^\top$
\STATE $\forall 1\le \ell,1\le \mu\le n, 1\le i \le d, ~~ \omega_{\ell,\mu\to i}^t=\frac{1}{\sqrt{d}}\sum\limits_{j\ne i} x^\mu_{\ell, j} \hat{w}_{j\to \mu}^t$
\STATE $\forall 1\le \ell,1\le \mu\le n, 1\le i \le d,$$ 
f^t_{\ell,\mu\to i}=\sum\limits_{\kappa=1}^L(V_{\mu\to i}^t)^{-1}_{\ell\kappa}\left(\prox(y_\mu,\omega^t_{\mu\to i},V^t_{\mu\to i}, \Gamma^t_{\mu\to i}, c^\mu)-\omega_{\mu\to i}^t\right)_\kappa  $
\STATE $\forall 1\le \mu\le n, 1\le i \le d,~~\eta^t_{\mu\to i}=\partial_3\ell\left(y_\mu, \prox(y_\mu,\omega^t_{\mu\to i},V^t_{\mu\to i}, \Gamma^t_{\mu\to i}, c^\mu), \Gamma^t_{\mu\to i}, c^\mu\right)$
\STATE $\forall 1\le \ell,\kappa\le L,1\le \mu\le n, 1\le i \le d,~~
g_{\ell\kappa, \mu\to i}^t=\nabla_{\omega_\kappa}f_{\ell,\mu\to i}^t
$
\STATE $\forall 1\le \mu\le n, 1\le i \le d,~~ A^t_{i\to\mu}=-\frac{1}{d}\sum\limits_{\ell,\kappa=1}^L \sum\limits_{\nu\ne \mu}(x^\nu_{\ell i})(x^\nu_{\kappa i})g_{\ell\kappa, \nu\to i}^t$
\STATE $\forall 1\le \mu\le n, 1\le i \le d,~~ C^t_{i\to\mu}=\frac{2}{d}\sum\limits_{\nu\ne \mu}\eta_{ \nu\to i}^t$
\STATE $\forall 1\le \mu\le n, 1\le i \le d,~~b^t_{i\to\mu}=\frac{1}{\sqrt{d}}\sum\limits_{\ell=1}^L
\sum\limits_{\nu\ne \mu}x^\nu_{\ell i} f_{\ell,\nu\to i}^t$
\STATE $\forall 1\le \mu\le n, 1\le i \le d,~~\hat{w}^{t+1}_{i\to\mu}=(\lambda\mathbb{I}_r+C^t_{i\to\mu}+A^t_{i\to\mu})^{-1}b_{i\to\mu}^t$
\STATE $\forall 1\le \mu\le n, 1\le i \le d,~~\hat{c}^{t+1}_{i\to\mu}=(\lambda\mathbb{I}_r+C^t_{i\to\mu}+A^t_{i\to\mu})^{-1}$
\ENDFOR
\STATE
\RETURN Estimator $\hat{\w}$
\end{algorithmic}
\end{algorithm}
We have denoted $\partial_3\ell$ the partial derivative of the loss function $\ell$ with respect to its third argument. We furthermore remind that $x^\mu_{\ell i}$ represents the $i-$th element of the $\ell-$th token of the $\mu-$th sample. 
In the rBP iterations \ref{alg:BP}, the variable $V^t_{\mu\to i} \in\mathbb{R}^{rL\times rL}$ is viewed as a block matrix with blocks of size $r\times r$, while $\omega^t_{\mu\to i}\in\mathbb{R}^{L\times r}$ is a matrix with rows $(\omega^t_{\mu\to i})_\ell\in \R^r$ for $1\le \ell\le L$. We also introduced the resolvent
 \begin{align}
 \label{eq:prox_GAMP}
     &\prox(y,\omega,V,\Gamma,c)\equiv\underset{X\in\R^{L\times r}}{\mathrm{arginf}}\Bigg\{ 
\frac{1}{2} \sum\limits_{1\le \ell,\kappa\le L}(X-\omega)_\ell^\top (V^{-1})_{\ell\kappa }(X-\omega)_\kappa
     +\ell(y, X,\Gamma,c)
     \Bigg\}\in\mathbb{R}^{L\times r}.
 \end{align}
We finally remind that $y_\mu=\sfrac{\x^\mu \teach}{\sqrt{d}} \in \R^{L\times t}$ correspond to the label associated to the $\mu-$th sample, in supervised settings. While formally simple, the rBP iterations (\ref{alg:BP}) nevertheless bear over a staggering $n\times d\times (2L^2r^2+2Lr+5r^2+2r)=\Theta_d(d^2)$ variables, and is thus computationally unwieldy. The rBP iterations (\ref{alg:BP}) can actually be rewritten in terms on only $\Theta_d(d)$ variables, yielding the more compact GAMP iterations (\ref{alg:GAMP}): 

\begin{algorithm}[H]
\caption{GAMP}
\label{alg:GAMP}
\begin{algorithmic}
\STATE \textbf{Inputs} : $\mathcal{D
}=\{\x^\mu, y^\mu,c^\mu\}_{\mu=1}^n $
\STATE \textbf{Initialize} $ \forall 1\le \mu\le n,~~1\le i\le d,~~\hat{w}^0_{i}=0_r, \hat{c}^0_{i}=\mathbb{I}_{r}$
\FOR{$t\le t_{\max}$}
\STATE $\forall 1\le \ell,\kappa\le L,1\le \mu\le n,~~(V_{\mu}^t)_{\ell\kappa}=\frac{1}{d}\sum\limits_{i} (x^\mu_{\ell i})(x^\mu_{\kappa i}) \hat{c}^t_{i} $
\STATE $\Gamma^t=\frac{1}{d}\sum\limits_{i}\hat{w}^t_{i}(\hat{w}^t_{i})^\top$
\STATE $\forall 1\le \ell,1\le \mu\le n, ~~ \omega_{\ell,\mu}^t=\frac{1}{\sqrt{d}}\sum\limits_{i} x^\mu_{\ell, i} \hat{w}_{i}^t-\sum\limits_\kappa(V^t_\mu)_{\ell\kappa} f_{\kappa\mu} $
\STATE $\forall 1\le \ell,1\le \mu\le n,$$~~ 
f^t_{\ell,\mu}=\sum\limits_{\kappa=1}^L (V_{\mu}^t)^{-1}_{\ell\kappa}\left(\prox(y_\mu,\omega^t_{\mu},V^t_{\mu}, \Gamma^t, c^\mu)-\omega_{\mu}^t\right)_\kappa  $
\STATE $\forall 1\le \mu\le n,~~\eta^t_{\mu}=\partial_3\ell\left(y_\mu, \prox(y_\mu,\omega^t_{\mu},V^t_{\mu}, \Gamma^t, c^\mu), \Gamma^t_{\mu}, c^\mu\right)$
\STATE $\forall 1\le \ell,\kappa\le L,1\le \mu\le n,~~
g_{\ell\kappa,\mu}^t=\nabla_{\omega_\kappa}f_{\ell,\mu}^t
$
\STATE $ 1\le i \le d,~~ A^t_{i}=-\frac{1}{d}\sum\limits_{\ell,\kappa=1}^L \sum\limits_{\mu}(x^\mu_{\ell i})(x^\mu_{\kappa i})g_{\ell\kappa, \mu}^t$
\STATE $ C^t=\frac{2}{d}\sum\limits_{\mu}\eta_{\mu}^t$
\STATE $1\le i \le d,~~b^t_{i}=\frac{1}{\sqrt{d}}\sum\limits_{\ell=1}^L
\sum\limits_{\mu}x^\mu_{\ell i} f_{\ell,\mu}^t+A^t_i\hat{w}^t_i$
\STATE $\forall 1\le \mu\le n, 1\le i \le d,~~\hat{w}^{t+1}_{i}=(\lambda\mathbb{I}_r+C^t+A^t_{i})^{-1}b_{i}^t$
\STATE $\forall 1\le \mu\le n, 1\le i \le d,~~\hat{c}^{t+1}_{i}=(\lambda\mathbb{I}_r+C^t+A^t_{i})^{-1}$
\ENDFOR
\STATE
\RETURN Estimator $\hat{\w}$
\end{algorithmic}
\end{algorithm}

The rBP (\ref{alg:BP}) and GAMP (\ref{alg:GAMP}) algorithms are in fact asymptotically equivalent. For completeness, we detail in Appendix \ref{app: BP} how rBP (\ref{alg:BP}) and GAMP (\ref{alg:GAMP}) can be derived from the \textit{Belief Propagation }\cite{Mzard2009InformationPA, bayati2011dynamics} algorithm, and explicate the asymptotic equivalence between the three schemes. We are thus now in possession of two equivalent algorithmic schemes (\ref{alg:BP}, \ref{alg:GAMP}) to evaluate the marginal $\hat{\w}$ of the distribution $\mathbb{P}_\beta$ \eqref{eq:Z } in the $\beta\to\infty$ limit -- the very same object and measure we characterized in subsection \ref{subsec:replica} using the replica method. In the next paragraphs, we will show that the replica equations \eqref{eq:intro:replica_SP_repeat} in fact describe the fixed points of the rBP and GAMP algorithms (\ref{alg:BP}, \ref{alg:GAMP}).

\subsubsection{State evolution}
\label{subsec:SE}
We provide in this subsection a detailed derivation of the connection between the SP equations \eqref{eq:intro:replica_SP_repeat} derived in subsection \ref{subsec:replica} and the just introduced GAMP algorithm (\ref{alg:GAMP}). Before doing so, let us sketch in broad strokes the elements of this connection.
As we shall see, all GAMP variables (\ref{alg:GAMP}) are asymptotically either deterministic, or Gaussian-distributed, with respect to the randomness of the train set $\mathcal{D}$. Their dynamics is thus entirely tracked by that of their limiting values or second-order statistics. These summary statistics in fact identically coincide with the RS order parameters \eqref{eq:intro:RS} of subsection \ref{subsec:replica}. Further even : the evolution of these summary statistics is given by the (time-indexed version of) the SP equations \eqref{eq:intro:replica_SP_repeat} -- implying in particular that the latter also describe fixed points of GAMP. \\

To build these connections, it is convenient to take as a starting point the equivalent rBP equations (\ref{alg:BP}), rather than the GAMP equations (\ref{alg:GAMP}).
 In the following, we examine each of the variables $V^t_{\mu\to i}$, $\omega^t_{\mu\to i}$, $f^t_{\mu\to i}$, $g^t_{\mu\to i}$, $A^t_{i\to \mu}$, $b^t_{i\to \mu}$, $\hat{w}^t_{i\to \mu}$, $\hat{c}^t_{i\to \mu}$ involved in the rBP iterations, and ascertain their probability distribution. 
As a convention, we note $\circ_{\mu}$ the version of a variable $\circ_{\mu\to i}$ where the summation also encompasses the index $i$, and $\circ_{i}$ the version of a variable $\circ_{i\to \mu}$ where the summation also encompasses the index $\mu$. Note that in all cases above the two variables $\circ_{\mu},\circ_{\mu\to i}$ or $\circ_{i},\circ_{i\to \mu} $ differ by at most $\Theta_d(\sfrac{1}{\sqrt{d}})$. Further, note that under the assumption of section \ref{sec:seq-GLM} that all covariances are jointly diagonalizable, one can assume without loss of generality all covariances $\Sigma_{\ell,k}$ to be diagonal. Let us make one final important remark. Given an index $1\le i\le d$ (resp. $1\le \mu \le n$), it is reasonable to expect from the scaling of equations (\ref{alg:BP}) that all incoming variables $\{\circ_{\nu\to i}\}_{1\le \nu\le n}$ (resp $\{\circ_{j\to \mu}\}_{1\le j\le d}$) are asymptotically weakly correlated, between themselves and with the data elements $\{(x^\nu_\ell)_i\}_{\ell,\nu}$ (resp $\{(x^\mu_\ell)_j\}_{\ell,j}$). These groups of variables can thus be assumed to be independent. This is in fact a standard assumption in the derivation and analysis of GAMP equations, see e.g. \cite{zdeborova2016statistical}.

\paragraph{Concentration of $ (V_{\mu\to i }^t)_{\ell\kappa}$, $ (\Gamma_{\mu\to i }^t)_{\ell\kappa}$} We first show that the variables $V_{\mu\to i }^t$ concentrate to a deterministic value. Let us first massage its expression as
\begin{align}
\label{eq:introduce_V}
    (V_{\mu\to i }^t)_{\ell\kappa}&=\frac{1}{d}\sum\limits_{j\ne i} (x^\mu_{\ell j})(x^\mu_{\kappa j}) \hat{c}^t_{j\to \mu}\notag\\
    &=\underbrace{\frac{1}{d}\sum\limits_{j\ne i} (\tilde{x}^\mu_{\ell })_j(\tilde{x}^\mu_{\kappa })_j \hat{c}^t_{j\to \mu}}_{\delta_{\ell\kappa}\Theta_d(1)+(1-\delta_{\ell\kappa})\Theta_d(\sfrac{1}{\sqrt{d}})}+\underbrace{
    \frac{1}{d}\sum\limits_{j\ne i} (\tilde{x}^\mu_{\ell })_j(\mu_{\kappa,c^\mu_\kappa })_j \hat{c}^t_{j\to \mu}+(\ell \leftrightarrow{}\kappa)}_{\Theta_d(\sfrac{1}{d})}+ \underbrace{\frac{1}{d}\sum\limits_{j\ne i} (\mu_{\ell ,c^\mu_\ell})_j (\mu_{\kappa ,c^\mu_\kappa})_j \hat{c}^t_{j\to \mu}}_{\Theta_d(\sfrac{1}{d})}\notag\\
    \end{align}
    where we denoted $\tilde{\x}_\ell=\x_\ell-\mean_\ell$ the centered data.  By the law of large numbers $(V_{\mu\to i }^t)_{\ell\kappa}$ then concentrates to leading order to the deterministic value
    \begin{align}
         (V_{\mu\to i }^t)_{\ell\kappa}&=\delta_{\ell\kappa}\frac{1}{d}\sum\limits_{j} (\Sigma_{\ell,c_\ell^\mu})_{jj}\hat{c}_j^t\equiv \delta_{\ell\kappa}V^t_{\ell,c^\mu_\ell}.
    \end{align}
    We introduced the summary statistic $V^t_{\ell,c_\ell}$. By the same token, $ (\Gamma_{\mu\to i }^t)_{\ell\kappa}$ concentrates to
    \begin{align}
        \label{eq:introduce_v}
        \Gamma_{\mu\to i}^t=\frac{1}{d}\sum\limits_{i}\hat{w}^t_{i}(\hat{w}^t_{i})^\top\equiv v^t,
    \end{align}
where we introduced the summary statistic $v^t$. 

\paragraph{Distribution of $\omega_{\ell,\mu\to i}^t$}
We now move to examine the probability distribution of $\omega_{\ell,\mu\to i}^t$. It shall prove convenient for later purposes to jointly examine the companion random variable
\begin{align}
    \Tilde{y}_{\mu,\ell}=\frac{1}{\sqrt{d}}\sum\limits_{i}(\tilde{x}^\mu_\ell)_i w^\star_i=y_{\mu\ell}-m^\star_{\ell,c^\mu_\ell},
\end{align}
which correspond to the centered label. As a sum of asymptotically independent variables, $\omega_{\ell,\mu\to i}^t$ is thus Gaussian-distributed according to the \textit{Central Limit Theorem} (CLT).
We can now ascertain the joint distribution of $\tilde{y}_{\mu,\ell},\omega_{\ell,\mu\to i}^t$. These variables have mean
\begin{align}
\label{eq:introduce_m}
    \mathbb{E}[\omega_{\ell,\mu\to i}^t]&=\frac{\mean_{\ell,c^\mu_\ell}^\top \hat{\w}^t}{\sqrt{d}}\equiv m_{\ell,c^\mu_\ell}^t,&& \mathbb{E}[\Tilde{y}_{\mu,\ell}]=0,
\end{align}
and respective variance
\begin{align}
\label{eq:introduce_qrhotheta}
    &\mathbb{E}[(\omega_{\ell,\mu\to i}^t-m_{\ell,c^\mu_\ell}^t)(\omega_{\kappa,\nu\to j}^t-m_{\kappa,c^\mu_\kappa}^t)^\top ]=\delta_{\mu\nu}\delta_{\ell\kappa}\frac{1}{d}\sum\limits_{i,j}\hat{w}^t_i(\Sigma_{\ell,c^\mu_\ell})_{ij}(\hat{w}^t_j)^\top \equiv \delta_{\mu\nu}\delta_{\ell\kappa} q_{\ell,c^\mu_\ell}^t,\\
    &\mathbb{E}[\tilde{y}_{\mu\ell}\tilde{y}_{\nu\kappa}^\top ]=\delta_{\mu\nu}\delta_{\ell\kappa}\frac{1}{d}\sum\limits_{i,j}w^\star_i(\Sigma_{\ell,c^\mu_\ell})_{ij}(w^\star_j)^\top  \equiv \delta_{\mu\nu}\delta_{\ell\kappa} \rho_{\ell,c^\mu_\ell},\\
    &\mathbb{E}[(\omega_{\kappa,\nu\to j}^t-m_{\kappa,c^\mu_\kappa}^t)\tilde{y}_{\mu\ell}^\top]=\delta_{\mu\nu}\delta_{\ell\kappa} \frac{1}{d}\sum\limits_{i,j}(\Sigma_{\ell,c^\mu_\ell})_{ij}\hat{w}^t_j (w^\star_i)^\top \equiv \delta_{\mu\nu}\delta_{\ell\kappa} \theta^t_{\ell,c^\mu_\ell}.
\end{align}
We introduced the summary statistics $q^t_{\ell,k},\rho_{\ell,k},\theta_{\ell,k}^t, m^t_{\ell,k}$.

\paragraph{Distribution of $b_{i\to\mu}^t$} Let us now study the distribution of $b_{i\to\mu}^t$. Most of the variables appearing as arguments of the resolvent, such as $\omega_{\nu\to i}$, can again be considered as independent with the $i-$the data component $(x^\nu_\ell)_i$. The first argument, namely the label $y_\nu$, is an exception, as it explicitly involves $(x^\nu_\ell)_i$. To disentangle these different statistical dependencies, we expand the resolvent in its first argument as
\begin{align}
    b_{i\to\mu}^t&=\frac{1}{\sqrt{d}}\sum\limits_{\ell}\sum\limits_{\nu\ne \mu}(x^\nu_\ell)_i f^t_{\ell, \nu\to i}\notag\\
    &=    \frac{1}{\sqrt{d}}\sum\limits_{\ell}\sum\limits_{\nu\ne \mu}((\tilde{x}^\nu_\ell)_i+(\mu_{\ell,c^\nu_\ell})_i)\left(1+\frac{1}{\sqrt{d}}\sum\limits_\gamma (\tilde{x}^\nu_\gamma)_i (w^\star_i \cdot \nabla_{y_\gamma})\right)\notag\\
    &\qquad\qquad\qquad\qquad\left[(V_{\nu\to i}^t)^{-1}\left(\prox(y_{\nu\to i},\omega^t_{\nu\to i},V^t_{\nu\to i},\Gamma^t_{\nu\to i},c_\nu)-\omega_{\nu\to i}^t\right)\right]_\ell\notag\\
    &=    \frac{1}{\sqrt{d}}\sum\limits_{\ell}\sum\limits_{\nu\ne \mu}((\tilde{x}^\nu_\ell)_i+(\mu_{\ell,c^\nu_\ell})_i)\left(1+\frac{1}{\sqrt{d}}\sum\limits_\gamma (\tilde{x}^\nu_\gamma)_i (w^\star_i \cdot \nabla_{y_\gamma})\right)\notag\\
    &\qquad\qquad\qquad\qquad(V_{\ell,c^\nu_\ell}^t)^{-1}\left(\prox(y_{\nu\to i},\omega^t_{\nu\to i},V^t_{\nu\to i},\Gamma^t_{\nu\to i},c_\nu)-\omega_{\nu\to i}^t\right)_\ell
\end{align}
We denoted $y_{\mu\to i}\equiv y_\mu- x^\mu_i w^\star_i$, and used in the last line the block-diagonal structure of $V^t_{\nu\to i}$ that follows from \eqref{eq:introduce_V}. All arguments of the resolvent $y_{\nu\to i},\omega^t_{\nu\to i}$ are now approximately independent of $(\tilde{x}^\nu_\ell)_i$. As for $\omega_{\ell,\mu\to i}^t$, it then follows from the CLT that $b_{i\to\mu}^t$ asymptotically follows a Gaussian distribution with mean
\begin{align}
\label{eq:introduce_thetahat}
    \mathbb{E}[b_{i\to\mu}^t]=&  \sum\limits_{\ell}\sum\limits_{k}(\sqrt{d} \mu_{\ell,k})_i \underbrace{\alpha\mathbb{E}_c\delta_{c_\ell,k}\mathbb{E}_{y^c,\Xi^c}(V^t_{\ell,k})^{-1}\left[\prox(y^c, m_c^t+\Xi^c, V^t_c,v^t,c)_\ell-
    (m^t_{\ell,k}+\Xi^c_\ell)\right]}_{\equiv \hat{m}^t_{\ell,k}}\notag\\
    &+ \sum\limits_\ell\sum\limits_{k}(\Sigma_{\ell,k})_{ii}
    \underbrace{\alpha\mathbb{E}_c\delta_{c_\ell,k}\mathbb{E}_{y^c,\Xi^c}(V^t_{\ell,k})^{-1}\nabla_{y_\ell}\left[\prox(y^c, m_c^t+\Xi^c, V^t_c,v^t,c)_\ell-
    (m^t_{\ell,k}+\Xi^c_\ell)\right] }_{\equiv \hat{\theta}^t_{\ell,k}}w^\star_i,
\end{align}
where the expectations bear over $\Xi^c\in\mathbb{R}^{L\times r}$ with colored Gaussian rows $(q^t_{\ell,c_\ell})^{\sfrac{1}{2}}\xi_\ell$, where $\xi_\ell\sim\mathcal{N}(0_r,\mathbb{I}_r)$, and $y\in\mathbb{R}^{L\times t}$ with rows $y^c_\ell\sim\mathcal{N}(m^\star_{\ell,c_\ell}+ (\theta^t_{\ell,c_\ell})^\top(q^t_{\ell,c_\ell})^{-\sfrac{1}{2}}\xi_\ell, \rho_{\ell,c_\ell}-(\theta^t_{\ell,c_\ell})^\top(q^t_{\ell,c_\ell})^{-1}\theta^t_{\ell,c_\ell})$. We further denoted by $V^t_c\in\mathbb{R}^{rL\times rL}$ the block-diagonal matrix with blocks $V^t_{\ell,c_\ell}$.  The variance of $b_{i\to\mu}^t$ can similarly be evaluated as
\begin{align}
\label{eq:intro_qhat}
    \mathbb{V}[b_{i}^t,b_{j}^t]=\delta_{ij} \sum\limits_\ell\sum\limits_{k}(\Sigma_{\ell,k})_{ii}
    \underbrace{\alpha\mathbb{E}_c\delta_{c_\ell,k}\mathbb{E}_{y^c,\Xi^c}(V^t_{\ell,k})^{-1}\left[\prox(y^c, m_c^t+\Xi^c, V^t_c,v^t,c)_\ell-
    (m^t_{\ell,k}+\Xi^c_\ell)\right]^{\otimes 2}(V^t_{\ell,k})^{-1}}_{\equiv \hat{q}^t_{\ell,k}}.
\end{align}
We introduced the summary statistics $\hat{q}^t_{\ell,k}, \hat{m}^t_{\ell,k},\hat{\theta}^t_{\ell,k}$.

\paragraph{Concentration of $A^t_{i\to\mu},C^t_{i\to\mu} $}
Finally, like $V^t_{\mu\to i}$, $A^t_{i\to\mu}$ concentrates to a deterministic value
\begin{align}
\label{eq:intro_Vhat}
    A^t_{i\to\mu}= \sum\limits_\ell 
    \sum\limits_k
    (\Sigma_{\ell,k})_{ii}
    \left[\underbrace{-\alpha\mathbb{E}_c\delta_{c_\ell,k}\left(\mathbb{E}_{y^c
    ,\Xi^c}(V^t_{\ell,k})^{-1}\nabla_{\omega_\ell}\prox(y^c, m^t_c+\Xi^c, V^t_c,v^t,c)_\ell
    -1
    \right)}_{\equiv\hat{V}^t_{\ell,k}}\right]
\end{align}
We introduced the summary statistics $\hat{V}^t_{\ell,c_\ell}$. Similarly, $C^t_{i\to\mu}$ concentrates to
\begin{align}
    \label{eq:introduce_vhat}
C^t_{i\to\mu}=2\alpha \mathbb{E}_c\mathbb{E}_{y^c,\Xi^c}\partial_3\ell\left(y^c, \prox(y^c, m^t_c+\Xi^c, V^t_c,v^t,c),  V^t_c
\right)\equiv \hat{v}^t.
\end{align}
\\

All the variables involved in the rBP iterations are thus either Gaussian-distributed or deterministic, making it possible to concisely capture their asymptotic dynamics with a small set of summary statistics $q^t_{\ell,k},\theta_{\ell,k}^t, m^t_{\ell,k},V^t_{\ell,k},v^t$, $\hat{q}^t_{\ell,k}, \hat{m}^t_{\ell,k},\hat{\theta}^t_{\ell,k}, \hat{V}^t_{\ell,k},\hat{v}^t$. In the following paragraph, we derive the update equations obeyed by these statistics, and show that they coincide with a time-indexed version of the SP equations \eqref{eq:intro:replica_SP_repeat} previously derived from the replica method. In particular, the set of self-consistent equations \eqref{eq:intro:replica_SP_repeat} is satisfied at convergence by the infinite-time iterates $q^\infty_{\ell,k},\theta_{\ell,k}^\infty, m^\infty_{\ell,k},V^\infty_{\ell,k},v^\infty$, $\hat{q}^\infty_{\ell,k}, \hat{m}^\infty_{\ell,k},\hat{\theta}^\infty_{\ell,k}, \hat{V}^\infty_{\ell,k},\hat{v}^\infty$.

\paragraph{Recovering equations \eqref{eq:intro:replica_SP_repeat}} Wrapping up, we now massage the equations \eqref{eq:introduce_m}, \eqref{eq:introduce_v}, \eqref{eq:introduce_V}, \eqref{eq:introduce_vhat}. \eqref{eq:introduce_thetahat} and \eqref{eq:introduce_qrhotheta} to access the update equations describing the dynamics of the summary statistics $q^t_{\ell,k},\theta_{\ell,k}^t, m^t_{\ell,k},V^t_{\ell,k},v^t$, $\hat{q}^t_{\ell,k}, \hat{m}^t_{\ell,k},\hat{\theta}^t_{\ell,k}, \hat{V}^t_{\ell,k},\hat{v}^t$.  Starting from $V_{\ell,k}^t$ \eqref{eq:introduce_V}, one reaches
\begin{align}
    V_{\ell,k}^t&=\frac{1}{d}\sum_i (\Sigma_{\ell,k})_{ii}\left(\lambda \mathbb{I}_r+\hat{v}^t+\sum\limits_\kappa\sum\limits_{j} \hat{V}^{t-1}_{\kappa,j} (\Sigma_{\kappa,j})_{ii}\right)^{-1}=\int d\nu(\gamma,\tau,\pi)\gamma_{\ell,k}\left(\lambda \mathbb{I}_r +\hat{v}^t+\sum\limits_\kappa\sum\limits_{j} \hat{V}^{t-1}_{\kappa,j} \gamma_{\kappa,j} \right)^{-1}.
\end{align}
Next, for $v^t$ \eqref{eq:introduce_qrhotheta}:
\begin{align}
    v^t&=\frac{1}{d}\sum\limits_{i}\left(\lambda \mathbb{I}_r+\hat{v}^t+\sum\limits_\kappa\sum\limits_{j} \hat{V}^{t-1}_{\kappa,j} (\Sigma_{\kappa,j})_{ii}\right)^{-1}
    \notag \\
    & \qquad\!\!\!\!\left[\left(
    \sum\limits_\kappa\sum\limits_{j} \sqrt{d}(\mu_{\kappa,j})_i \hat{m}_{\kappa,j}^{t-1}+ (\Sigma_{\kappa,j})_{ii}\hat{\theta}_{\kappa,j}^{t-1}w^\star_i
    \right)^{\otimes 2}\!\!\!\!\!\!\!\!+ \sum\limits_\kappa\sum\limits_{j} (\Sigma_{\kappa,j})_{ii}\hat{q}^{t-1}_{\kappa,j}
    \right]\left(\lambda \mathbb{I}_r+\hat{v}^t+\sum\limits_\kappa\sum\limits_{j} \hat{V}^{t-1}_{\kappa,j} (\Sigma_{\kappa,j})_{ii}\right)^{-1}\notag\\
    &=\int d\nu(\gamma,\tau,\pi)  \left(\lambda \mathbb{I}_r +\hat{v}^t+\sum\limits_\kappa\sum\limits_{j} \hat{V}^{t-1}_{\kappa,j} \gamma_{\kappa,j} \right)^{-1}\notag\\
    &\qquad\left[\left(
    \sum\limits_\kappa\sum\limits_{j} (\tau_{\kappa,j} \hat{m}_{\kappa,j}^{t-1}+ \gamma_{\kappa,j}\hat{\theta}_{\kappa,j}^{t-1}\pi
    \right)^{\otimes 2}\!\!\!\!\!\!\!\!+ \sum\limits_\kappa\sum\limits_{j} \gamma_{\kappa,j}\hat{q}^{t-1}_{\kappa,j}
    \right]
    \left(\lambda \mathbb{I}_r +\hat{v}^t+\sum\limits_\kappa\sum\limits_{j} \hat{V}^{t-1}_{\kappa,j} \gamma_{\kappa,j} \right)^{-1}.
\end{align}
Next, for $q^t_{\ell,k}$ \eqref{eq:introduce_qrhotheta}:
\begin{align}
    q^t_{\ell,k}&=\frac{1}{d}\sum\limits_{i}(\Sigma_{\ell,k})_{ii}\left(\lambda \mathbb{I}_r+\hat{v}^t+\sum\limits_\kappa\sum\limits_{j} \hat{V}^{t-1}_{\kappa,j} (\Sigma_{\kappa,j})_{ii}\right)^{-1}
    \notag \\
    & \qquad\!\!\!\!\left[\left(
    \sum\limits_\kappa\sum\limits_{j} \sqrt{d}(\mu_{\kappa,j})_i \hat{m}_{\kappa,j}^{t-1}+ (\Sigma_{\kappa,j})_{ii}\hat{\theta}_{\kappa,j}^{t-1}w^\star_i
    \right)^{\otimes 2}\!\!\!\!\!\!\!\!+ \sum\limits_\kappa\sum\limits_{j} (\Sigma_{\kappa,j})_{ii}\hat{q}^{t-1}_{\kappa,j}
    \right]\left(\lambda \mathbb{I}_r+\hat{v}^t+\sum\limits_\kappa\sum\limits_{j} \hat{V}^{t-1}_{\kappa,j} (\Sigma_{\kappa,j})_{ii}\right)^{-1}\notag\\
    &=\int d\nu(\gamma,\tau,\pi) \gamma_{\ell k} \left(\lambda \mathbb{I}_r +\hat{v}^t+\sum\limits_\kappa\sum\limits_{j} \hat{V}^{t-1}_{\kappa,j} \gamma_{\kappa,j} \right)^{-1}\notag\\
    &\qquad\left[\left(
    \sum\limits_\kappa\sum\limits_{j} (\tau_{\kappa,j} \hat{m}_{\kappa,j}^{t-1}+ \gamma_{\kappa,j}\hat{\theta}_{\kappa,j}^{t-1}\pi
    \right)^{\otimes 2}\!\!\!\!\!\!\!\!+ \sum\limits_\kappa\sum\limits_{j} \gamma_{\kappa,j}\hat{q}^{t-1}_{\kappa,j}
    \right]
    \left(\lambda \mathbb{I}_r +\hat{v}^t+\sum\limits_\kappa\sum\limits_{j} \hat{V}^{t-1}_{\kappa,j} \gamma_{\kappa,j} \right)^{-1}.
\end{align}
For $\theta^t_{\ell,k}$\eqref{eq:introduce_qrhotheta}:
\begin{align}
    \theta^t_{\ell,k}&=\frac{1}{d}\sum\limits_i(\Sigma_{\ell,k})_{i i}  \left(\lambda \mathbb{I}_r +\hat{v}^t+\sum\limits_\kappa\sum\limits_{j} \hat{V}^{t-1}_{\kappa,j} \gamma_{\kappa,j} \right)^{-1}\left(
    \sum\limits_\kappa\sum\limits_{j} (\sqrt{d}(\mu_{\kappa,j})_i \hat{m}_{\kappa,j}^{t-1}+ (\Sigma_{\kappa,j})_{ii}\hat{\theta}_{\kappa,j}^{t-1}w^\star_i
    \right)(w^\star_i)^\top\notag\\
    &=\int d\nu(\gamma,\tau,\pi) \gamma_{\ell,k } \left(\lambda \mathbb{I}_r +\hat{v}^t+\sum\limits_\kappa\sum\limits_{j} \hat{V}^{t-1}_{\kappa,j} \gamma_{\kappa,j} \right)^{-1} \left(
    \sum\limits_\kappa\sum\limits_{j} (\tau_{\kappa,j} \hat{m}_{\kappa,j}^{t-1}+ \gamma_{\kappa,j}\hat{\theta}_{\kappa,j}^{t-1}\pi
    \right)\pi^\top.
\end{align}
For $m^t_{\ell,k}$ \eqref{eq:introduce_m}:
\begin{align}
    m^t_{\ell,k}&=\frac{1}{d}\sum\limits_i (\sqrt{d} \mu_{\ell,k})_i\left(\lambda \mathbb{I}_r +\hat{v}^t+\sum\limits_\kappa\sum\limits_{j} \hat{V}^{t-1}_{\kappa,j} \gamma_{\kappa,j} \right)^{-1}\left(
    \sum\limits_\kappa\sum\limits_{j} (\sqrt{d}(\mu_{\kappa,j})_i \hat{m}_{\kappa,j}^{t-1}+ (\Sigma_{\kappa,j})_{ii}\hat{\theta}_{\kappa,j}^{t-1}w^\star_i
    \right)\notag\\
    &=\int d\nu(\gamma,\tau,\pi) \tau_{\ell,k } \left(\lambda \mathbb{I}_r +\hat{v}^t+\sum\limits_\kappa\sum\limits_{j} \hat{V}^{t-1}_{\kappa,j} \gamma_{\kappa,j} \right)^{-1} \left(
    \sum\limits_\kappa\sum\limits_{j} (\tau_{\kappa,j} \hat{m}_{\kappa,j}^{t-1}+ \gamma_{\kappa,j}\hat{\theta}_{\kappa,j}^{t-1}\pi
    \right).
\end{align}
For $\hat{m}^t_{\ell,k}$ \eqref{eq:introduce_thetahat}:
\begin{align}
    \hat{m}^t_\ell= \alpha\mathbb{E}_{c}\delta_{c_\ell,k}\mathbb{E}_{y,\Xi}(V^t_{\ell,k})^{-1}\left[
    \prox_\ell^c-(q^t_{\ell,k})^{\sfrac{1}{2}}\xi_\ell-m^t_{\ell,k}
    \right],
\end{align}
while for $\hat{\theta}^t_{\ell,k}$ \eqref{eq:introduce_thetahat}:
\begin{align}
    \hat{\theta}^t_{\ell,k}&= \alpha \mathbb{E}_{c}\delta_{c_\ell,k}\mathbb{E}_{y,\Xi}(V^t_{\ell,k})^{-1}\nabla_{y_\ell}\left[
    \prox_\ell^c-(q^t_{\ell,k})^{\sfrac{1}{2}}\xi_\ell-m^t_{\ell,k}
    \right]\notag\\
    &=\alpha \mathbb{E}_{c}\delta_{c_\ell,k}\mathbb{E}_{y,\Xi}(V^t_{\ell,k})^{-1}\left[
    \prox_\ell^c-(q^t_{\ell,k})^{\sfrac{1}{2}}\xi_\ell-m^t_{\ell,k}
    \right]\notag\\
    &\qquad\qquad\qquad\qquad\qquad~~\left(y_\ell-m^\star_{\ell,k}-(\theta^t_{\ell,k})^\top(q^t_{\ell,k})^{-\sfrac{1}{2}}\xi_\ell\right)^\top\left(\rho_{\ell,k}-(\theta^t_{\ell,k})^\top(q^t_{\ell,k})^{-1}\theta^t_{\ell,k}\right)^{-1}.
\end{align}
Now turning to $\hat{q}^t_{\ell,k}$:
\begin{align}
    \hat{q}^t_{\ell,k}=\alpha\mathbb{E}_{c}\delta_{c_\ell,k}\mathbb{E}_{y,\Xi}\left[(V^t_{\ell,k})^{-1}\left[
    \prox_\ell^c-(q^t_{\ell,k})^{\sfrac{1}{2}}\xi_\ell-m^t_{\ell,k}
    \right]^{\otimes 2}(V^t_{\ell,k})^{-1}
    \right].
\end{align}
For $\hat{v}^t$:
\begin{align}
    \hat{v}^t=2\alpha \mathbb{E}_c\mathbb{E}_{y,\Xi}\partial_3\ell\left(y^c, \prox^c,  V^t_c,v^t,c
\right)\equiv \hat{v}^t
\end{align}
Finally, for $\hat{V}^t_{\ell,k}$ \eqref{eq:intro_Vhat}:
\begin{align}
    \hat{V}^t_{\ell,k}&=-\alpha \mathbb{E}_{c}\delta_{c_\ell,k}\mathbb{E}_{y,\Xi}(V^t_{\ell,k})^{-1}\left[
    \nabla_{\omega_\ell}\prox_\ell^c-1
    \right]\notag\\
    &=-\alpha\mathbb{E}_{c}\delta_{c_\ell,k} \mathbb{E}_{y,\Xi}(V^t_{\ell,k})^{-1}\left[
   \nabla_{\xi_\ell}(\prox_\ell^c-(q^t_{\ell,k})^{\sfrac{1}{2}}\xi_\ell-m^t_{\ell,k})(q^t_{\ell,k})^{-\sfrac{1}{2}}
    \right]\notag\\
    &=\alpha \mathbb{E}_{c}\delta_{c_\ell,k}\mathbb{E}_{y,\Xi}\left[
    (V^t_{\ell,k})^{-1}
    (\prox_\ell^c-(q^t_{\ell,k})^{\sfrac{1}{2}}\xi_\ell-m^t_{\ell,k})\right.\notag\\
    &\left. \qquad
    \left[\left(y_\ell-m^\star_{\ell,k}-(\theta^t_{\ell,k})^\top(q^t_{\ell,k})^{-\sfrac{1}{2}}\xi_{\ell,k}\right)^\top\left(\rho_{\ell,k}-(\theta^t_{\ell,k})^\top(q^t_{\ell,k})^{-1}\theta^t_{\ell,k}\right)^{-1}(\theta^t_{\ell,k})^\top (q^t_{\ell,k})^{-\sfrac{1}{2}}-\xi_{\ell,k}^\top\right](q^t_{\ell,k})^{-\sfrac{1}{2}}
    \right]\notag\\
    &= \hat{\theta}^t_{\ell,k}(\theta^t_{\ell,k})^\top(q^t_{\ell,k})^{-1}-\alpha  \mathbb{E}_{c}\delta_{c_\ell,k}\mathbb{E}_{y,\Xi}(V^t_{\ell,k})^{-1}(\prox_\ell^c-(q^t_{\ell,k})^{\sfrac{1}{2}}\xi_\ell-m^t_{\ell,k})\xi_\ell^\top(q^t_{\ell,k})^{-\sfrac{1}{2}}.
\end{align}
This concludes the derivation of the update equations satisfied by the set of summary statistics $q^t_{\ell,k},\theta_{\ell,k}^t, m^t_{\ell,k},V^t_{\ell,k}$, $\hat{q}^t_{\ell,k}, \hat{m}^t_{\ell,k},\hat{\theta}^t_{\ell,k}, \hat{V}^t_{\ell,k}$, called the SE equations. These equations concisely describe the macroscopic asymptotic behaviour of the rBP (equivalently GAMP) iterates, thus abstracting away the precise dynamics of the $\Theta_d(d^2)$ variables $V^t_{\mu\to i}$, $\omega^t_{\mu\to i}$, $f^t_{\mu\to i}$, $g^t_{\mu\to i}$, $A^t_{i\to \mu}$, $b^t_{i\to \mu}$, $\hat{w}^t_{i\to \mu}$, $\hat{c}^t_{i\to \mu}$. This reductionist viewpoint is, much like the replica approach of section \ref{subsec:replica}, very characteristic of statistical physics.

\paragraph{Summary : State evolution equations}
We now regroup the SE equations derived in the previous paragraph:
\begin{align}
    \label{eq:SE}
    &\begin{cases}
         V_{\ell,k}^t=\int d\nu(\gamma,\tau,\pi)\gamma_{\ell,k}\left(\lambda \mathbb{I}_r +\hat{v}^t+\sum\limits_\kappa\sum\limits_{j} \hat{V}^{t-1}_{\kappa,j} \gamma_{\kappa,j} \right)^{-1}\\
        q^t_{\ell,k}=\int d\nu(\gamma,\tau,\pi) \gamma_{\ell, k} \left(\lambda \mathbb{I}_r +\hat{v}^t+\sum\limits_\kappa\sum\limits_{j} \hat{V}^{t-1}_{\kappa,j} \gamma_{\kappa,j} \right)^{-1}\\
    \qquad~~\left[\left(
    \sum\limits_\kappa\sum\limits_{j} (\tau_{\kappa,j} \hat{m}_{\kappa,j}^{t-1}+ \gamma_{\kappa,j}\hat{\theta}_{\kappa,j}^{t-1}\pi
    \right)^{\otimes 2}\!\!\!\!\!\!\!\!+ \sum\limits_\kappa\sum\limits_{j} \gamma_{\kappa,j}\hat{q}^{t-1}_{\kappa,j}
    \right]
    \left(\lambda \mathbb{I}_r +\hat{v}^t+\sum\limits_\kappa\sum\limits_{j} \hat{V}^{t-1}_{\kappa,j} \gamma_{\kappa,j} \right)^{-1}\\
    \theta^t_{\ell,k}=\int d\nu(\gamma,\tau,\pi) \gamma_{\ell,k } \left(\lambda \mathbb{I}_r +\hat{v}^t+\sum\limits_\kappa\sum\limits_{j} \hat{V}^{t-1}_{\kappa,j} \gamma_{\kappa,j} \right)^{-1} \left(
    \sum\limits_\kappa\sum\limits_{j} (\tau_{\kappa,j} \hat{m}_{\kappa,j}^{t-1}+ \gamma_{\kappa,j}\hat{\theta}_{\kappa,j}^{t-1}\pi
    \right)\pi^\top\\
    m^t_{\ell,k}=\int d\nu(\gamma,\tau,\pi) \tau_{\ell,k } \left(\lambda \mathbb{I}_r +\hat{v}^t+\sum\limits_\kappa\sum\limits_{j} \hat{V}^{t-1}_{\kappa,j} \gamma_{\kappa,j} \right)^{-1} \left(
    \sum\limits_\kappa\sum\limits_{j} (\tau_{\kappa,j} \hat{m}_{\kappa,j}^{t-1}+ \gamma_{\kappa,j}\hat{\theta}_{\kappa,j}^{t-1}\pi
    \right)\\
    v^t=\int d\nu(\gamma,\tau,\pi)  \left(\lambda \mathbb{I}_r +\hat{v}^t+\sum\limits_\kappa\sum\limits_{j} \hat{V}^{t-1}_{\kappa,j} \gamma_{\kappa,j} \right)^{-1}\\
    \qquad~~\left[\left(
    \sum\limits_\kappa\sum\limits_{j} (\tau_{\kappa,j} \hat{m}_{\kappa,j}^{t-1}+ \gamma_{\kappa,j}\hat{\theta}_{\kappa,j}^{t-1}\pi
    \right)^{\otimes 2}\!\!\!\!\!\!\!\!+ \sum\limits_\kappa\sum\limits_{j} \gamma_{\kappa,j}\hat{q}^{t-1}_{\kappa,j}
    \right]
    \left(\lambda \mathbb{I}_r +\hat{v}^t+\sum\limits_\kappa\sum\limits_{j} \hat{V}^{t-1}_{\kappa,j} \gamma_{\kappa,j} \right)^{-1}
    \end{cases}\\
    &\begin{cases}
        \hat{V}^t_{\ell,k}=\hat{\theta}^t_{\ell,k}(\theta^t_{\ell,k})^\top(q^t_{\ell,k})^{-1}-\alpha  \mathbb{E}_{c}\delta_{c_\ell,k}\mathbb{E}_{y,\Xi}(V^t_{\ell,k})^{-1}(\prox_\ell^c-(q^t_{\ell,k})^{\sfrac{1}{2}}\xi_\ell-m^t_{\ell,k})\xi_\ell^\top(q^t_{\ell,k})^{-\sfrac{1}{2}} \\
        \hat{q}^t_{\ell,k}=\alpha\mathbb{E}_{c}\delta_{c_\ell,k}\mathbb{E}_{y,\Xi}\left[(V^t_{\ell,k})^{-1}\left[
    \prox_\ell^c-(q^t_{\ell,k})^{\sfrac{1}{2}}\xi_\ell-m^t_{\ell,k}
    \right]^{\otimes 2}(V^t_{\ell,k})^{-1}
    \right]\\
    \hat{\theta}^t_{\ell,k}=\alpha \mathbb{E}_{c}\delta_{c_\ell,k}\mathbb{E}_{y,\Xi}(V^t_{\ell,k})^{-1}\left[
    \prox_\ell^c-(q^t_{\ell,k})^{\sfrac{1}{2}}\xi_\ell-m^t_{\ell,k}
    \right]\\
    \qquad~~\left(y_\ell-m^\star_{\ell,k}-(\theta^t_{\ell,k})^\top(q^t_{\ell,k})^{-\sfrac{1}{2}}\xi_\ell\right)^\top\left(\rho_{\ell,k}-(\theta^t_{\ell,k})^\top(q^t_{\ell,k})^{-1}\theta^t_{\ell,k}\right)^{-1}\\
    \hat{m}^t_{\ell,k}= \alpha\mathbb{E}_{c}\delta_{c_\ell,k}\mathbb{E}_{y,\Xi}(V^t_{\ell,k})^{-1}\left[
    \prox_\ell^c-(q^t_{\ell,k})^{\sfrac{1}{2}}\xi_\ell-m^t_{\ell,k}
    \right]\\
     \hat{v}^t=2\alpha \mathbb{E}_c\mathbb{E}_{y,\Xi}\partial_3\ell\left(y^c, \prox^c,  V^t_c,v^t,c\right)
    \end{cases}    
\end{align}
The astute reader will have recognized the replica SP equations \eqref{eq:intro:replica_SP_repeat} derived in section \ref{subsec:replica}, with the difference that in the dynamical SE equations \eqref{eq:SE}, the summary statistics bear time indices. This subsection has thus established that the equations \eqref{eq:intro:replica_SP_repeat} describe the summary statistics capturing the dynamics of GAMP iterations (\ref{alg:GAMP}), provided the relevant time indices are included. In particular, the equations \eqref{eq:intro:replica_SP_repeat} -- without time indices -- are verified by the fixed points of GAMP. The next subsection finally shows that the latter also correspond to critical (zero-gradient) points of the empirical landscape \eqref{eq:ERM}, i.e. fixed points of GD.

%% file: Sections_masked/GD_mapping.tex
\subsubsection{Fixed points of GAMP are fixed points of GD }

In this subsection, we finally show that fixed points of GAMP (\ref{alg:GAMP}) --as asymptotically described by the replica equations \eqref{eq:intro:replica_SP_repeat} derived in section \ref{subsec:replica}-- coincide with critical (zero gradient) points of the empirical ERM landscape \eqref{eq:ERM}. Let us depart from the fixed point equations of GAMP, which read

\begin{subnumcases}{\label{eq:state_fixed_pts_GAMP}}
        b_{i}^\infty=\frac{1}{\sqrt{d}}\sum\limits_{\ell=1}^L
\sum\limits_{\mu}x^\mu_{\ell i} f_{\ell,\mu}^\infty+A_i^\infty\hat{w}_i^\infty\label{eq:state_fixed_pts_GAMP_1}\\
\omega_{\ell,\mu}^\infty=\frac{1}{\sqrt{d}}\sum\limits_{i} x^\mu_{\ell, i} \hat{w}_{i}^\infty-\sum\limits_\kappa(V_\mu)_{\ell\kappa}^\infty f_{\kappa\mu}^\infty\label{eq:state_fixed_pts_GAMP_2}\\
f_{\ell,\mu}^\infty=\sum\limits_{\kappa=1}^L (V_{\mu}^\infty)^{-1}_{\ell\kappa}\left(\prox(y_\mu,\omega_{\mu}^\infty,V_{\mu}^\infty, \Gamma^\infty, c^\mu)-\omega_{\mu}^\infty\right)_\kappa\label{eq:state_fixed_pts_GAMP_3}\\
C^\infty=\frac{2}{d}\sum\limits_{\mu}\partial_3\ell\left(y_\mu, \prox(y_\mu,\omega_{\mu}^\infty,V_{\mu}^\infty, \Gamma^\infty, c^\mu), \Gamma^\infty, c^\mu\right)\label{eq:state_fixed_pts_GAMP_4}\\
\Gamma^\infty=\frac{1}{d}\sum\limits_{i}\hat{w}_{i}^\infty(\hat{w}_{i}^\infty)^\top\label{eq:state_fixed_pts_GAMP_5}\\
b_{i}^\infty=(\lambda\mathbb{I}_r+C^\infty+A_{i}^\infty) \hat{w}_{i}\label{eq:state_fixed_pts_GAMP_6}
    \end{subnumcases}

In \eqref{eq:state_fixed_pts_GAMP}, all GAMP variables are understood as the infinite-time iterates, evaluated after convergence. Combining \eqref{eq:state_fixed_pts_GAMP_1} with \eqref{eq:state_fixed_pts_GAMP_3} and \eqref{eq:state_fixed_pts_GAMP_6},the following identity can be constructed:
\begin{align}
\label{eq:zero_GD_expr_1}
    \frac{1}{\sqrt{d}}\sum\limits_{\ell,\kappa}
\sum\limits_{\mu}x^\mu_{\ell i}  (V_{\mu}^\infty)^{-1}_{\ell\kappa}\left(\omega_{\mu}^\infty-\prox(y_\mu,\omega_{\mu}^\infty,V_{\mu}^\infty, \Gamma^\infty, c^\mu)\right)_\kappa+(\lambda\mathbb{I}_r+C^\infty)\hat{w}_i^\infty=0
\end{align}
From the definition of the proximal \eqref{eq:prox_GAMP}, it further follows that
\begin{align}
\label{eq:prox_expr1}
    \sum\limits_\kappa (V_{\mu}^\infty)^{-1}_{\ell\kappa}\left(\omega_{\mu}^\infty-\prox(y_\mu,\omega_{\mu}^\infty,V_{\mu}^\infty, \Gamma^\infty, c^\mu)\right)_\kappa=\nabla_{X_\ell}\ell(y_\mu,\prox(y_\mu,\omega_{\mu}^\infty,V_{\mu}^\infty, \Gamma^\infty, c^\mu), \Gamma^\infty, c^\mu),
\end{align}
where the derivative bears over the $\ell-$th row of the second argument of the loss function $\ell$.
On the other hand, combining \eqref{eq:state_fixed_pts_GAMP_2} and \eqref{eq:state_fixed_pts_GAMP_3}, the resolvent $\prox(y_\mu,\omega_{\mu}^\infty,V_{\mu}^\infty, \Gamma^\infty, c^\mu)$ can also be expressed as
\begin{align}
\label{eq:prox_expr2}
    \prox(y_\mu,\omega_{\mu}^\infty,V_{\mu}^\infty, \Gamma^\infty, c^\mu)_\ell=\frac{1}{\sqrt{d}}\sum\limits_{i} x^\mu_{\ell, i} \hat{w}_{i}^\infty=\left(\frac{\x^\mu \hat{\w}^\infty}{\sqrt{d}}\right)_\ell.
\end{align}
Incorporating the two simplifications \eqref{eq:prox_expr1} and \eqref{eq:prox_expr2} into \eqref{eq:zero_GD_expr_1}, and further replacing $\Gamma, C$  by \eqref{eq:state_fixed_pts_GAMP_5} and \eqref{eq:state_fixed_pts_GAMP_6},yields
\begin{align}
    \sum\limits_\mu \sum\limits_{\ell} \frac{x^\mu_{\ell i}}{\sqrt{d}} \nabla_{X_\ell}\ell\left(y_\mu,\frac{\x^\mu \hat{\w}^\infty}{\sqrt{d}}, \frac{(\hat{\w}^\infty)^\top \hat{\w}^\infty}{d}, c^\mu \right)+\frac{2}{d}\sum\limits_d \partial_3 \ell\left(y_\mu,\frac{\x^\mu \hat{\w}^\infty}{\sqrt{d}}, \frac{(\hat{\w}^\infty)^\top \hat{\w}^\infty}{d}, c^\mu \right)\hat{w}_i^\infty+\lambda \hat{w}_i^\infty=0.
\end{align}
The astute reader might have noticed that the left-hand side corresponds, in fact, to a derivative, and can be more evocatively and compactly expressed as 
\begin{align}
    \nabla_{\w}\left[
    \sum\limits_\mu 
   \ell\left(y_\mu,\frac{\x^\mu \w}{\sqrt{d}}, \frac{\w^\top \w}{d}, c^\mu \right)+\frac{\lambda}{2}\lVert \w\lVert^2  \right]\Bigg|_{\hat{\w}^\infty}=0,
\end{align}
which is exactly a zero-gradient condition for the empirical risk \eqref{eq:ERM}. In other words, we established that fixed points $\hat{\w}^\infty$ of the GAMP algorithm \ref{alg:GAMP} also correspond to fixed points of GD, i.e. critical points of the empirical loss landscape \eqref{eq:ERM}.

\subsection{Summary}
Section \ref{sec: Derivation} presented a detailed derivation and discussion of a tight asymptotic characterization of the learning of the \seq, using techniques borrowed from statistical physics. In subsection \ref{subsec:replica}, we demonstrated how, using the replica method, a sharp characterization of the minimizer $\hat{\w}$ of the ERM \eqref{eq:ERM} could be reached in terms of a sufficient set of finite-dimensional summary statistics. These statistics are solutions of a system of self-consistent SP equations \eqref{eq:intro:replica_SP_repeat}. The analysis thus importantly enables the reduction of the original high-dimensional optimization problem into a set of equations in \emph{finite} dimensions.  Subsection \ref{subsec:SE} then provided an algorithmic viewpoint on the SP equations as the fixed point conditions of a GAMP algorithm. Furthermore, the latter were shown to also correspond to critical points of the ERM landscape -- i.e. fixed points of GD --. The study of the set of finite-dimensional SP equations thus offers an informative and insightful perspective on the ERM landscape, and affords a particularly powerful framework to analyze ML learning tasks in high dimensions. Techniques such as the replica method and GAMP --or variants thereof-- underlie a large number of statistical physics analyses of NNs \cite{gardner1989three, gardner1988optimal, sompolinsky1990learning, schwarze1993learning,mignacco2020role, cornacchia2023learning, aubin2020generalization, loureiro2021learning2, cui2023high, cui2024phase, gerace_generalisation_2020, loureiro2021learning,pesce2023gaussian,dAscoli2021OnTI,baldassi2016learning, Saglietti2021SolvableMF, cui2020large, ZavatoneVeth2022ContrastingRA, gabrie2018entropy, manoel2017multi, ichikawa2023dataset, krzakala2012probabilistic}.

%% file: Sections_masked/Phenomenology.tex
\subsection{Some insights and phenomenology}
\label{subsec:phenomeno}

\begin{figure}
    \centering
    \includegraphics[scale=0.56]{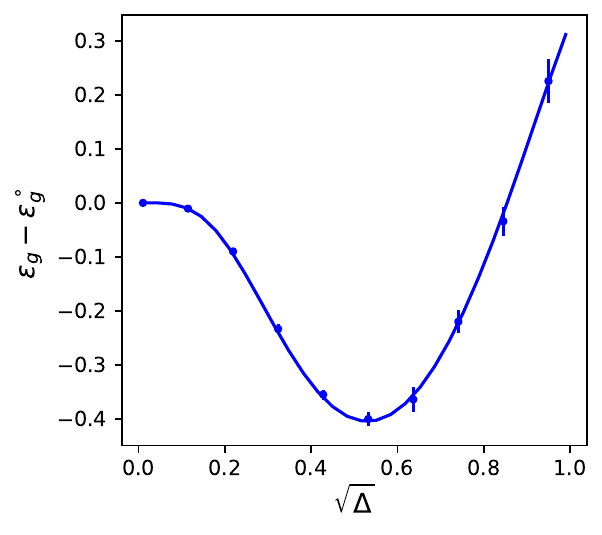}
    \includegraphics[scale=0.58]{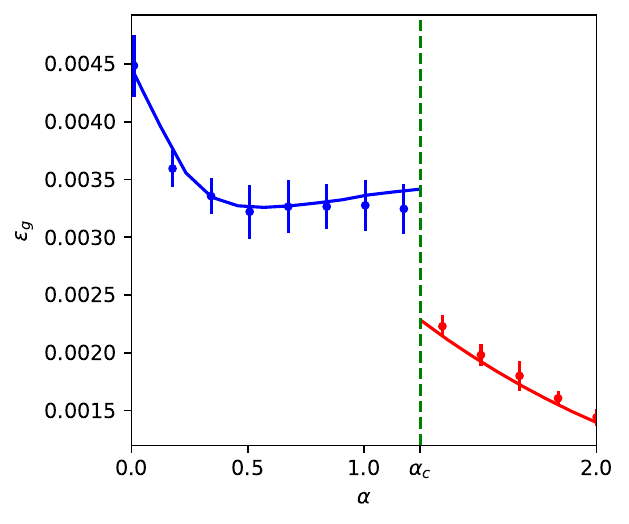}
    \caption{Example of learning curves obtained from the theoretical characterization \eqref{eq:intro:replica_SP_repeat} of section \ref{sec: Derivation} (solid lines), contrasted with numerical experiments (dots), in dimensions $d=700, 1000$, of the corresponding networks trained with the Pytorch \cite{paszke2019pytorch} implementation of the Adam \cite{kingma2014adam} optimizer. (left) Reproduced from \cite{cui2023high}. Denoising test error achieved by a DAE \eqref{eq:AE:model}, minus that achieved by a simple linear baseline $f_b(\x)=b\x$, as a function of the variance of the corrupting noise. (right) Reproduced from \cite{cui2024phase}. Test error achieved by a single layer attention \eqref{eq:Attention:student}. A first order phase transition happens at sample complexity $\alpha=\alpha_c$, signalling the learning by the model of a qualitatively new algorithmic mechanism. All the details can be found in the original works, and are not exhaustively reported here for conciseness. }
    \label{fig:examples}
\end{figure}

In this section, we turn to briefly reviewing some phenomenological insights that can be extracted from the analysis presented in section \ref{sec: Derivation}, providing very short accounts of the discussions reported in \cite{mignacco2020role, aubin2018committee, loureiro2021learning2, cornacchia2023learning, cui2023high, cui2024phase}. These very short sketches have the dual ambition of showcasing the power of this analytical approach, and of illustrating the breadth of the questions that can be explored therewith.

\paragraph{GLMs -- approaching Bayes optimality, and the role of regularization} 
We start this section by briefly reviewing some insights garnered for single-layer networks. For a perceptron target function, \cite{aubin2020generalization} find that a GLM trained with optimally-regularized logistic loss achieves a performance quantitatively very close to the Bayes-optimal performance. This conclusion is shared by \cite{cornacchia2023learning} in the more generic setting of multiclass classification with a cross-entropy loss. In the related context of classifying Gaussian mixture data, \cite{mignacco2020role} show that for a balanced binary mixture, the optimal regularization corresponds to the limit of infinite $\ell_2$ regularization $\lambda\to\infty$, and the corresponding classifiers then achieves Bayes-optimal error. Interestingly, this property holds for any twice-differentiable loss function with finite second derivative at the origin. In the case of mixtures with $K\ge 2$ clusters, again in the balanced case, \cite{loureiro2021learning2} also find that the Bayes-optimal error is approached for limit of infinite $\ell_2$ regularization $\lambda\to\infty$. This series of works bring valuable insight into how information-theoretically optimal performances can be approached with simple ERM procedures, when the regularization is carefully tuned.

\paragraph{AEs-- architecture and phase diagram} Consider a denoising AE (DAE) (see also \ref{subsec:DAE}). This already rather complex architecture comprises two components, namely a skip connection, and a bottleneck network. What respective roles do they play in the full architecture ? \cite{cui2023high} explore this question by analyzing the learning of these components when they are trained in isolation, as standalone denoiser networks. Surprisingly, the bottleneck network then learns to perform a simple linear algorithm -- namely Principal Component Analysis --. Thus, both components of an AE architecture learn linear algorithms when trained in isolation, but when jointly trained as components of the full architecture, they result in a non-linear, and better performing, model. \cite{ichikawa2023dataset} leverage a very similar replica analysis as the one presented in section \ref{sec: Derivation} -- although their model does not strictly fall under the \seq of section \ref{sec:seq-GLM}-- to analyze a linear Variational AE (VAE). They evidence a rich phase diagram, as a function of the sample complexity and hyperparameters of the VAE loss, delineating regions where the model generalizes and regions where it suffers from the posterior collapse phenomenon. 

\paragraph{Attention -- emergence of semantic learning }  To complete a supervised regression task, \cite{cui2024phase} show that a single attention layer can learn to implement two qualitatively different algorithmic mechanisms -- positional attention (with tokens attending to each other based on their respective positions) at small sample complexities, and semantic attention (with tokens attending to each other based on their meanings) past a critical sample complexity $\alpha_c$. The emergence of the semantic mechanism corresponds to a \textit{first-order} phase transition in statistical physics, and is accompanied by a sharp drop in test error. In fact, the two distinct mechanisms correspond to two distinct minima in the non-convex empirical loss landscape, with the positional (resp. semantic) minimum becoming global below (resp. beyond) $\alpha_c$.

%% file: Sections_masked/Perspectives.tex
\section{Perspectives}
\label{sec:perspectives}

Subsection \ref{subsec:phenomeno} concludes the review of statistical physics approaches to high-dimensional NNs, through the lens of the case study of the \seq. The tools and ideas presented in sections \ref{sec:seq-GLM} and \ref{sec: Derivation} delineate a very sizeable patch, though not the entirety, of the statistical physics of ML researchscape. To provide a somewhat more exhaustive picture of the field, we give in this conclusive section an overview of related research questions, and recent progress therein.  

\subsection{Connex questions}

This subsection gathers an overview of recent progress in adjoining research questions, that bear direct connection to the tight asymptotic exploration of high-dimensional neural networks using statistical physics ideas, as reported and reviewed in precedent sections. We also take the opportunity to delineate current challenges and directions.

\subsubsection{Data structure}

\begin{figure}
    \centering
    \includegraphics[scale=0.27]{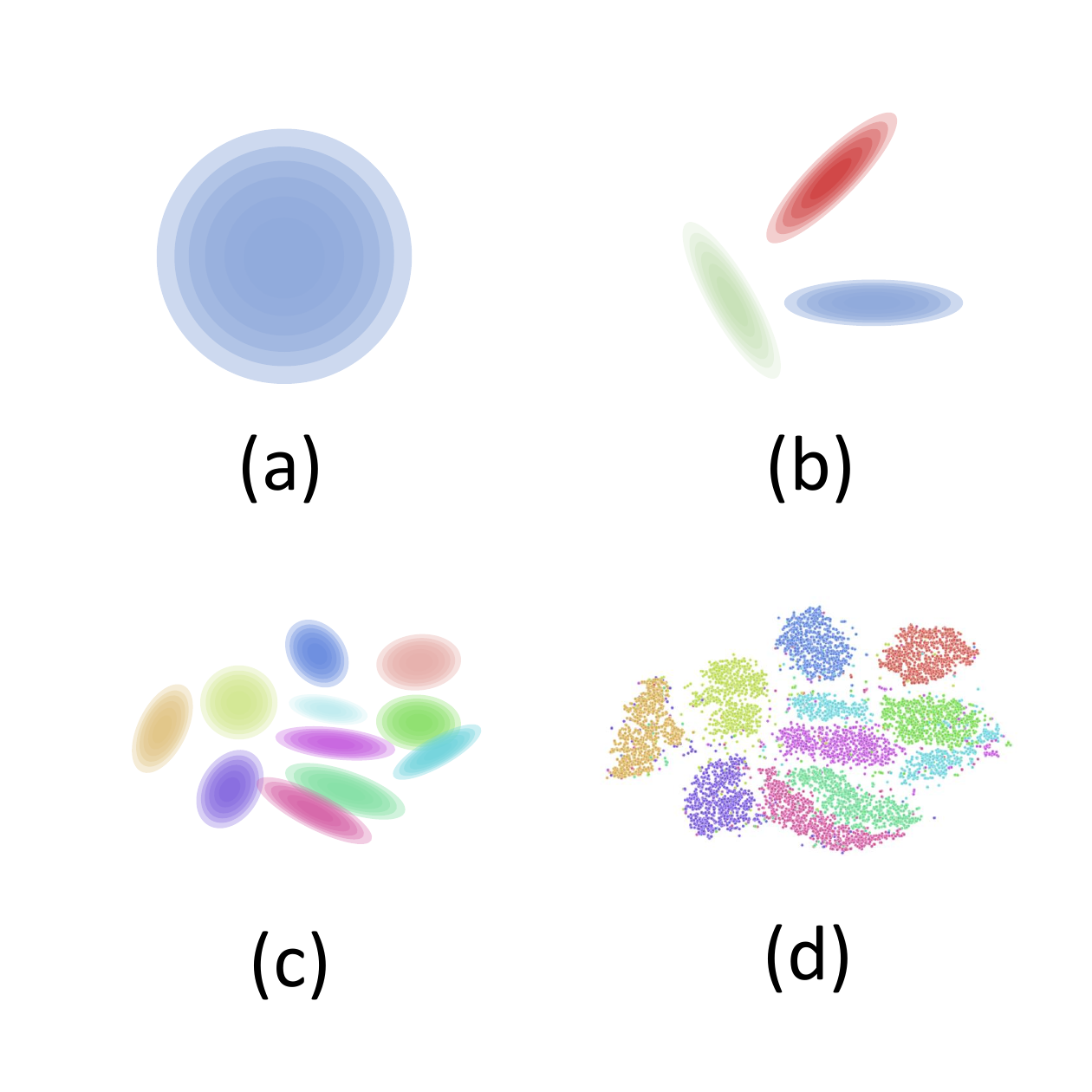}
    \includegraphics[scale=0.55]{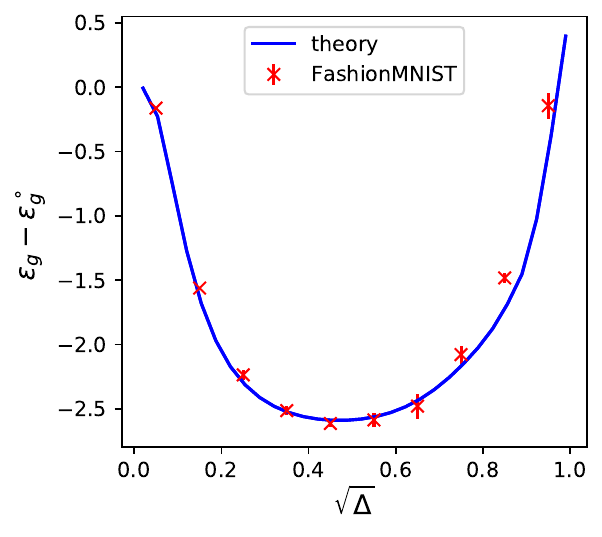}
    
    \caption{(left) Statistical physics studies typically leverage stylized data assumptions, such as Gaussian densities (a) or Gaussian mixture densities (b). In sharp contrast, real data ((d) represents a tSNE \cite{van2008visualizing} visualization of the MNIST \cite{lecun1998gradient} train set) displays much more intricate structure. A strategy employed in e.g. \cite{bordelon2021learning, loureiro2021learning, loureiro2021learning2, cui2023high, Refinetti2022TheDO} consists in considering surrogate analytical densities (c) with matching relevant statistics. This allows to obtain theoretical predictions capturing the learning curves of real data experiments, like denoising FashionMNIST \cite{xiao2017} with a DAE (right). The plot is reproduced from \cite{cui2023high}, and represents the test error achieved by the DAE, minus that achieved by a simple baseline $f_b(\x)=b\x$, see also caption of Fig.\,\ref{fig:examples}. }
    \label{fig:real_data}
\end{figure}

A large part of the versatility of the \seq framework expounded in section \ref{sec:seq-GLM} resides in its particularly flexible data model \eqref{eq:data_distrib}, which subsumes a large class of commonly employed data models \cite{aubin2018committee, mignacco2020role, cornacchia2023learning, loureiro2021learning2, cui2023high, cui2024phase}. While versatile in the diversity of tasks it can accommodate and model, this distribution remains however rather stylized.
It is in fact fair to say that, generally, in the building of a model of a ML task, the modeling of the data distribution poses a singular challenge. In contrast to NNs, which are by design mathematical constructs and can thus be formally modelled rather straightforwardly, it is to a large extent unclear how to model the distribution of real data. What indeed is the probability distribution of e.g. images of cats and dogs? Of natural language? Satisfyingly answering these interrogations is to a large extent an open question in ML theory. As of the end of the $2010$s, a sizeable part of the research effort in sharp asymptotic studies of ML methods was in fact set under the assumption of isotropic, unimodal Gaussian data \cite{aubin2018committee, aubin2020generalization, barbier2019optimal, zdeborova2016statistical, cornacchia2023learning, gardner1988optimal, sompolinsky1990learning}, or random orthogonal design \cite{kabashima2008inference, shinzato2008learning, shinzato2008perceptron}. Overcoming those somewhat restrictive assumptions to consider colored (or bimodal) Gaussian densities, was an endeavour initiated in e.g. \cite{mignacco2020role, goldt2020modeling, dAscoli2021OnTI}. Let us also mention the work of \cite{chung2018classification}, which addressed distributions of extended manifolds. Still, these more structured models remained highly stylized in nature. Further morphing these idealized densities into more \emph{realistic} distributions, is an essential step towards building \emph{realistic} models which can ultimately be descriptive of day-to-day ML practice. \\

A possible research agenda is to aim at identifying, for any given setting, the set of relevant statistical or structural descriptors of the data distribution which is relevant for the learning -- i.e. the \textit{universality class} of the data model. A real dataset can then be replaced, in the analysis, by a surrogate stylized data density \textit{with matching descriptors}, with the latter being tractable and thus amenable to further theoretical study. For instance, models for which these descriptors corresponds to the first two moments are regrouped under the umbrella of \textit{Gaussian universality}. Gaussian universality has been rigorously characterized in theoretically controlled settings in a large body of works \cite{goldt2022gaussian, montanari2022universality, hu2022universality, dandi2023universality, pesce2023gaussian}, and empirically observed for single-layer networks \cite{loureiro2021learning, bordelon2021learning,loureiro2021learning2}, kernels and wide networks \cite{Canatar2021SpectralBA, bordelon2020, cui2021generalization, cui2023error, bordelon2022self}, and AEs \cite{cui2023high, Refinetti2022TheDO}, making it possible to build theoretical descriptions that quantitatively capture the learning curves of experiments on real data sets, such as MNIST \cite{lecun1998gradient} (images of handwritten digits), FashionMNIST \cite{xiao2017} (images of clothing items) or CIFAR \cite{krizhevsky2009learning} (images of animals, means of transport...). An example is provided in Fig.\,\ref{fig:real_data}, right, for denoising using an AE \cite{cui2023high}. Most of these data models are comprised in the \seq \eqref{eq:data_distrib}.
We however know that there exist settings where more complicated data universalities -- beyond Gaussian universality -- hold. For instance, the universality class of manifold classification \cite{chung2018classification} relies on more intricate geometric descriptors, rather than simply the first two statistical moments. It is a fair observation that, in the current state of research, we lack a principled way of identifying the relevant data universality class -- if any-- for a given setting, architecture, and scaling. Before closing this discussion, let us mention an alternative path, trodden in e.g. \cite{li2021statistical, ariosto2022statistical, bordelon2020}, which consists in formulating effective theories directly in terms of $n\times n$ kernel matrices, thereby allowing one to seamlessly plug-in real data kernels -- but on the other hand not allowing fully asymptotic descriptions as $n\to\infty$.

\subsubsection{Infinite and extensive-width scalings}

The \seq of section \ref{sec:seq-GLM} models high-dimensional neural network architectures with $r=\Theta_d(1)$ hidden units. Understanding NN models with wider architectures, i.e. along the horizontal direction of the diagram of Fig.\,\ref{fig:zoo}, is essential in probing the learning of overparametrized model -- arguably a core research question in ML theory. While, as we reviewed in subsection \ref{subsec:modelscape}, this line of research is rather well trodden for fully-connected models, the bottom-right sector of Fig.\,\ref{fig:zoo} remains to a large extent uncharted, with the exception of a few works such as \cite{NGUYEN2019BenefitsOJ, Nguyen2021AnalysisOF, bordelon2024infinite}. Even on the level of fully-connected models, the exact asymptotic analysis of the learning of extensive width ($r=\Theta_d(d)$) networks still eludes current technical know-how. Existing tight asymptotic studies rely on alternative training protocols, such as freezing all the layers but the readout \cite{gerace_generalisation_2020, schroder2023deterministic, bosch2023precise, schroder2024asymptotics, ZavatoneVeth2022ContrastingRA, zavatone2023learning}, Bayesian learning \cite{ZavatoneVeth2022ContrastingRA, li2021statistical, ariosto2022statistical, cui2023bayes, hanin2024bayesian, hanin2022bayesian, tiberi2024dissecting, van2024coding, maillard2024bayes}, or only performing a single gradient step \cite{moniri2023theory, cui2024asymptotics}.

\subsubsection{Dynamics}
This review focuses on \textit{static} statistical analyses of the minimizers (and properties thereof) of ERM problems, irrespective of how they can be reached, or not, by optimizers such as (stochastic) gradient descent. The study of the dynamics of learning in neural networks constitutes a large and important part of ML theory, with many works also leveraging the statistical physics toolbox. The dynamics of full-batch (stochastic) gradient descent can be tightly captured with Dynamical Mean-Field Theory (DMFT) (see e.g. \cite{cugliandolo2023recent} for a review) for single-layer networks \cite{mignacco2020dynamical, mignacco2021stochasticity, mignacco2022effective, gerbelot2024rigorous, celentano2021high, collins2024high}, wide multi-layer neural networks \cite{bordelon2022self, bordelon2023dynamics, bordelon2023depthwise, bordelon2022influence}, and transformers \cite{bordelon2024infinite}. Learning with online stochastic gradient descent has also been characterized in a long line of works \cite{goldt2020modeling, goldt2019dynamics, Refinetti2021ClassifyingHG, refinetti2021align, refinetti2023neural, patel2023rl, straat2019line} for two-layer network with a finite number of hidden units $r=\Theta_d(1)$, impulsed by the seminal works of \cite{biehl1995learning, saad.solla_1995_line, saad1995exact, saad1996learning, riegler1995line}. The dynamics of deep, but linear networks has been explored in \cite{saxe2013exact, saxe2019mathematical, braun2022exact, atanasov2021neural, tarmoun2021understanding, arora2018, fukumizu1998effect}.  Let us also mention that the online stochastic gradient dynamics have also been characterized in AE architectures in \cite{Refinetti2022TheDO}.

\subsection*{Conclusions}
The present manuscript reviews recent progress in the tight asymptotic analysis of neural networks trained with empirical risk minimization, in the limit of comparably large data dimension $d$ and number of training samples $n$, and finite number of hidden units $r=\Theta_d(1)$. Introducing and using a versatile class of models -- the \seq-- as a case study, we unify and present the analyses reported in many previous works. This allows us to review central techniques in the field of statistical physics of ML, such as the replica method and approximate message passing algorithms. We finally present an overview of the current state of the field, and of the adjoining researchscape, outlining connex topics and modern challenges. This review should be a useful primer for machine learning theoreticians curious of statistical physics approaches; it should also be of value to statistical physicists interested in the transfer of such ideas to the study of neural networks.

\subsection*{Acknowledgements}
The author acknowledges support from the Center of Mathematical Sciences and Applications (CMSA) of Harvard University and from the Swiss National Science Foundation grant SNFS SMArtNet (grant number 212049).
This review is based on parts of the author's PhD dissertation \cite{cui2024topics}, performed in the Statistical Physics of Computation laboratory in EPFL, under the guidance of Lenka Zdeborová. The author further thanks Bruno Loureiro, Lenka Zdeborová, C\'edric Koller, Fabrizio Boncoraglio and Emanuele Troiani for their valuable feedback on this manuscript.

%% file: Sections_masked/BP.tex
\section{Derivation of GAMP from Belief Propagation}
\label{app: BP}

\subsection{BP iterations}
We detail in this appendix how the rBP and GAMP algorithm \ref{alg:BP}, \ref{alg:GAMP} can be obtained from the iterations of the \textit{Belief-Propagation algorithm} (BP). Historically, the BP algorithm was independently derived in many seminal works, in the context of magnetism \cite{ pearl:88}, error correcting codes \cite{gallager1962low}, and statistical inference \cite{pearl2022reverend}, see also \cite{Mzard2009InformationPA} for a review. In fact, rBP \ref{alg:BP} and GAMP \ref{alg:GAMP} correspond to asymptotically relaxed versions of BP, in the sense that they provide simplified and easier implementable, but asymptotically equivalent, iterations. We here make this equivalence explicit. For the factor graph of Fig.\,\ref{fig:Graphical_model}, the BP algorithm reads

\begin{algorithm}[H]
\caption{BP}
\label{alg:MP}
\begin{algorithmic}
\STATE \textbf{Inputs} : $\mathcal{D
}=\{\x^\mu, y^\mu,c^\mu\}_{\mu=1}^n $
\FOR{$t\le t_{\max}$}
\STATE $\forall 1\le i \le d, 1\le \mu\le n, ~~m^{t+1}_{i\to\mu }(w_i)=\frac{e^{-\frac{\beta\lambda}{2}\lVert w_i\lVert^2 }\prod\limits_{\nu\ne \mu}\hat{m}^t_{\nu \to i}(w_i)}{\int dz e^{-\frac{\beta\lambda}{2}\lVert z\lVert^2 }\prod\limits_{\nu\ne \mu}\hat{m}^t_{\nu \to i}(z)}$
\STATE $\forall 1\le i \le d, 1\le \mu\le n, ~~\hat{m}^t_{\mu\to i }(w_i)=\frac{\int \prod\limits_{j\ne i} dw_j \prod\limits_{j\ne i} m^t_{j\to \mu}(w_j) e^{-\beta \ell(y^\mu, \sfrac{\x^\mu \w}{\sqrt{d}}, \sfrac{\w^\top \w}{d}, c^\mu)}}{\int d\w \prod\limits_{j\ne i}  m^t_{j\to \mu}(w_j) e^{-\beta \ell(y^\mu, \sfrac{\x^\mu \w}{\sqrt{d}}, \sfrac{\w^\top \w}{d}, c^\mu)}}$
\ENDFOR
\STATE Compute the marginals $\forall 1\le i\le d ,~~ m_i^{t_{\max}}(w_i)=\frac{\prod\limits_\mu \hat{m}^{t_{\max}}_{\mu\to i}(w_i)}{\int dz \prod\limits_\mu \hat{m}^{t_{\max}}_{\mu\to i}(z)}$
\RETURN $\{m^{t_{\max}}_i(w_i)\}_{i=1}^d$
\end{algorithmic}
\end{algorithm}
The algorithms iterates over variables $m_{i\to\mu}^t, \hat{m}_{\mu\to i}^t$ (the eponymous \textit{beliefs},or messages) which correspond to probability distribution over $\R^r$, and outputs the marginal probability $m_i$ over the $i-$th row $w_i\in\R^r$ of the trained weights $\hat{\w}$. Because of the complex nature of the iterates, this algorithmic scheme proves in many extents unwieldy, whence the incentive to find a simpler formulation, involving \textit{scalar, vector, or matrix} -- rather than probability distribution-- variables : rBP \ref{alg:BP} and GAMP \ref{alg:GAMP}. In the following, we detail this reduction.

\subsection{Equivalence BP/rBP}
\subsubsection{Relaxing BP}
\paragraph{Simplification of $\hat{m}_{\mu\to i}^t$}
Le us first consider the factor beliefs $\hat{m}_{\mu\to i}^t$. The starting point is to realize that, under the usual BP assumption that all incoming messages $m_{j\to\mu}^t$ are statistically independent, the variable $z=\sfrac{x^\mu\w}{\sqrt{d}}\in \R^{L\times r}$ is Gaussian from the CLT. Each of its rows $z_\ell\in\R^r$ has mean and joint covariance
\begin{align}
    &\mathbb{E}[z_\ell|w_i]=\underbrace{\frac{1}{\sqrt{d}}\sum\limits_{1\le j\ne i\le d} (x^\mu_\ell)_j \hat{w}_{j\to\mu}^t}_{\equiv (\omega^t_{\mu\to i})_\ell}+ \frac{1}{\sqrt{d}}(x^\mu_\ell)_i  w_i, \notag\\
    &\mathbb{C}[z_\ell, z_\kappa|w_i]=\underbrace{\frac{1}{d}\sum\limits_{1\le j\ne i\le d} (x^\mu_\ell)_j(x^\mu_\kappa)_j \hat{c}_{j\to\mu}^t}_{\equiv (V^t_{\mu\to i})_{\ell\kappa}}.
\end{align}
The expectation bears over the variables $w_j\sim m^t_{j\to\mu}$, for $1\le j\ne i \le d$.
We introduced the rBP variables $\omega^t_{\mu\to i} \in\R^{L\times r}$ and $V^t_{\mu\to i}\in\R^{Lr\times Lr}$ (viewed as a $L\times L$ block matrix with blocks $(V^t_{\mu\to i})_{\ell\kappa}\in\R^{r\times r}$.  We further defined $\hat{w}^t_{j\to\mu}$ and $\hat{c}^t_{j\to\mu}$ as the mean and variance of the message $m^t_{j\to\mu}$, namely
\begin{align}
\label{app:what-c}
    \hat{w}^t_{j\to\mu}=\int dw m^t_{j\to\mu}(w) w, && \hat{c}^t_{j\to\mu}=\int dw m^t_{j\to\mu}(w) w^2-(\hat{w}^t_{j\to\mu})^2.
\end{align}
Let us further observe that the variable $\sfrac{\w^\top \w}{d}$ concentrates to 
\begin{align}
   \sfrac{\w^\top \w}{d}=\underbrace{\frac{1}{d}\sum\limits_{1\le j\ne i \le d} (\hat{w}^t_{j\to \mu})(\hat{w}^t_{j\to \mu})^\top}_{\equiv \Gamma^t_{\mu\to i}} +\underbrace{\frac{1}{d}\sum\limits_{1\le j\ne i \le d} \hat{c}^t_{j\to\mu}}_{\equiv \Tilde{\Gamma}^t_{\mu\to i}}+\frac{1}{d}w_iw_i^\top.
\end{align}

In order to reach a compact expression, let us introduce the shorthand $\tau(w_i)\equiv\in\R^{L\times r}$ with rows $\tau(w_i)_\ell=(x^\mu_\ell)_iw_i\in\R^r$. The message $\hat{m}_{\mu\to i}^t$ can be now compactly rewritten as
\begin{align}
\label{app:hatm1}
    \hat{m}_{\mu\to i}^t(w_i)\propto\int dze^{-\frac{1}{2}\sum\limits_{1\le \ell,\kappa\le L}(z- \omega^t_{\mu\to i}-\frac{1}{\sqrt{d}}\tau(w_i))_\ell^\top (V^t_{\mu\to i})^{-1}_{\ell\kappa}(z- \omega^t_{\mu\to i}-\frac{1}{\sqrt{d}}\tau(w_i))_\kappa}e^{-\beta \ell(y^\mu, z, \Gamma^t_{\mu\to i}+ \Tilde{\Gamma}^t_{\mu\to i}+\sfrac{w_iw_i^\top}{d}, c^\mu)}.
\end{align}
For the sake of conciseness, we did not explicitly write the normalization factor, subsumed in the proportionality symbol $\propto$. Let us introduce the density associated to the partition function
\begin{align}
\label{app:Z_b}
    Z_\beta(y, \omega, V, \Gamma, c)\equiv \int dz \frac{1}{\sqrt{\det(2\pi V)}}e^{-\frac{1}{2}\sum\limits_{1\le \ell,\kappa\le L}(z-\omega)_\ell^\top V^{-1}_{\ell\kappa}(z-\omega)_\kappa}e^{-\beta\ell(y,z, \Gamma,c)},
\end{align}
and by brackets $\langle\cdot \rangle$ the expectation with respect to $Z_\beta(y^\mu,\omega^t_{\mu\to i}, V^t_{\mu\to i},\Gamma^t_{\mu\to i}+\tilde{\Gamma}^t_{\mu\to i}, c^\mu)$. Then, expanding the exponent of $ \hat{m}_{\mu\to i}^t(w_i)$ \eqref{app:hatm1},
\begin{align}
\label{qpp:mhat2}
     \hat{m}_{\mu\to i}^t(w_i)\propto& \sum\limits_{\ell\kappa}\frac{1}{\sqrt{d}}\tau(w_i)_\ell^\top\underbrace{(V^t_{\mu\to i})^{-1}_{\ell\kappa}\langle (z-\omega^t_{\mu\to i})_\kappa\rangle}_{\equiv f^t_{\kappa,\mu\to i}  } +\frac{1}{2d}\left(\sum\limits_{\ell\kappa}\tau(w_i)_\ell^\top(V^t_{\mu\to i})^{-1}_{\ell\kappa}\langle (z-\omega^t_{\mu\to i})_\kappa\rangle\right)^2
     \notag\\
     &-\frac{1}{2d}\sum\limits_{\ell\kappa}\tau(w_i)_\ell^\top(V^t_{\mu\to i})^{-1}_{\ell\kappa}\tau(w_i)_\kappa-\beta \Tr[ \underbrace{\langle\partial_3\ell(y^\mu, z, \Gamma^t_{\mu\to i}+ \Tilde{\Gamma}^t_{\mu\to i}, c^\mu)\rangle}_{\equiv \eta_{\mu\to i}^t}\frac{w_iw_i^\top}{d}].
\end{align}
We introduced the rBP variables $f^t_{\mu\to i}\in \R^{L\times r}$ and $\eta^t_{\mu\to i} \in \R^{r\times r}$. Let us further introduce the variance $g^t_{\mu\to i}\in \R^{Lr\times Lr}$, seen as a $L\times L$ block matrix with components
\begin{align}
    g^t_{\ell\kappa,\mu\to i} \equiv \nabla_{\omega_\kappa} f^t_{\ell,\mu\to i} =-(V^t_{\mu\to i})^{-1}_{\ell\kappa}-f^t_{\ell,\mu\to i }(f^t_{\kappa, \mu\to i} )^\top +\left\langle \sum\limits_{\rho\sigma} (V^t_{\mu\to i})^{-1}_{\ell\rho}(z-\omega^t_{\mu\to i})_\rho
    (z-\omega^t_{\mu\to i})^\top_\sigma  (V^t_{\mu\to i})^{-1}_{\sigma\kappa}\right\rangle,
\end{align}
so that \eqref{qpp:mhat2} can be rewritten in more compact fashion as
\begin{align}
    \hat{m}_{\mu\to i}^t(w_i)&\propto \frac{1}{\sqrt{d}}\sum\limits_\ell\tau(w_i)_\ell f^t_{\ell, \mu\to i}+\frac{1}{2d} \sum\limits_{\ell,\kappa}\tau(w_i)_\ell^\top g^t_{\ell\kappa,\mu\to i}\tau(w_i)_\kappa +\frac{1}{2d}\sum\limits_{\ell,\kappa}\tau(w_i)^\top_\ell f^t_{\ell,\mu\to i}(f^t_{\kappa, \mu\to i})^\top \tau(w_i)_\kappa\notag\\
    &\qquad -\beta \Tr[\eta^t_{\mu\to i}\frac{w_iw_i^\top}{d}]\notag \\
    &\propto e^{\frac{1}{\sqrt{d}}\sum\limits_{\ell}\tau(w_i)_\ell f^t_{\ell \mu\to i}+\frac{1}{2d}\sum\limits_{\ell,\kappa} \tau(w_i)_\ell^\top g^t_{\ell\kappa,\mu\to i}\tau(w_i)_\kappa -\beta \Tr[\eta^t_{\mu\to i}\frac{w_iw_i^\top}{d}]}\notag\\
    &\propto e^{\frac{1}{\sqrt{d}}\sum\limits_\ell (x^\mu_\ell)_i w_i^\top f^t_{\ell,\mu\to i}+\frac{1}{2d}\sum\limits_{\ell,\kappa} (x^\mu_\ell)_i(x^\mu_\kappa)_i w_i^\top g^t_{\ell\kappa,\mu\to i}w_i-\frac{\beta}{d}w_i^\top\eta^t_{\mu\to i}w_i}.
\end{align}

\paragraph{Simplication of $m^t_{i\to\mu}$} We now turn to examine the variable messages $m^t_{i\to\mu}$. Again keeping the normalization implicit:
\begin{align}
    \label{app:m1}
    m^t_{\mu\to i}(w_i)&\propto e^{w_i^\top \left(\underbrace{\frac{1}{\sqrt{d}}\sum\limits_{\nu\ne \mu} \sum\limits_\ell (x^\nu_\ell)_i f^t_{\ell,\nu\to i}}_{\equiv b^t_{i\to\mu}}\right)-\frac{1}{2}w_i^\top \left(\underbrace{-
    \frac{1}{d}\sum\limits_{\nu\ne \mu}
    \sum\limits_{\ell,\kappa} (x^\nu_\ell)_i(x^\nu_\kappa)_i  g^t_{\ell\kappa,\nu\to i}}_{\equiv A^t_{i\to\mu}}+\underbrace{\frac{2}{d}\sum\limits_{\nu\ne \mu}\beta\eta^t_{\nu\to i}}_{\equiv C^t_{i\to\mu}}
    \right)w_i-\frac{\beta\lambda}{2}w_i^\top w_i}.
\end{align}
in other words, $ m^t_{\mu\to i}$ is a Gaussian dstribution.
One is now in a position to close the equations on $\hat{w}^t_{i\to\mu}$ and $\hat{c}^t_{i\to\mu}$ \eqref{app:what-c}, leading to 
\begin{align}
    \hat{w}^t_{i\to\mu}=\left(\beta \lambda \mathbb{I}_r+A^t_{i\to\mu}+C^t_{i\to\mu}\right)^{-1} b^t_{i\to\mu}, && \hat{c}^t_{i\to\mu}=\left(\beta \lambda \mathbb{I}_r+A^t_{i\to\mu}+C^t_{i\to\mu}\right)^{-1}
\end{align}

\subsubsection{Summary of rBP equations}
We summarize all relaxed BP  for $\beta \to 0$

\begin{algorithm}[H]
\caption{rBP at temperature $\sfrac{1}{\beta}$}
\label{alg:beta-rBP}
\begin{algorithmic}
\STATE \textbf{Inputs} : $\mathcal{D
}=\{\x^\mu, y^\mu,c^\mu\}_{\mu=1}^n $
\STATE \textbf{Initialize} $ \forall 1\le \mu\le n,~~1\le i\le d,~~\hat{w}^0_{i\to \mu}=0_r, \hat{c}^0_{i\to \mu}=\mathbb{I}_{r}, \{f_{\ell\mu\to i}^0=0_r\}_{\ell=1}^L$
\FOR{$t\le t_{\max}$}
\STATE $\forall 1\le \ell,\kappa\le L,1\le \mu\le n, 1\le i \le d,~~(V_{\mu\to i }^t)_{\ell\kappa}=\frac{1}{d}\sum\limits_{j\ne i} (x^\mu_{\ell j})(x^\mu_{\kappa j}) \hat{c}^t_{j\to \mu} $
\STATE $1\le \mu\le n, 1\le i \le d,~~\Gamma_{\mu\to i}^t=\frac{1}{d}\sum\limits_{j\ne i}\hat{w}^t_{j\to\mu}(\hat{w}^t_{j\to\mu})^\top$
\STATE $1\le \mu\le n, 1\le i \le d,~~\tilde{\Gamma}_{\mu\to i}^t=\frac{1}{d}\sum\limits_{j\ne i}\hat{c}^t_{j\to\mu}$
\STATE $\forall 1\le \ell,1\le \mu\le n, 1\le i \le d, ~~ \omega_{\ell,\mu\to i}^t=\frac{1}{\sqrt{d}}\sum\limits_{j\ne i} x^\mu_{\ell, j} \hat{w}_{j\to \mu}^t$
\STATE $\forall 1\le \ell,1\le \mu\le n, 1\le i \le d,$$ 
f^t_{\ell,\mu\to i}=\left\langle\sum\limits_\kappa (V_{\mu\to i}^t)^{-1}_{\ell\kappa}\left(z-\omega_{\mu\to i}^t\right)_\kappa \right\rangle  $
\STATE $\forall 1\le \mu\le n, 1\le i \le d,~~\eta^t_{\mu\to i}=\left\langle \partial_3\ell\left(y_\mu, z, \Gamma^t_{\mu\to i}+\tilde{\Gamma}^t_{\mu\to i}, c^\mu\right)\right\rangle$
\STATE $\forall 1\le \ell,\kappa\le L,1\le \mu\le n, 1\le i \le d,~~
g_{\ell\kappa, \mu\to i}^t=\nabla_{\omega_\kappa}f_{\ell, \mu\to i}^t
$
\STATE $\forall 1\le \mu\le n, 1\le i \le d,~~ A^t_{i\to\mu}=-\frac{1}{d}\sum\limits_{\ell,\kappa=1}^L \sum\limits_{\nu\ne \mu}(x^\nu_{\ell i})(x^\nu_{\kappa i})g_{\ell\kappa, \nu\to i}^t$
\STATE $\forall 1\le \mu\le n, 1\le i \le d,~~ C^t_{i\to\mu}=\frac{2\beta}{d}\sum\limits_{\nu\ne \mu}\eta_{ \nu\to i}^t$
\STATE $\forall 1\le \mu\le n, 1\le i \le d,~~b^t_{i\to\mu}=\frac{1}{\sqrt{d}}\sum\limits_{\ell=1}^L
\sum\limits_{\nu\ne \mu}x^\nu_{\ell i} f_{\ell,\nu\to i}^t$
\STATE $\forall 1\le \mu\le n, 1\le i \le d,~~\hat{w}^{t+1}_{i\to\mu}=(\beta\lambda\mathbb{I}_r+C^t_{i\to\mu}+A^t_{i\to\mu})^{-1}b_{i\to\mu}^t$
\STATE $\forall 1\le \mu\le n, 1\le i \le d,~~\hat{c}^{t+1}_{i\to\mu}=(\beta\lambda\mathbb{I}_r+C^t_{i\to\mu}+A^t_{i\to\mu})^{-1}$
\ENDFOR
\STATE
\RETURN Estimator $\hat{\w}$
\end{algorithmic}
\end{algorithm}
We remind that the bracket notation $\langle \cdot\rangle$ denotes the expectation with respect to the measure with partition function $Z_\beta(y^\mu,\omega^t_{\mu\to i}, V^t_{\mu\to i},\Gamma^t_{\mu\to i}+\tilde{\Gamma}^t_{\mu\to i}, c^\mu)$ \eqref{app:Z_b}. We are now in a position to take the zero-temperature $\beta \to\infty$ limit. Rescaling the iterates as
\begin{align}
    &\frac{1}{\beta}V^t_{\mu\to i} \leftarrow V^t_{\mu\to i}, &&\frac{1}{\beta}\tilde{\Gamma}^t_{\mu\to i} \leftarrow \tilde{\Gamma}^t_{\mu\to i}, && \frac{1}{\beta}\hat{c}^t_{i\to \mu}\leftarrow \hat{c}^t_{i\to \mu},\notag\\
    & \beta f^t_{\mu\to i} \leftarrow f^t_{\mu\to i}, &&\beta g^t_{\mu\to i} \leftarrow g^t_{\mu\to i}, &&\beta b^t_{i\to \mu}\leftarrow b^t_{i\to \mu},\notag \\
    &\beta C^t_{i\to \mu}\leftarrow C^t_{i\to \mu}, && \beta A^t_{i\to \mu}\leftarrow A^t_{i\to \mu}.
\end{align}
This allows to greatly simplify the $\beta-$rBP iterations \ref{alg:beta-rBP}. First, note that $\tilde{\Gamma}^t_{\mu\to i}=\Theta_d(\sfrac{1}{\beta})$ is always negligible compared to $\Gamma^t_{\mu\to i}=\Theta_d(1)$, and can be removed as an iterate. Secondly, the measure $Z_\beta(y^\mu,\omega^t_{\mu\to i}, V^t_{\mu\to i},\Gamma^t_{\mu\to i}, c^\mu)$ concentrates, and 
\begin{align}
    \langle z\rangle =\underset{X\in\R^{L\times r}}{\mathrm{arginf}}\Bigg\{ 
\frac{1}{2} \sum\limits_{\ell,\kappa}(X-\omega^t_{\mu\to i})_\ell(V^t_{\mu\to i})^{-1}_{\ell\kappa}(X-\omega^t_{\mu\to i})_\kappa
     +\ell(y^\mu,X,\Gamma^t_{\mu\to i}, c^\mu)
     \Bigg\}\equiv \prox(y^\mu,\omega^t_{\mu\to i}, V^t_{\mu\to i},\Gamma^t_{\mu\to i}, c^\mu).
\end{align}
By the same token, 
\begin{align}
    \left\langle \partial_3\ell\left(y_\mu, z, \Gamma^t_{\mu\to i}+\tilde{\Gamma}^t_{\mu\to i}, c^\mu\right)\right\rangle=
    \partial_3\ell\left(y_\mu, \prox(y^\mu,\omega^t_{\mu\to i}, V^t_{\mu\to i},\Gamma^t_{\mu\to i}, c^\mu), \Gamma^t_{\mu\to i}, c^\mu\right).
\end{align}
The zero-temperature rBP iterations thus read

\begin{algorithm}[H]
\caption{rBP}
\begin{algorithmic}
\STATE \textbf{Inputs} : $\mathcal{D
}=\{\x^\mu, y^\mu,c^\mu\}_{\mu=1}^n $
\STATE \textbf{Initialize} $ \forall 1\le \mu\le n,~~1\le i\le d,~~\hat{w}^0_{i\to \mu}=0_r, \hat{c}^0_{i\to \mu}=\mathbb{I}_{r}$
\FOR{$t\le t_{\max}$}
\STATE $\forall 1\le \ell,\kappa\le L,1\le \mu\le n, 1\le i \le d,~~(V_{\mu\to i }^t)_{\ell\kappa}=\frac{1}{d}\sum\limits_{j\ne i} (x^\mu_{\ell j})(x^\mu_{\kappa j}) \hat{c}^t_{j\to \mu} $
\STATE $1\le \mu\le n, 1\le i \le d,~~\Gamma_{\mu\to i}^t=\frac{1}{d}\sum\limits_{j\ne i}\hat{w}^t_{j\to\mu}(\hat{w}^t_{j\to\mu})^\top$
\STATE $\forall 1\le \ell,1\le \mu\le n, 1\le i \le d, ~~ \omega_{\ell,\mu\to i}^t=\frac{1}{\sqrt{d}}\sum\limits_{j\ne i} x^\mu_{\ell, j} \hat{w}_{j\to \mu}^t$
\STATE $\forall 1\le \ell,1\le \mu\le n, 1\le i \le d,$$ 
f^t_{\ell,\mu\to i}=\sum\limits_\kappa (V_{\mu\to i}^t)^{-1}_{\ell\kappa}\left(\prox(y_\mu,\omega^t_{\mu\to i},V^t_{\mu\to i}, \Gamma^t_{\mu\to i}, c^\mu)-\omega_{\mu\to i}^t\right)_\kappa  $
\STATE $\forall 1\le \mu\le n, 1\le i \le d,~~\eta^t_{\mu\to i}=\partial_3\ell\left(y_\mu, \prox(y_\mu,\omega^t_{\mu\to i},V^t_{\mu\to i}, \Gamma^t_{\mu\to i}, c^\mu), \Gamma^t_{\mu\to i}, c^\mu\right)$
\STATE $\forall 1\le \ell,\kappa\le L,1\le \mu\le n, 1\le i \le d,~~
g_{\ell\kappa, \mu\to i}^t=\nabla_{\omega_\kappa}f_{\ell, \mu\to i}^t
$
\STATE $\forall 1\le \mu\le n, 1\le i \le d,~~ A^t_{i\to\mu}=-\frac{1}{d}\sum\limits_{\ell,\kappa=1}^L \sum\limits_{\nu\ne \mu}(x^\nu_{\ell i})(x^\nu_{\kappa i})g_{\ell\kappa, \nu\to i}^t$
\STATE $\forall 1\le \mu\le n, 1\le i \le d,~~ C^t_{i\to\mu}=\frac{2}{d}\sum\limits_{\nu\ne \mu}\eta_{ \nu\to i}^t$
\STATE $\forall 1\le \mu\le n, 1\le i \le d,~~b^t_{i\to\mu}=\frac{1}{\sqrt{d}}\sum\limits_{\ell=1}^L
\sum\limits_{\nu\ne \mu}x^\nu_{\ell i} f_{\ell,\nu\to i}^t$
\STATE $\forall 1\le \mu\le n, 1\le i \le d,~~\hat{w}^{t+1}_{i\to\mu}=(\lambda\mathbb{I}_r+C^t_{i\to\mu}+A^t_{i\to\mu})^{-1}b_{i\to\mu}^t$
\STATE $\forall 1\le \mu\le n, 1\le i \le d,~~\hat{c}^{t+1}_{i\to\mu}=(\lambda\mathbb{I}_r+C^t_{i\to\mu}+A^t_{i\to\mu})^{-1}$
\ENDFOR
\STATE
\RETURN Estimator $\hat{\w}$
\end{algorithmic}
\end{algorithm}
This recovers Algorithm \ref{alg:BP}.

\subsection{Equivalence rBP/GAMP}
The previous subsection established the asymptotic equivalence of BP \eqref{alg:MP} and rBP \ref{alg:BP} equations. We now further demonstrate that rBP \ref{alg:BP} is also asymptotically equivalent to GAMP \ref{alg:GAMP}, which has affords a simpler algorithm with less variables, and further bears connections to the fixed points of GD, see main text. The basis of this derivation is the observation that variables $\circ_{i\to \mu}$ (resp. $\circ_{\mu\to i}$) have a weak dependence on the target node $\mu$ (resp. $i$). Therefore, $V^t_{\mu\to i}\approx V^t_\mu, \Gamma^t_{\mu\to i}\approx \Gamma^t_\mu, \omega^t_{\mu\to i}\approx \omega^t_{\mu}, f^t_{\mu\to i}\approx f^t_\mu, g^t_{\mu\to i}\approx g^t_\mu, \eta^t_{\mu\to i}\approx \eta^t_\mu, A^t_{i\to\mu}\approx A^t_{i}, C^t_{i\to\mu}\approx C^t_i, b^t_{i\to\mu}\approx b^t_i, \hat{w}^t_{i\to\mu}\approx\hat{w}_i, \hat{c}^t_{i\to\mu}\approx \hat{c}^t_i$, where
\begin{align}
\label{app:intro_GAMP_var}
&(V_{\mu}^t)_{\ell\kappa}=\frac{1}{d}\sum\limits_{j} (x^\mu_{\ell j})(x^\mu_{\kappa j}) \hat{c}^t_{j\to \mu}, &&\Gamma_{\mu}^t=\frac{1}{d}\sum\limits_{j}\hat{w}^t_{j\to\mu}(\hat{w}^t_{j\to\mu})^\top, &&\omega_{\ell,\mu}^t=\frac{1}{\sqrt{d}}\sum\limits_{j} x^\mu_{\ell, j} \hat{w}_{j\to \mu}^t\notag\\
&A^t_{i}=-\frac{1}{d}\sum\limits_{\ell,\kappa=1}^L \sum\limits_{\nu}(x^\nu_{\ell i})(x^\nu_{\kappa i})g_{\ell\kappa, \nu\to i}^t, &&C^t_{i}=\frac{2}{d}\sum\limits_{\nu}\eta_{ \nu\to i}^t, &&b^t_{i}=\frac{1}{\sqrt{d}}\sum\limits_{\ell=1}^L
\sum\limits_{\nu}x^\nu_{\ell i} f_{\ell,\nu\to i}^t,\notag\\
&\hat{w}^{t+1}_{i}=(\lambda\mathbb{I}_r+C^t_{i}+A^t_{i})^{-1}b_{i}^t, &&\hat{c}^{t+1}_{i}=(\lambda\mathbb{I}_r+C^t_{i}+A^t_{i})^{-1}, &&f^t_{\ell,\mu}=\sum\limits_\kappa \scriptstyle (V_{\mu\to i}^t)^{-1}_{\ell\kappa}\left(\prox(y_\mu,\omega^t_{\mu},V^t_{\mu}, \Gamma^t_{\mu}, c^\mu)-\omega_{\mu}^t\right)_\kappa, \notag\\
& g_{\ell\kappa, \mu}^t=\nabla_{\omega_\kappa}f_{\ell, \mu}^t, && && \eta^t_{\mu}=\partial_3\ell\left(\scriptstyle y_\mu, \prox(y_\mu,\omega^t_{\mu},V^t_{\mu}, \Gamma^t_{\mu}, c^\mu), \Gamma^t_{\mu}, c^\mu\right).
\end{align}
This constitutes a smaller set of variables. We note, however, that the equations \eqref{susbsec:GAMP} are not yet closed, as they still involve $\circ_{i\to\mu},\circ_{\mu\to i}$ variables. In the following, we thus focus on obtaining a closed set of iteration equations, which will result in the GAMP algorithm \ref{alg:GAMP}.

\paragraph{Simplifying $V^t_{\mu}, \Gamma^t_{\mu}, A^t_{i}, C^t_i$} Let us first examine the variable $V^t_{\mu\to i}\approx V^t_{\mu}$. We can further simplify $V^t_{\mu}$ as
\begin{align}
    (V^t_{\mu})_{\ell\kappa}= \frac{1}{d}\sum\limits_{j}(x^\mu_\ell)_j(x^\mu_\kappa)_j \hat{c}_{j\to \mu}^t=\frac{1}{d}\sum\limits_{j}(x^\mu_\ell)_j(x^\mu_\kappa)_j \hat{c}_{j}^t+O_d(\sfrac{1}{d}).
\end{align}
By the same token,
\begin{align}
    \Gamma^t_{i\to\mu } =\frac{1}{d}\sum\limits_{j}\hat{w}_j^t(\hat{w}_j^t)^\top +O_d(\sfrac{1}{d}).
\end{align}
and 
\begin{align}
    A^t_i=-\frac{1}{d}\sum\limits_{\ell,\kappa=1}^L \sum\limits_{\nu}(x^\nu_{\ell i})(x^\nu_{\kappa i})g_{\ell\kappa, \nu}^t+O_d(\sfrac{1}{d}),
\end{align}
and finally
\begin{align}
    C^t_{i}=\frac{2}{d}\sum\limits_{\nu}\eta_{ \nu}^t
+O_d(\sfrac{1}{d}).
\end{align}

\paragraph{Onsager terms}
We now turn our attention to $\omega^t_{\ell,\mu}$. In this term, one may not straightforwardly replace $\hat{w}^t_{j\to \mu}$ by $\hat{w}^t_j$, as the neglected term is correlated with $x^\mu_{\ell,j}$ and results in a $\Theta_d(1)$ contribution when summed over. To make progress, let us more carefully control $\hat{w}^t_{j\to\mu}$ as 
\begin{align}
    \hat{w}^t_{j\to\mu}=\hat{w}^t_{j}-\frac{1}{\sqrt{d}}\sum\limits_\ell x^\mu_{\ell j}(\lambda\mathbb{I}_r+C^t_j+A^t_j)^{-1} f^t_{\ell,\mu}+O_d(\sfrac{1}{d}).
\end{align}
Thus, 
\begin{align}
    \omega^t_{\ell,\mu}&=\frac{1}{\sqrt{d}}\sum\limits_j x^\mu_{\ell,j}\hat{w}^t_j -\frac{1}{d} \sum\limits_j\sum\limits_{\kappa} x^\mu_{\ell,j}
    x^\mu_{\kappa,j} \hat{c}^t_j f^t_{\kappa,\mu}+o_d(1)\notag\\
    &=\frac{1}{\sqrt{d}}\sum\limits_j x^\mu_{\ell,j}\hat{w}^t_j -\sum\limits_{\kappa}(V^t_\mu)_{\ell,\kappa} f^t_{\kappa,\mu}.
\end{align}
The last expression to close, bearing on $b^t_i$, can be treated in the same fashion. Carefully expanding $f^t_{\ell,\nu\to i}$ as
\begin{align}
    f^t_{\ell,\nu\to i}=f^t_{\ell,\nu}- \frac{1}{\sqrt{d}}\sum\limits_\kappa (g^t_\nu)_{\ell\kappa} x^\mu_{\kappa, i} \hat{w}^t_i+O_d(\sfrac{1}{d}),
\end{align}
we reach
\begin{align}
    b_i^t&=\frac{1}{\sqrt{d}}\sum\limits_\ell \sum\limits_\nu x^\nu_{\ell, i} f^t_{\ell,\nu} - \frac{1}{d}\sum\limits_{\ell,\kappa} \sum\limits_\nu x^\nu_{\ell, i}x^\nu_{\kappa, i} (g^t_{\nu})_{\ell,\kappa}\hat{w}_i+o_d(1)\notag\\
    &=\frac{1}{\sqrt{d}}\sum\limits_\ell \sum\limits_\nu x^\nu_{\ell, i} f^t_{\ell,\nu} +A^t_i \hat{w}_i^t.
\end{align}

\paragraph{Summary of GAMP equations}
We have now closed the set of equations \ref{app:intro_GAMP_var}. Summarizing, we reach the iterative scheme

\begin{algorithm}[H]
\caption{GAMP}
\begin{algorithmic}
\STATE \textbf{Inputs} : $\mathcal{D
}=\{\x^\mu, y^\mu,c^\mu\}_{\mu=1}^n $
\STATE \textbf{Initialize} $ \forall 1\le \mu\le n,~~1\le i\le d,~~\hat{w}^0_{i}=0_r, \hat{c}^0_{i}=\mathbb{I}_{r}$
\FOR{$t\le t_{\max}$}
\STATE $\forall 1\le \ell,\kappa\le L,1\le \mu\le n,~~(V_{\mu}^t)_{\ell\kappa}=\frac{1}{d}\sum\limits_{i} (x^\mu_{\ell i})(x^\mu_{\kappa i}) \hat{c}^t_{i} $
\STATE $\Gamma^t=\frac{1}{d}\sum\limits_{i}\hat{w}^t_{i}(\hat{w}^t_{i})^\top$
\STATE $\forall 1\le \ell,1\le \mu\le n, ~~ \omega_{\ell,\mu}^t=\frac{1}{\sqrt{d}}\sum\limits_{i} x^\mu_{\ell, i} \hat{w}_{i}^t-\sum\limits_\kappa(V^t_\mu)_{\ell\kappa} f_{\kappa\mu} $
\STATE $\forall 1\le \ell,1\le \mu\le n,$$~~ 
f^t_{\ell,\mu}=\sum\limits_\kappa (V_{\mu}^t)^{-1}_{\ell\kappa}\left(\prox(y_\mu,\omega^t_{\mu},V^t_{\mu}, \Gamma^t, c^\mu)-\omega_{\mu}^t\right)_\kappa  $
\STATE $\forall 1\le \mu\le n,~~\eta^t_{\mu}=\partial_3\ell\left(y_\mu, \prox(y_\mu,\omega^t_{\mu},V^t_{\mu}, \Gamma^t, c^\mu), \Gamma^t_{\mu}, c^\mu\right)$
\STATE $\forall 1\le \ell,\kappa\le L,1\le \mu\le n,~~
g_{\ell\kappa, \mu}^t=\nabla_{\omega_\kappa}f_{\ell,\mu}^t
$
\STATE $ 1\le i \le d,~~ A^t_{i}=-\frac{1}{d}\sum\limits_{\ell,\kappa=1}^L \sum\limits_{\mu}(x^\mu_{\ell i})(x^\mu_{\kappa i})g_{\ell\kappa, \mu}^t$
\STATE $ C^t=\frac{2}{d}\sum\limits_{\mu}\eta_{\mu}^t$
\STATE $1\le i \le d,~~b^t_{i}=\frac{1}{\sqrt{d}}\sum\limits_{\ell=1}^L
\sum\limits_{\mu}x^\mu_{\ell i} f_{\ell,\mu}^t+A^t_i\hat{w}^t_i$
\STATE $\forall 1\le \mu\le n, 1\le i \le d,~~\hat{w}^{t+1}_{i}=(\lambda\mathbb{I}_r+C^t+A^t_{i})^{-1}b_{i}^t$
\STATE $\forall 1\le \mu\le n, 1\le i \le d,~~\hat{c}^{t+1}_{i}=(\lambda\mathbb{I}_r+C^t+A^t_{i})^{-1}$
\ENDFOR
\STATE
\RETURN Estimator $\hat{\w}$
\end{algorithmic}
\end{algorithm}

which recovers Algorithm \ref{alg:GAMP}.